# Pre-trained knowledge elevates large language models beyond traditional chemical reaction optimizers

Robert MacKnight,[1] Jose Emilio Regio,[2] Jeffrey G. Ethier,[3] Luke A. Baldwin,[3] Gabe Gomes*[1,2,4,5]

1. Department of Chemical Engineering, Carnegie Mellon University, Pittsburgh, PA 15213, USA
2. Department of Chemistry, Carnegie Mellon University, Pittsburgh, PA 15213, USA
3. Materials and Manufacturing Directorate, Air Force Research Laboratory, Wright-Patterson AFB, OH 45433, USA
4. Machine Learning Department, Carnegie Mellon University, Pittsburgh, PA 15213, USA
5. Wilton E. Scott Institute for Energy Innovation, Carnegie Mellon University, Pittsburgh, PA 15213, USA

* corresponding author: gabegomes@cmu.edu

# Abstract

Modern optimization in experimental chemistry employs algorithmic search through black-box parameter spaces. Here we demonstrate that pre-trained knowledge in large language models (LLMs) fundamentally changes this paradigm. Using six fully enumerated categorical reaction datasets (768–5,684 experiments), we benchmark LLM-guided optimization (LLM-GO) against Bayesian optimization (BO) and random sampling. Frontier LLMs consistently match or exceed BO performance across five single-objective datasets, with advantages growing as parameter complexity increases and high-performing conditions become scarce (<5% of space). BO retains superiority only for explicit multi-objective trade-offs. To understand these contrasting behaviors, we introduce a topology-agnostic information theory framework quantifying sampling diversity throughout optimization campaigns. This analysis reveals that LLMs maintain systematically higher exploration Shannon entropy than BO across all datasets while achieving superior performance, with advantages most pronounced in solution-scarce parameter spaces where high-entropy exploration typically fails—suggesting that pre-trained domain knowledge enables more effective navigation of chemical parameter space rather than replacing structured exploration strategies. To enable transparent benchmarking and community validation, we release Iron Mind (https://gomes.andrew.cmu.edu/iron-mind), a no-code platform for side-by-side evaluation of human, algorithmic, and LLM optimization campaigns with public leaderboards and complete trajectories. Our findings establish that LLM-GO excels precisely where traditional methods struggle: complex categorical spaces requiring domain understanding rather than mathematical optimization. For practitioners, deploy LLM-GO for high-dimensional categorical problems under tight experimental budgets; prefer BO for continuous parameters, multi-objective optimization, or unfamiliar chemical spaces. These results demonstrate that foundation model knowledge can effectively bypass exploration-exploitation paradigms that have dominated experimental optimization for decades.

# Introduction

In chemistry, the optimization of experiments requires searching for the process parameters that optimize a predefined objective or property (e.g., yield, enantiomeric excess, bandgap).[1,2] However, the parameter space is often complex and multi-dimensional, requiring considerable time and resources to explore. A common strategy, often reported via anecdote, relies on chemical intuition. This approach can be valuable due to the fundamental complexity and unpredictability inherent in chemical systems.[3,4] However, intuitive decision-making may not consistently lead to optimal outcomes. To overcome these limitations, researchers turn to more systematic methodologies. One-factor-at-a-time (OFAT)[5–7] strategies follow a more structured approach, systematically varying one parameter while holding others constant.[1] In Design of Experiments (DoE)[8–10], a statistical model is constructed to describe the relationship between the process parameters and the objective function. Intuition-based optimization and OFAT strategies are inefficient and risk identifying suboptimal parameters. DoE also suffers from inefficiency since the number of experiments increases exponentially with parameter count and is best suited for continuous variables, while categorical variables require special considerations.[1,2,11]

Bayesian optimization (BO) is a global optimization algorithm for expensive black-box objective functions.[2,4,12,13] The algorithm combines assumptions about the response surface along with previous observations of the objective function to train a probabilistic surrogate model. An acquisition function takes the surrogate model as input and identifies the next experiments to perform. After running the experiment, the new observations are used to update the surrogate model. This process repeats until either an objective criterion is met, or resources are exhausted. Through a balance of exploration and exploitation, encoded by the acquisition function, BO is able to efficiently navigate parameter space and thus find better optima with fewer experiments.[2,4,13] BO has been used to optimize a wide range of parameter spaces in chemistry and materials science.[14] Some examples are: design of energy storage materials,[15] flow synthesis of pyridinium salts,[16] Cu-catalyzed C-N coupling of pyrazines,[4] and design of electrochemical devices.[17] In this paper, we benchmark BO on six chemical reaction datasets, outlined in the methods section. Two of these datasets have been previously studied as benchmark datasets.[2,18]

Many have investigated the impact of leveraging descriptor-based representations for BO of complex chemical reactions.[2,19–21] Implementation is challenging; effective descriptor selection requires deep domain knowledge to capture relevant molecular properties for modeling the objective, and can even hinder performance.[20,21] Another complexity of experiment optimization, is that problems can be multi-objective. Multi-objective approaches,[22] like those facilitated by tools like Chimera,[23] often require some human logic. These methods typically rely on expert-defined weights, hierarchies, or thresholds that may not fully capture the complexity of the optimization landscape. Additionally, extracting interpretable rationale from BO campaigns is difficult. Ultimately, these challenges highlight the need for more interpretable optimization methods that require less specialized expertise, such as those leveraging Large Language Models (LLMs).

The rise of LLMs offers a new approach to the optimization of experiments. Based on the transformer architecture,[24] LLMs are text-based foundational models that have found success in domains such as natural language processing and code generation.[25–27] Of recent interest is the performance of LLMs in the chemistry domain, where LLMs have been shown to outperform state-of-the-art machine learning models especially in the low-data regime.[27,28] LLM models such as MolGPT[29] and Llamol[30] provide avenues for molecular discovery via generation from a set of conditions or properties. Researchers presented MolLEO[31], incorporating LLMs into evolutionary algorithms, concluding that chemistry-aware LLMs result in superior performance across multiple benchmarks. Furthermore, it has been shown that LLMs carry useful information (via embeddings) to aid in Bayesian optimization for materials discovery, particularly when a fine-tuned chemistry-specific LLM is used.[32] Other language models such as Bidirectional Encoder Representations from Transformers (BERT) can take chemical reagents as SMILES notations and predict reaction yields or reaction success.[33,34] The capabilities of LLMs can additionally be expanded when given access to tools. A number of works have demonstrated high-level capabilities of LLM-based agentic systems in the chemistry and biochemistry domains.[25,26,35–37]

While LLMs offer promising new approaches, understanding how they compare to traditional methods requires analyzing their underlying sampling strategies. Central to data-driven optimization approaches is how methods balance exploration of unknown parameter regions versus exploitation of promising areas.[38–40] BO relies on acquisition functions to navigate this exploration-exploitation trade-off, but these mathematical frameworks may not capture the strategic reasoning that expert chemists employ. Understanding and analyzing these sampling behaviors becomes crucial for evaluating whether different optimization methods are making chemically sound decisions throughout campaigns. Recent work has demonstrated the value of entropy-based metrics to quantify exploration in optimization by measuring the spatial distribution of observation points in the objective space.[41] However, this approach assumes well-structured objective landscapes where spatial dispersion correlates with effective exploration. We introduce a fundamentally different approach that computes Shannon entropy[42] over the parameter selections, providing a topology-agnostic method for quantifying sampling strategies that does not rely on assumptions about the objective space structure.

Human experimental design involves complex decision-making that combines pattern recognition, theoretical knowledge, and intuitive reasoning. Scientists routinely "warm start" optimization campaigns by mining literature databases (Google Scholar,[43] SciFinder,[44] Reaxys[45]) to inform initial hypotheses and guide subsequent experiments. Despite the growing adoption of algorithmic approaches like BO, direct performance comparisons with human experts remain scarce. This gap stems partly from a lack of standardized datasets and evaluation frameworks designed to capture human decision-making processes rather than simply test machine learning (ML) algorithms. Without systematic comparisons, it remains difficult to evaluate when and why data-driven methods outperform human expertise in real-world chemistry scenarios.

Beyond performance evaluation, capturing human optimization trajectories offers another critical advantage: enhanced trustworthiness of AI-driven approaches. By cross-referencing LLM reasoning and optimization strategies with human decision-making patterns, we can provide validation frameworks that demonstrate when and why AI recommendations align with expert human judgment. This transparency could prove pivotal in encouraging adoption among practitioners who remain skeptical of "black box" AI recommendations in experimental settings. To bridge the gap and train new machine learning models that understand how chemists think, it is crucial to capture and codify the qualitative reasoning and experiential insights that underlie human decision-making. Incorporating these qualitative insights alongside quantitative data could lead to more nuanced and effective AI-driven experimental strategies.

To address these challenges, we present LLM-guided optimization (**LLM-GO**) and develop a user-friendly web application (https://gomes.andrew.cmu.edu/iron-mind) to systematically capture human optimization decisions for comparison with AI approaches. We benchmark our LLM-GO strategy on six chemical reaction datasets, generating detailed rationales for each experimental suggestion. To quantify and compare the sampling strategies of these different optimization methods, we introduce a Shannon entropy-based analysis framework that evaluates parameter selection behavior independent of objective space structure. This approach provides insights into how LLM-GO and traditional algorithms navigate parameter spaces in chemistry, offering both performance benchmarks and mechanistic understanding of optimization decision-making.

# Methods

## Datasets

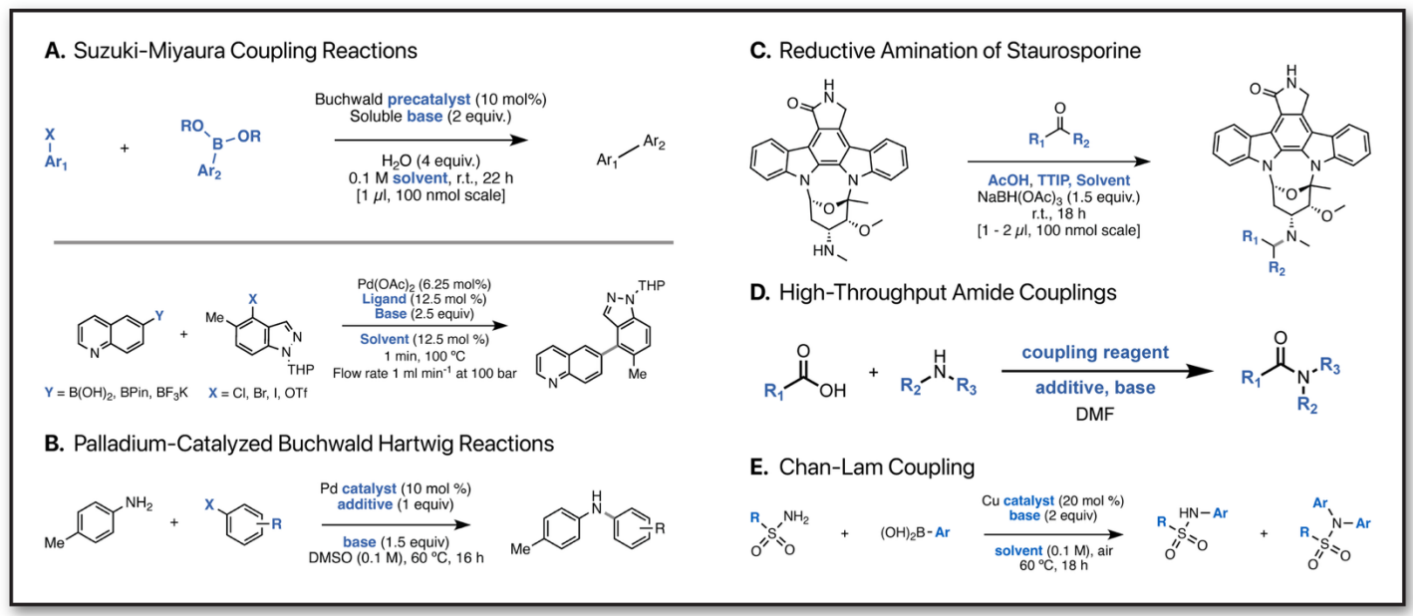

**Figure 1 — Benchmarking datasets.** All datasets were gathered from the literature and translated into lookup tables for the fully mapped parameter space in relation to the measured objective. A) Two Suzuki-Miyaura coupling datasets were used, one measuring conversion and another measuring yield. B.) A dataset of Buchwald Hartwig Reactions measuring yield. C.) A dataset of reactions for the reductive amination of Staurosporine measuring conversion. D.) An amide coupling reaction dataset measuring yield E.) Lastly, a multi-objective Chan-Lam coupling dataset of primary sulfonamides, quantifying the yield of the desired (mono) and undesired (bis) products. To compare with all other benchmark datasets, the objective value is taken as the difference between desired and undesired yields.

## Suzuki-Miyaura Coupling Reactions

Two distinct datasets were evaluated for Suzuki-Miyaura coupling reactions (**Figure 1A**), addressing different optimization objectives. The first dataset from Gesmundo et al. contains 960 measured conversion rates.[46] This dataset contains three categorical parameters: 1.) aryl halide, 2.) boronate building block, 3.) reaction conditions. There is a total of eight aryl halides, ten boronate building blocks, and twelve sets of reaction conditions. A set of reaction conditions contains a catalyst, base, solvent, and cosolvent. The conversion distribution, **Figure 2**, reveals a bimodal pattern with peaks at both very low (~0%) and very high (~100%) conversion values. The second dataset from Perera et al. focused on reaction yield as the primary objective, exploring a parameter space of five categorical parameters: 1.) electrophile, 2.) nucleophile, 3.) base, 4.) ligand, 5.) solvent.[47] A total of four electrophiles, three nucleophiles, seven bases, 11 ligands, and four solvents were considered, making for a dataset of 3,696 measured yields. The yield distribution, **Figure 2**, contains a peak in the 10-25% range and a broad distribution of higher-yielding conditions extending to 100%. Both datasets represent valuable benchmarks for optimization algorithms, presenting different challenges in navigating the complex parameter spaces of these widely used coupling reactions in organic synthesis.[48]

## Palladium-Catalyzed Buchwald Hartwig Reactions

This dataset (**Figure 1B**) from Ahneman et al. examines Buchwald-Hartwig amination reactions, focusing on yield optimization across a diverse parameter space.[49] The dataset explores 4 categorical parameters: 1.)

base, 2.) ligand, 3.) aryl halide, 4.) additive. A total of three bases, four ligands, 16 aryl halides, and 24 additives were investigated, providing a dataset of 4,608 measured yields. The yield distribution, **Figure 2**, for this dataset contains a peak at 0% and a broad distribution up to 100%. The Buchwald-Hartwig reaction represents a critical tool in medicinal chemistry and materials science, making this dataset particularly relevant for optimizing synthetic routes to complex nitrogen-containing compounds.[50]

## Reductive Amination of Staurosporine

This dataset (**Figure 1C**) investigates the reductive amination of Staurosporine, exploring the impact of five parameters on reaction conversion.[46] The parameters include: 1.) substrate, 2.) acetic acid equivalents, 3.) titanium isopropoxide (TTIP) equivalents, 4.) solvent, 5.) reaction concentration. A total of 16 substrates, three acetic acid equivalents, four TTIP equivalents, two solvents, and two reaction concentrations were evaluated. The dataset contains 768 measured conversions. The conversion distribution, **Figure 2**, reveals a strongly skewed pattern with approximately 35% of experiments showing near-zero conversion, followed by a long tail of moderate to high-converting reactions. This challenging distribution, with many failed reactions, presents an interesting optimization problem typical of complex pharmaceutical transformations.

## High-Throughput Amide Couplings

A fully mapped dataset (**Figure 1D**) measuring the percent yield for amide couplings was extracted from the literature.[51] This dataset explores the impact of three categorical parameters: 1.) carboxylic acid, 2.) amine, 3.) reaction conditions. There is a total of three carboxylic acids, three amines, and 72 sets of reaction conditions. A set of reaction conditions contains a coupling reagent, base, additive, and solvent. The solvent was kept constant as dimethylformamide (DMF). In total this dataset contains 648 measured yields. The yield distribution, **Figure 2**, is dominated by poor performing solutions (yield ~0-5 percent), while all other performance levels are somewhat uniformly distributed. It is important to note the well-explored nature of this reaction type, along with its prevalence in medicinal chemistry.[52]

## Chan-Lam Coupling of Primary Sulfonamides

This dataset (**Figure 1E**) is the only multi-objective benchmark dataset. The percent yield of the desired, mono-arylated product, and undesired, bis-arylated product, was measured across five categorical parameters.[53] A total of 10 sulfonamides, two boronic acids, four copper catalysts, four solvents, and six bases were evaluated. The dataset contains 5684 measured desired/undesired yields. In this, there are 1920 unique parameters combinations, where a given parameter set may yield multiple measurements. To compare with all other benchmark datasets, we compute a weighted selectivity metric:

$$s = \left(\frac{desired\ yield}{desired\ yield + undesired\ yield}\right) * desired\ yield \quad (1)$$

When visualizing results the objective value is taken as the *lower-bound weighted selectivity* within a group of multiple measurements. However, during the optimization process all methods receive all measurements associated with a given set of conditions. The lower-bound weighted selectivity distribution, **Figure 2**, shows that conditions resulting in little selectivity toward the desired product are most prevalent. This dataset is distinct from the others due to its multi-objective nature where one objective (desired yield) is sought to be maximized while the other is minimized (undesired yield).

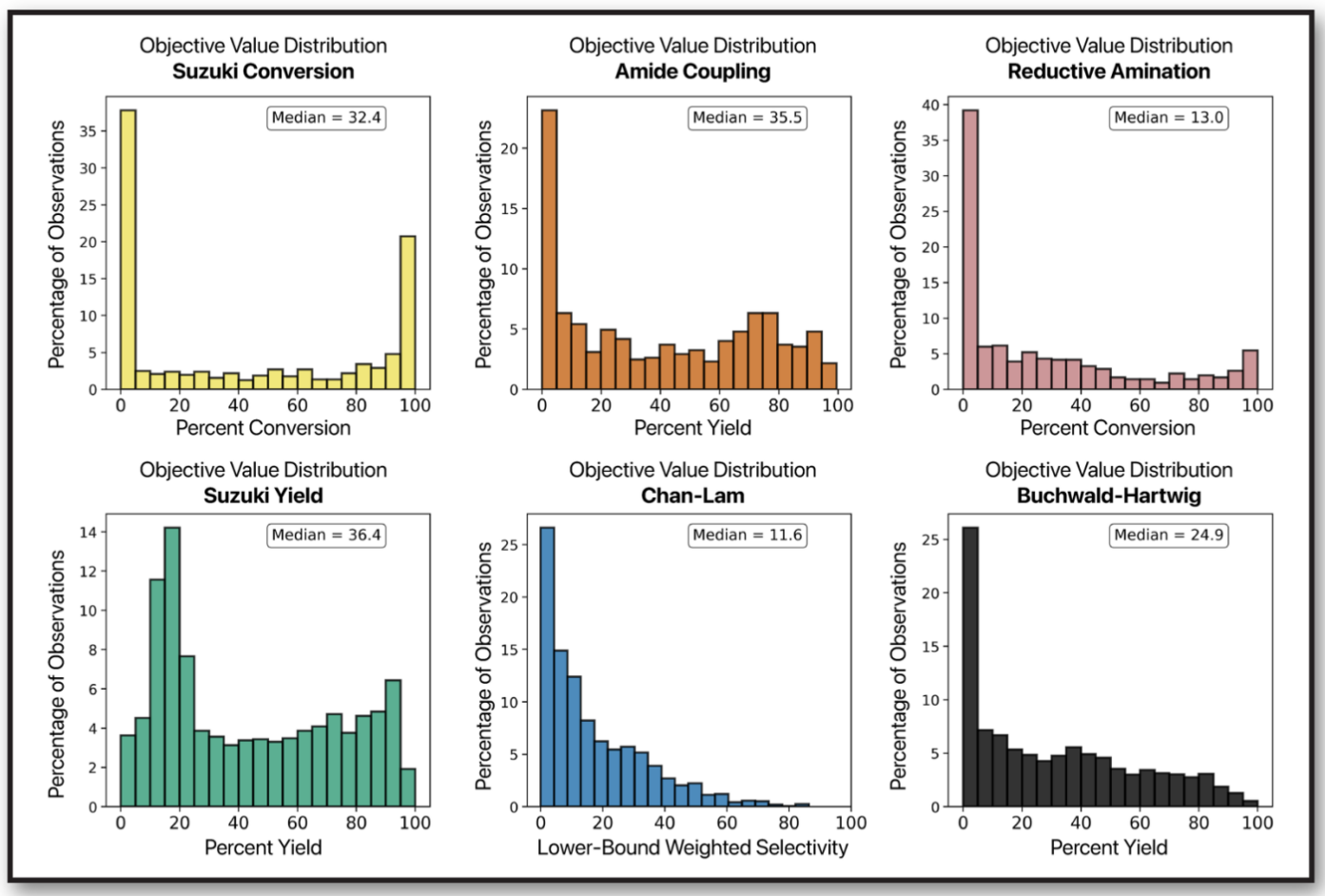

**Figure 2 — Objective value distributions.** Each histogram illustrates the objective value distribution for a given dataset. Each dataset is assigned a color, colors are used to refer to a given dataset throughout this paper. Each histogram was generated with 20 bins. Raw counts of observations in each bin were converted to a percentage of observations for each bin.

## Quantifying the optimization complexity of benchmark datasets

Analyzing the objective value distributions presented in **Figure 2** gives an idea of the optimization campaign complexity for each benchmark dataset. To further articulate complexity, we collect six features from each dataset, shown in **Figure 3**. First, we calculate the average number of parameter options for each dataset (AOP), this is facilitated by the fact that each dataset is fully discrete. Next, we gather the number of manipulatable parameters (NP) for each dataset. The reported parameter space size is simply the product of options for each parameter (PSS). We report two measures related to the objective value distributions shown in **Figure 2**, skewness (SKEW) and scarcity index (SI). The skewness measures the departure from normality in the objective values for each dataset. To calculate this metric, we leveraged the SciPy implementation, which obtains the Fisher-Pearson coefficient of skewness.[54] For all datasets, the skewness value is positive, meaning poor performing measurements are dominant. However, the extent to which each dataset is skewed in this manner varies greatly. A scarcity index was computed for each dataset as the one minus the fraction of values greater than the 95 percent of the maximum objective value. Lastly, we train a Random Forest Regressor (RFR) on one-hot-encoded representations to obtain normalized parameter importances. From here we compute the standard deviation in parameter importances. To obtain a metric positively correlated with complexity, we report one minus the parameter importance variation, referred to as parameter importance balance (PIB). We min-max normalize each of the six metrics across all benchmark datasets to prepare radar plots. From the radar plots we compute areas and normalize with respect to the largest area (Buchwald-Hartwig), which are shown underneath each plot. To compute the complexity metrics for the sole multi-objective dataset of Chan-Lam couplings, we take the lower-bound weighted selectivity (**Equation 1**) for a group of measurements as the objective value.

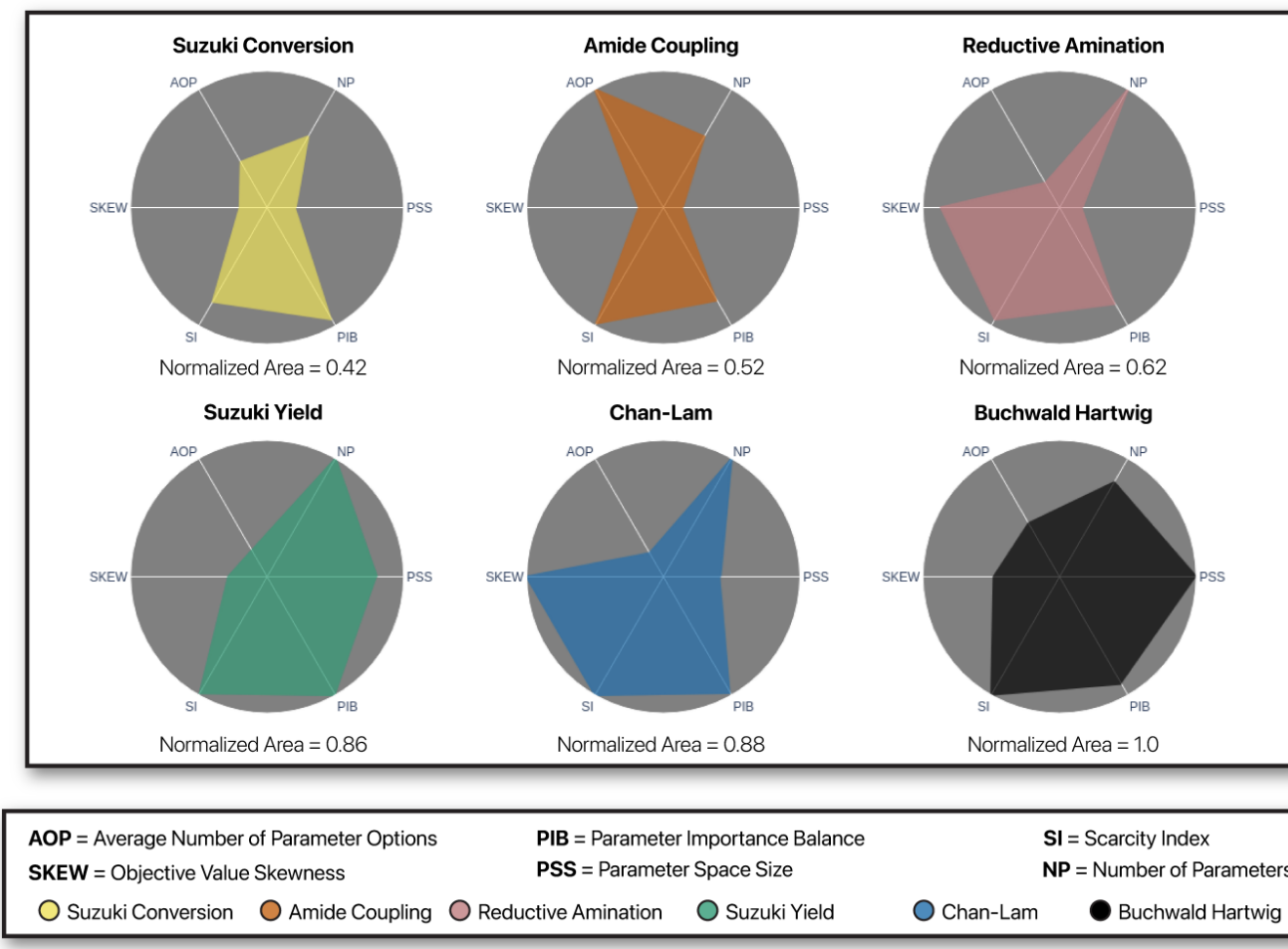

**Figure 3 — Reaction optimization campaign complexity.** Each dataset possesses a radar plot indicating the normalized metrics. The area of each radar plot was used to position each dataset relative to the other benchmark datasets. Thus, the "normalized area" can be interpreted as a complexity score for a reaction optimization campaign relative to the Buchwald-Hartwig dataset.

## Optimization strategies

We evaluated two optimization approaches for experimental design: a BO strategy using an established implementation,[55] and our LLM-GO strategy. As a baseline comparison, we implemented a random sampling strategy ("random baseline") that uniformly selects parameter combinations from the available parameter space without replacement, ensuring no duplicate experiments are suggested during optimization.

### Random sampling baseline

Random sampling served as a baseline method that samples uniformly without replacement from available parameter combinations. For the fully categorical/discrete parameter spaces in this study, all possible combinations are enumerated and sampled without duplication within the campaign. This implementation provides a naive baseline for comparison, representing the performance achievable through uninformed parameter space exploration while avoiding wasted experimental budget on repeated combinations.

### Bayesian optimization baseline

For a second baseline comparison, we utilized the BO implementation from the Atlas[55] package, a Python library for automated experiment optimization. This implementation uses Gaussian Process Regression (GPR) to approximate the objective function, providing both mean predictions and uncertainty estimates across the parameter space. A schematic of this optimization strategy is shown in **Figure 4A.** Furthermore, this is an outline of the BO method:

> 1. **Train** a surrogate model (GP) on all observed data
> 2. **Compute** the acquisition function for each unobserved parameter combination
> 3. **Recommend** a batch of suggestions to test that maximize the acquisition function

Specific details pertaining to initial point generation, kernels, and acquisition function optimizers can be found in the Supporting Information (**Section SJ**).

For each optimization campaign, we employed one of three standard acquisition functions: expected improvement (EI), probability of improvement (PI), or upper confidence bound (UCB). These functions represent different strategies for balancing exploration and exploitation in the parameter space.

The categorical parameters were encoded using one-hot encoding (OHE) or molecular descriptors. For the descriptors approach we follow the work done by Shields et al. in the development of Experimental Design via Bayesian Optimization (EDBO),[2] which leverages Mordred[56] descriptors. For each categorical component, all molecular descriptors are computed. From here, uninformative features (zero variance, non-numeric) are removed, then the features are adaptively filtered. More information on feature selection can be found in the Supporting Information (**Section SJ**).

For the multi-objective Chan-Lam coupling dataset, all BO methods leverage the Chimera[23] achievement scalarizing function (ASF) to conduct the optimization for the maximization of desired yield and minimization of undesired yield. The relative tolerance for each objective was set to 0.3. Chimera hierarchically optimizes desired yield first, accepting candidate solutions with predicted desired yields in the top 30% of observed performance, before optimizing undesired yield as the secondary objective, constraining candidate solutions to predicted undesired yields in the top 30% of observed performance. This hierarchical scalarization is applied to the surrogate model predictions during acquisition function optimization, ensuring balanced and selective multi-objective optimization throughout the search process.

### Large Language Model-guided optimization

We developed an optimization strategy that leverages a LLM to guide experimental design decisions. The optimization process is structured through a specific prompting strategy that provides information to the LLM in two stages. First, is the system prompt design, here the following information is provided to the model:

> 1. Complete parameter space definition
>     a. Parameter names and types (categorical/continuous/discrete)
>     b. Valid options for categorical parameters
>     c. Bounds for continuous parameters
>     d. Allowed values for discrete parameters
> 2. Optimization objectives and their goals (minimize/maximize)
> 3. The number of parameter combinations to suggest per iteration (batch size)
> 4. Key guidelines for optimization
>     a. Avoid infeasible experiments
>     b. Minimize the number of experiments
>     c. Avoid suggesting previously tested parameter combinations
>     d. Consider the physical/chemical meaning of observed data

A generalized and truncated version of the system prompt used is shown in **Figure 4C**.

Second, each iteration follows a specific prompt design allowing the model to reason and provide an explanation for suggested experiments. At each iteration, the LLM receives a complete history of all previously observed experiments, including parameter values tested, resulting objective values, and any infeasible experiments. The LLM then follows a structured response protocol:

> 1. **Analyze** trends in the observed data
> 2. Form a **hypothesis** about important factors
> 3. Provide explicit **reasoning** for the next suggestion
> 4. **Recommend** a batch of suggestions to test

To better ensure suggested parameter values are in the parameter space, we leverage function calling to enforce the model to select values from the list of valid options for a given parameter, although this does not always hold. On invalid suggestions, we mark the objective value as nan and proceed as normal, with the suggestion counting against the budget, **Figure S11** summarizes the rates. Within each campaign, r, we count $D_r$: the number of unique parameter configurations suggested ≥ 2 times; **Figure 6** shows the distribution of $D_r$ across 20 campaigns.

For LLMs that allow for setting the model temperature, the value was held constant at a value of 0.7. For Anthropic models, we scaled the value of 0.7 (from the range of 0-2 to the range of 0-1) linearly, resulting in a constant model temperature of 0.35. The specific API model versions can be found in **Table S1**. Responses from the model were limited to a maximum of 8,192 tokens. For Anthropic models that produce thinking tokens we allow for up to 4,096 thinking tokens and 4,096 output tokens. Both OpenAI and Google models leverage the OpenAI API, thus thinking/reasoning tokens were specified by level (i.e. 'low' or 'medium'). This setting corresponds to a number of allowed thinking/reasoning tokens.

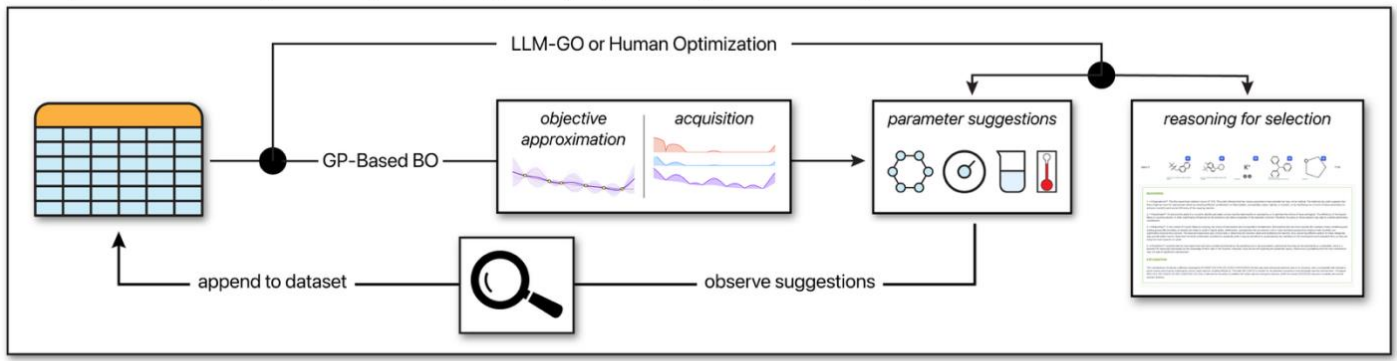

**A. Optimization Strategies — Bayesian, Large Language Model, and Human Optimization**

**B. Generalized Experiment Optimization Algorithm**

**Algorithm:** Experiment Optimization
**Result:** Optimize **objective function**, $f(x)$
**Input: batch size** $b$, **budget** $B$
**while** $i < B$ **do**
  **suggest** $\{x_{next}^{(1)}, x_{next}^{(2)}, \ldots, x_{next}^{(b)}\}$;
  **observe** $\{f(x_{next}^{(1)}), f(x_{next}^{(2)}), \ldots, f(x_{next}^{(b)})\}$;
  $i \leftarrow i + b$;
**end**

**C. Automatic System Prompt Generation**

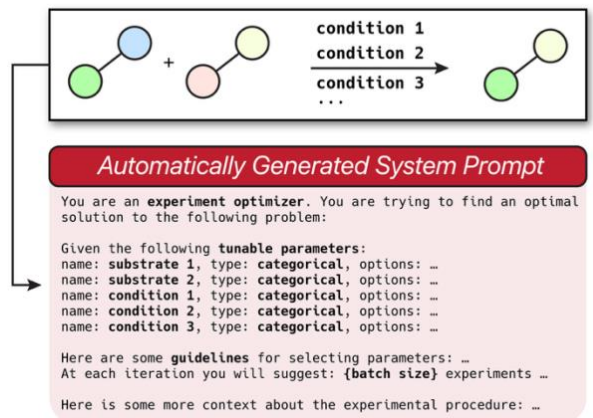

**Figure 4 — Summary of Optimization Strategies.** (A) Shown is the information flow for each optimization method. For BO, existing data is leveraged to train a surrogate model and evaluate the objective function, yielding parameter suggestions. For LLM-GO, existing data is sent directly to the LLM where both parameter suggestions and reasoning for those suggestions is returned, mirroring the process of human optimization. (B) Experiment optimization seeks to optimize an objective function through iterative experiment suggestions of batch size b, until a budget B is exhausted. (C) The LLM-GO method transforms parameter space definitions and any existing procedural information to automatically generate system prompts for the LLM optimizer.

## Shannon entropy analysis of sampling strategies

To quantify and compare the sampling strategies of different optimization methods, we developed a Shannon entropy-based analysis framework. For each optimization method and dataset, we analyze parameter selection behavior across 20 independent optimization runs.

At each optimization run $r$, we compute parameter selection diversity across all iterations as follows: (1) For each parameter, we count how many times each possible parameter option was selected across all $T$ iterations within that single run (Equation 1), (2) convert these counts to selection probabilities within the run (Equation 2), (3) calculate the normalized Shannon entropy for each parameter to quantify diversity on a 0-1 scale (Equation 3), and (4) average the normalized entropies across all parameters in the dataset to obtain a single cumulative diversity measure for that run (Equation 4).

To outline the mathematical formulation of this analysis, take a dataset with $K$ parameters. Each parameter, $p_i$, has a fixed set of options, $V_i$:

$$V_i = \{v_{i1}, v_{i2}, \ldots, v_{in_i}\}$$

The number of options for a parameter in the dataset is denoted as $n_i$. A number of runs, $R$, are conducted each with $T$ iterations, First, we define a count function for each individual run $r$.

$$c^r_{ij} = number\ of\ times\ run\ r\ selected\ v_{ij}\ for\ p_i\ across\ all\ T\ iterations \tag{2}$$

From this, the probability of selecting a given parameter option within run $r$ is computed:

$$P^r_{ij} = \frac{c^r_{ij}}{T}; \sum_{j=1}^{n_i} P^r_{ij} = 1 \tag{3}$$

This yields a probability distribution across all options $V_i$ for parameter $p_i$ within run $r$:

$$H^r_i = -\sum_{j=1}^{n_i} \frac{P^r_{ij} \times log_2(P^r_{ij})}{log_2(n_i)} \tag{4}$$

We follow standard convention in computing Shannon entropy $H^r_i$, where $P \times log_2(P) = 0$, if $P = 0$.

Finally, the normalized Shannon entropy $H^r$ for each parameter within run $r$ is averaged, giving a cumulative exploration measure for that run:

$$H^r = \frac{1}{K} \sum_{i=1}^{K} H^r_i \tag{5}$$

This cumulative entropy metric quantifies the overall sampling diversity within individual optimization runs, where high entropy (approaching 1) indicates broad exploration across parameter options throughout the entire run and low entropy (approaching 0) indicates exploitation of specific parameter values. For each run, we count how frequently each parameter option was selected across all $T$ iterations, convert these counts to probabilities, and compute normalized Shannon entropy for each parameter. The normalization ensures parameters with different numbers of options are comparable, while averaging across parameters provides an overall measure of more diverse sampling strategies, while methods with lower variance in cumulative entropy exhibit more consistent behavior across repeated runs. We also report entropy-to-best (cumulative entropy up to the first occurrence of the run's best objective), used in **Section SF**.

## Olympus integration

To ensure broad applicability of our optimization approach, we integrated the LLM-guided optimizer with the Olympus framework, a Python library for standardizing experimental optimization tasks. This integration allows our methods to be readily applied to any dataset in the Olympus database without modification. Olympus provides a unified interface for handling diverse experimental datasets, managing parameter spaces, and tracking optimization progress.

## No code web application

To increase the transparency of the experimental optimization techniques and facilitate data collection from human experimenters, we developed a web-based platform that requires no coding experience to operate. This platform serves as an important methodological contribution, enabling direct comparison between human, algorithmic, and LLM-driven optimization approaches within a unified interface. Furthermore, we present a [leaderboard](#) ranking optimization method by median/mean performance. Human optimizers are given the option to publish their optimization campaigns to the public leaderboard.

The web application supports three distinct optimization modalities: human, Bayesian, and LLM optimization. We provide an intuitive interface for users to propose experimental parameters based on their domain expertise and reasoning. Users can review past experimental results, suggest new combinations, and document their decision-making process through structured explanation fields. This captures both the quantitative performance and qualitative reasoning behind experimental decisions. The web application can be found here: https://gomes.andrew.cmu.edu/iron-mind.

To enable systematic comparison with LLM optimization strategies and ensure high-fidelity reasoning data for analysis, we encourage human users to structure their reasonings/explanations around four key elements, while recognizing that these components may naturally overlap in expert reasoning.

1. **Analyze** any experimental data obtained thus far
2. Form **hypotheses** about the most critical factors achieving high performance based on both observed data and domain knowledge, drawing on fundamental principles
3. Provide explicit **rationale** for your next experimental suggestion(s)
4. **Recommend** a batch of suggestions to test

These guidelines are intended to capture the depth and systematic nature of expert chemical reasoning rather than constrain creative problem-solving — overlapping discussions of data trends and chemical hypotheses, or integrated rationale that spans multiple elements, are expected and valuable for understanding how human experts navigate complex optimization landscapes.

## Results

First we benchmarked our LLM-GO approach on the five single objective chemical reaction datasets and a single multi-objective dataset shown in **Figure 1**, using both BO and random sampling as baseline methods. The random sampling baseline is shown in the Supporting Information (**Section SJ**). Importantly, **Figures 2 and 3** provides insights into the optimization landscape and complexity. The datasets span a range of difficulty levels, with the Suzuki coupling and Amide coupling datasets exhibiting relatively favorable distributions of outcomes (median ~32-36%), while the Chan-Lam coupling and reductive amination datasets presents more challenging optimization problems with lower typical objective values (median ~11-13%). Beyond the objective value distributions shown, these datasets vary in the six metrics presented in **Figure 3**. Overall, the Buchwald-Hartwig dataset is deemed the most complex. All datasets are fully mapped with no continuous parameters. This allows for objective values to be retrieved from a lookup table, rather than relying on a regression model to predict objective values.

We evaluated LLMs from three different providers: Anthropic, Google, and OpenAI. Results for six Anthropic models, six Google models, and six OpenAI models are presented in **Figure 5**. For each provider, we evaluate at least one model trained to reason/think before responding. Details on the LLM configurations used are presented in the methods section. BO was evaluated via six separate methods spanning three acquisition functions, with and without descriptors. The generated optimization trajectories can be found on HuggingFace (https://huggingface.co/datasets/gomesgroup/iron-mind-data).

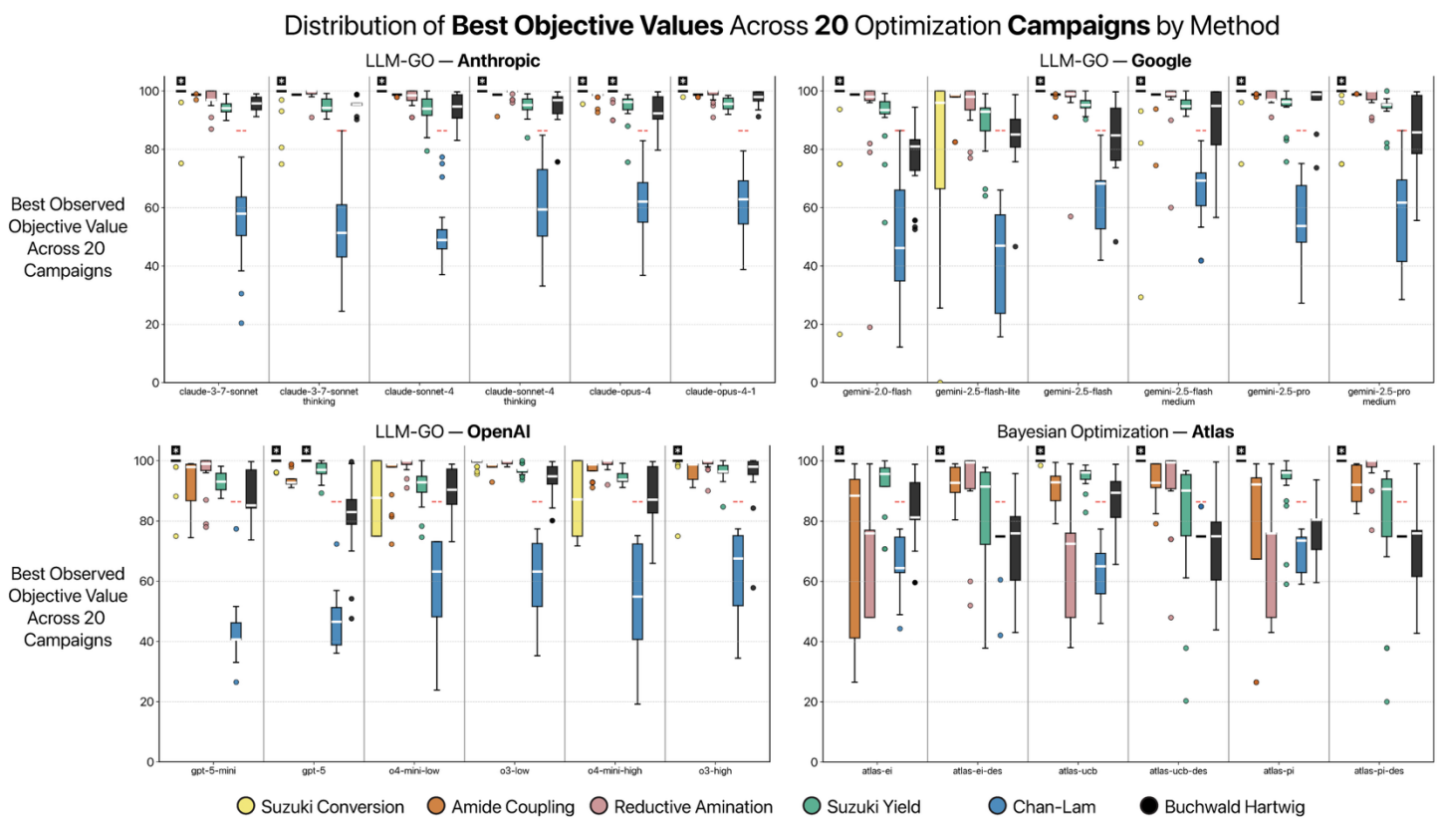

**Figure 5 — Method performance across optimization datasets.** Each panel shows a different optimization provider: three LLM-based providers (Anthropic, Google, OpenAI) and one Bayesian optimization provider (Atlas). Boxplots (**n=20**) display the distribution of best objective values achieved across 20 independent optimization campaigns for each method-dataset combination. Colors are used to represent the six different optimization datasets. Asterisks (*) indicate methods where the median equals the maximum objective value and the interquartile range is zero. Objective values are the raw values from each dataset's objective function. Individual points beyond the whiskers indicate outliers in the performance distribution. A dashed red line is shown for the Chan-Lam dataset to indicate the best possible performance, all other datasets have best possible performance values of 100 (percent yield/conversion). Random baseline results and stats in **Figure S12-S13**.

To provide comprehensive statistical validation of the performance differences observed in **Figure 5**, we conducted pairwise statistical comparisons among all LLM-GO methods, BO methods, and the random baseline for each dataset. We applied the Wilcoxon[57] rank-sum test to assess statistical significance and calculated Cliff's delta[58] ($\delta$) to quantify effect sizes for each method pair. The Wilcoxon test determines whether two methods have significantly different performance distributions ($p < 0.05$), while Cliff's delta measures the magnitude and direction of performance differences, with positive values indicating superior performance relative to the comparison method and negative values indicating inferior performance. Additionally, we computed 95% bootstrap confidence intervals[59] for all median performance values to quantify the precision of our median estimates (1,000 bootstrap samples per method-dataset combinations). Complete pairwise matrices comparing all LLM-GO methods to each BO method are shown in **Figures S1-2**. Corresponding information comparing all optimization methods to a random sampling baseline is shown in **Figure S13**. For all methods, including the random baseline, bootstrap confidence intervals are provided in **Figure S3**.

The system prompt used for LLM optimization instructed the models to never suggest a previously suggested set of parameters. However, in developing LLM-GO, this criterion was not strictly enforced, and as is the nature of LLMs this criterion cannot be guaranteed via one-time prompting in the system prompt. This presents a test for LLMs in-context retrieval capabilities, akin to "needle in a haystack" evaluations.[60] We find that several models are proficient in adhering to this criteria, while others struggle, often hindering optimization performance. In contrast, BO explicitly adheres to this criterion. In **Figure 6**, we show the distribution of repeated parameter suggestions for each method across all 20 campaigns on a given dataset.

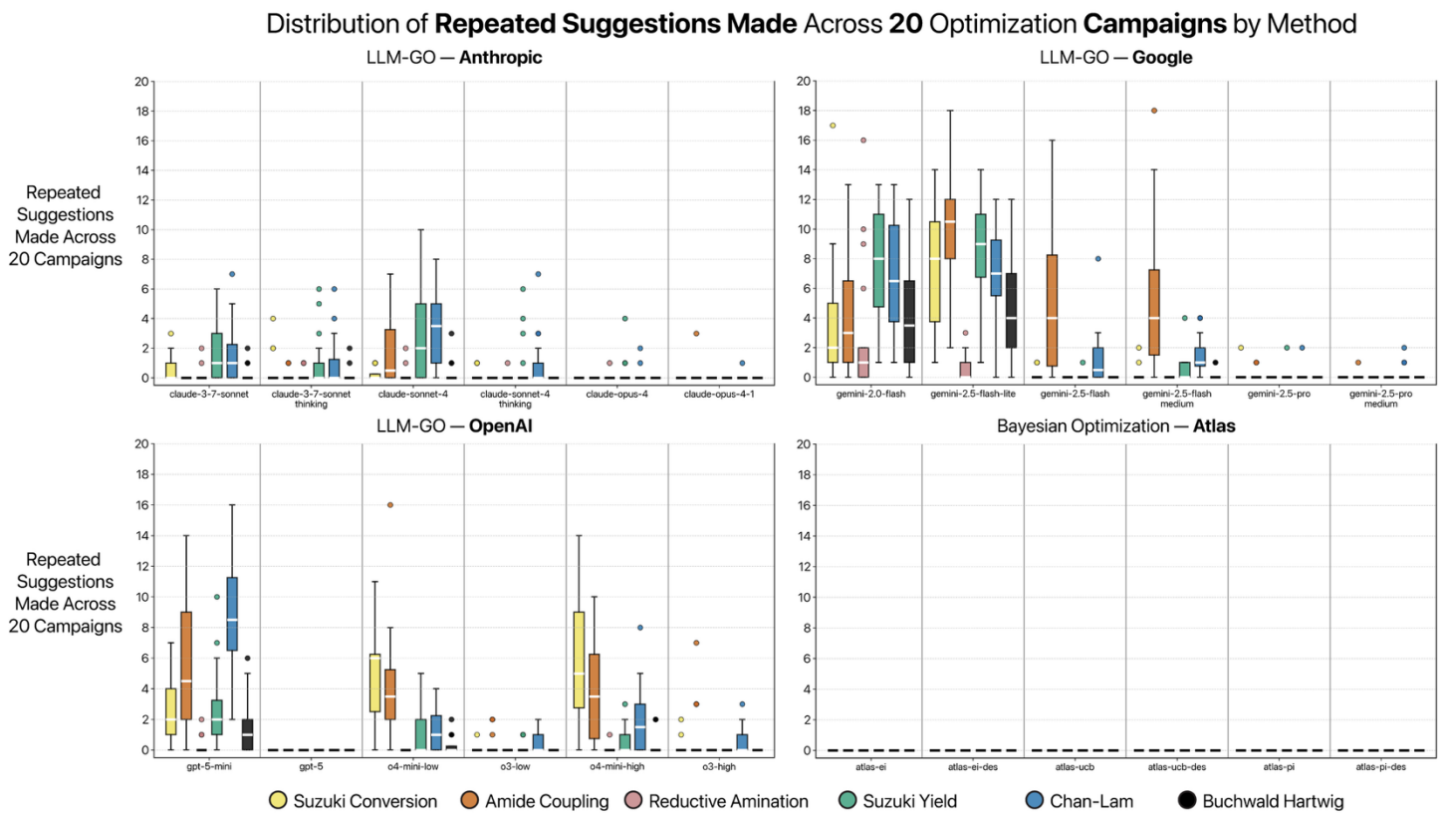

**Figure 6 — Distribution of repeated suggestions across optimization campaigns by method and dataset.** Within each campaign, for each method-dataset combination, we measure the number of parameter configurations that were suggested multiple times across 20 independent optimization campaigns. Boxplots show the distribution of repeated suggestions across campaigns, where each datapoint represents a campaign, with each color representing a different dataset. Methods are grouped by provider (LLM-GO for Anthropic, Google, and OpenAI; Bayesian Optimization for Atlas). Lower values indicate more efficient exploration, as the method avoids redundant suggestions. The median is shown as a white line within each box, except where the median is 0 and the interquartile range is 0 (shown as a black line). Outliers are displayed as individual points.

To understand how different optimization methods navigate parameter selection strategies, we analyzed parameter selection diversity using cumulative Shannon entropy across entire optimization campaigns. This analysis quantifies how broadly or narrowly methods sample the parameter space throughout their complete runs, providing insights into their overall search strategies. Methods with high cumulative entropy explore broadly across parameter options throughout the campaign, while those with low entropy focus on specific parameter combinations.

**Figure 7** shows the distribution of cumulative entropy values across 20 independent campaigns for each method and dataset. These distributions reveal distinct strategic approaches: some methods consistently maintain broad parameter exploration throughout their campaigns, while others focus on exploiting specific parameter combinations. The entropy distributions help explain performance differences and highlight fundamental strategic differences between LLM-based and Bayesian optimization approaches.

We conducted similar pairwise statistical comparisons for the entropy distributions across runs, focusing on comparisons between optimization methods to characterize differences in sampling strategies. Complete pairwise entropy comparison matrices are provided in the Supporting Information (**Figure S6-8**).

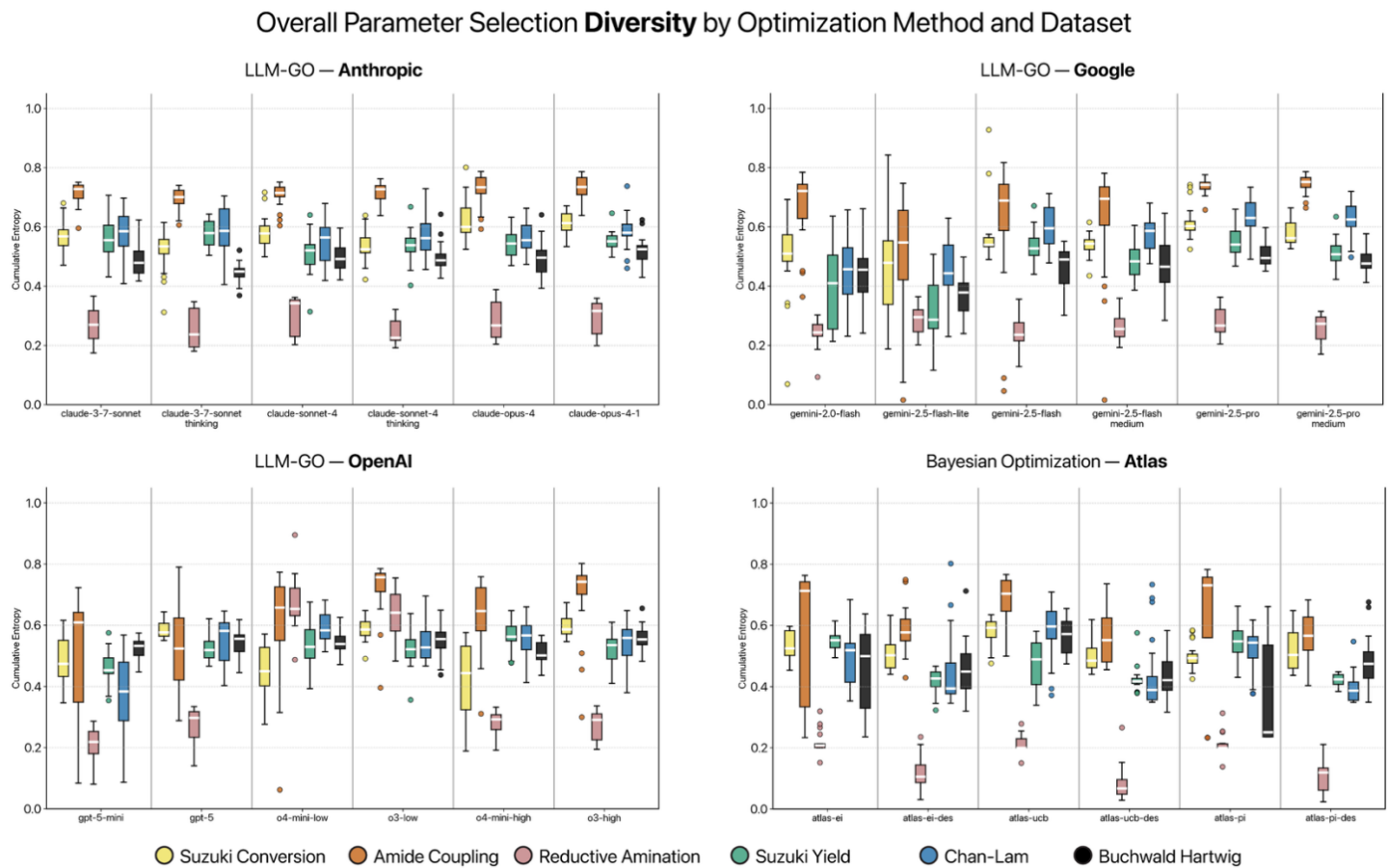

**Figure 7 — Cumulative parameter selection Shannon entropy by optimization method and dataset.** Parameter selection diversity is measured using normalized Shannon entropy averaged across all parameters for each method-dataset combination over complete 20-iteration optimization runs (see Methods for mathematical formulation). Each colored boxplot represents a different dataset, with methods grouped by provider (LLM-GO for Anthropic, Google, and OpenAI; Bayesian Optimization for Atlas). The y-axis ranges from 0 (focused exploitation of specific parameter options) to 1 (uniform exploration across all parameter options). Boxplots show the distribution of cumulative entropy across 20 independent runs, revealing each method's overall exploration strategy: higher entropy indicates broader parameter space exploration throughout the optimization process, while lower entropy shows more focused exploitation of specific parameter regions.

# Discussion

## Optimization method performance and limitations

The performance shown in **Figures 5** and **6** reveals strengths and limitations of the evaluated optimization methods. Our systematic evaluation across six chemical reaction datasets demonstrates that optimization effectiveness is intimately tied to the underlying characteristics of the chemical parameter space. Performance patterns across datasets with varying literature precedent and solution scarcity provide insights into how LLMs leverage pre-trained chemical knowledge versus training data memorization.

The random sampling baseline comparisons (**Figure S13**) serve as a critical benchmark for evaluating the practical value of sophisticated optimization methods across our chemical reaction datasets. Given the known characteristics of these parameter spaces — particularly their favorable solution densities as captured in our complexity analysis and objective value distribution visualization (**Figures 2 and 3**) — random sampling predictably achieves high performance across all datasets. This establishes a stringent practical threshold: optimization methods must demonstrably exceed random performance to justify their computational complexity and implementation effort. More importantly, the random baseline enables us to distinguish genuine parameter space navigation from coincidental success due to abundant good solutions. Suzuki conversion, with the lowest scarcity index, shows saturated performance where random sampling achieves near-optimal results (~100% median, **Figure S12**), making meaningful distinctions between methods difficult. In contrast, five datasets exhibit high scarcity indices where good solutions are sparse: Amide Coupling, Suzuki Yield, Chan-Lam, Buchwald-Hartwig, and Reductive Amination. Within this high-scarcity group, Reductive Amination represents the least scares case, still allowing relatively strong random performance. As scarcity intensifies across the remaining datasets, random sampling performance deteriorates progressively. Across these high-scarcity datasets, LLM and BO methods show advantages over random sampling (**Figure S13**), indicated by effect sizes. This progression confirms that both BO and LLM-GO represent structured parameter space navigation strategies distinct from random exploration, validating meaningful method-to-method comparisons across all datasets regardless of baseline performance levels.

BO performance varies substantially across datasets, with strong results on some datasets (Suzuki Conversion, Suzuki Yield) while showing greater variability on others (Reductive Amination, Buchwald-Hartwig). The statistical significance heatmaps (**Figures S1 and S2**) demonstrate that BO methods achieve statistically significant advantages over many LLM methods on certain datasets, with effect sizes showing medium to large positive effects in these cases. The effect sizes are most pronounced in cases where molecular descriptors are included and result in improved performance over the OHE case. However, molecular descriptor inclusion shows highly problem-dependent effects on BO performance, underscoring that descriptor selection is not a one-size-fits-all process. Descriptors provide clear improvements for Amide Coupling — likely because representing the complex "reaction conditions" categorical parameter as independent molecular descriptors for each chemical component better captures relevant information than treating conditions as monolithic categories. However, descriptors deteriorate performance on Suzuki Yield and Buchwald-Hartwig, while showing minimal impact on other datasets. These inconsistent patterns highlight a fundamental challenge of BO-based optimization: effective implementation requires problem-specific feature engineering decisions that demand domain expertise. This contrasts with LLM-based methods that demonstrate more consistent performance across diverse chemical optimization landscapes without requiring explicit feature selection.

Some LLMs demonstrate remarkably consistent performance across all datasets regardless of complexity metrics. The most striking finding is the exceptional robustness of Anthropic models beginning with claude-3-7-sonnet, achieving consistently high performance with minimal variability between campaigns (**Figure 5**). Statistical comparisons reveal that Anthropic and OpenAI models significantly outperform BO methods on complex datasets like Buchwald-Hartwig and Reductive Amination ($p < 0.001$, **Figure S1**), with large effect sizes (Cliff's delta > 0.5, **Figure S2**). Google models show similar robustness with slightly more variability while gemini-2.5-pro matches the best Anthropic/OpenAI performance across most datasets. Notably, LLM performance remains strong even on Amide Coupling — a recently published dataset unlikely to appear in training corpora. This suggests LLMs successfully leverage generalizable chemical knowledge rather than memorizing specific datasets.

However, performance becomes more variable on the Chan-Lam dataset; this dataset presents unique challenges: it is the only multi-objective benchmark, contains multiple measurements per parameter set (~3 replicates), and explores under-studied chemistry. Performance assessment requires aggregating multiple measurements into a single objective value — a decision that profoundly impacts apparent method performance (**Figure S14A**). Using average aggregation, LLMs achieve comparable performance to BO in some cases, while upper-bound aggregation indicates superior LLM performance in a number of cases. This sensitivity suggests a potential LLM reasoning pitfall: models may overweight individual high-performing measurements rather than properly integrating evidence across replicate observations showing conflicting outcomes.

To address both the under-studied chemistry and multi-measurement challenges, we evaluated LLM performance when provided with the complete Chan-Lam paper in an initial user prompt, using three aggregation strategies (**Figure S14B**). The lower-bound aggregation — which most conservatively penalizes inconsistent replicates — provides the most stringent performance metric. With this approach and paper context, several LLMs show substantial improvement, with gemini-2.5-flash now demonstrating positive effect sizes versus BO methods (**Figure S14D**), indicating statistically superior performance. Claude-sonnet-4 and gemini-2.0-flash similarly improve with this intervention.

These results clarify that Chan-Lam's challenge stems from the combination of under-studied chemistry, multi-objective optimization, and multi-measurement variability rather than simply the absence from the training corpora. The strong LLM performance on Amide Coupling reinforces that LLMs can effectively leverage pre-trained chemical knowledge to navigate genuinely novel substrate-condition combinations. However, the aggregation sensitivity reveals that LLMs may struggle to appropriately weight conflicting experimental evidence, representing an area where structured probabilistic approaches like BO maintain advantages through explicit uncertainty quantification.

The duplicate suggestion analysis (**Figure 6**) reveals a critical limitation directly impacting LLM optimizer effectiveness. Models that frequently suggest duplicate experiments demonstrate a cascading failure pattern: rapid convergence to suboptimal solutions coupled with poor final performance. This reflects a breakdown where models fail to maintain experimental memory, leading to inefficient parameter space exploration and wasted experimental budget, which is particularly problematic in laboratory settings where each experiment consumes valuable resources. However, this limitation can be addressed through improved LLM planner design implementing explicit duplicate checks, structured memory systems for previously tested combinations, and dynamic prompting strategies that increasingly emphasize exploration of untested parameter space as campaigns progress.

Statistical analysis of Shannon entropy patterns (**Figure S6-S7**) reveals that LLM methods engage in fundamentally different sampling strategies compared to BO methods. Across most datasets and LLM methods, we observe statistically significant differences in entropy distributions (extensive blue regions in **Figure S6**, $p < 0.05$), with effect sizes (**Figure S7**) predominately showing positive values. This indicates that LLM methods consistently employ more exploratory sampling strategies than BO methods. The entropy analysis (**Figure 7**) demonstrates that sampling strategies are strongly dataset-dependent, with all methods showing similar patterns: highest entropy on Amide Coupling, lowest on Reductive Amination, and intermediate values on other datasets. However, within these dataset-specific ranges, LLM methods maintain systematically higher entropy than BO methods, as confirmed by the predominately positive effect sizes in **Figure S7**. The statistical comparisons reveal that the exploratory bias of LLM methods relative to BO is largely consistent across Anthropic, Google, and OpenAI models, suggesting this represents a fundamental characteristic of LLM-GO rather than provider-specific implementation differences.

Several important limitations constrain the generalizability of these findings. First, LLM performance likely depends on the chemical representation formats (i.e. SMILES strings vs. common names) and system prompt formulations used in this study, potentially limiting generalizability across different representation schemes. Second, our complexity metrics and entropy analysis framework applies specifically to fully mapped categorical/discrete parameter spaces, restricting applicability to continuous optimization problems or mixed parameter spaces. Finally, we cannot provide definitive explanations as to why certain LLMs outperform others, as performance differences do not correlate cleanly with model size, release, date, or other publicly available characteristics, and we lack access to training data compositions that might explain these performance variations.

## Implementation Strategies and Future Directions

While our results demonstrate the potential of LLM-GO, several considerations affect real-world adoption. We focused on predominantly categorical parameter spaces where BO often struggles. Furthermore, our Shannon entropy analysis demonstrates that BO methods are forced into suboptimal exploration patterns on high-complexity datasets even when including descriptors. This positions LLMs as particularly valuable for chemical optimization problems. Our integration with Olympus, an optimization algorithm benchmarking platform, enables seamless handling of categorical, discrete, and continuous parameters, providing a clear pathway for practitioners to define parameter spaces that LLMs can reason over and make recommendations within.

One potential barrier to widespread adoption is cost (see **Table S2**): LLM API calls are significantly more expensive than the near-zero computational cost of BO, particularly for extended experimental campaigns. However, the optimizer cost is likely to pale in comparison to the cost of running experiments, in which case, greater optimizer cost is justified in the name of increased performance. Moreover, BO has achieved broad industrial adoption, especially in pharmaceutical and materials research, suggesting that demonstrated value can overcome cost considerations. We designed our approach to encourage similar adoption through accessible tooling and demonstrated performance advantages on challenging optimization landscapes.

Several strategies can mitigate cost concerns while maintaining optimization quality. Hybrid approaches could leverage high-quality LLMs for reasoning at each iteration while using cheaper, even local, models for parameter recommendations, or employ premium models only every $N$ iterations. The substantial reasoning data generated from our campaigns (400 reasoning traces per method per dataset) provides opportunities to fine-tune open-source models, potentially achieving competitive performance using local resources rather than external APIs. Additionally, improved prompting strategies — such as explicit duplicate prevention checks and dynamic budget awareness that informs LLM methodology as experiments progress — could enhance efficiency. Unlike BO, which relies on implicit exploration-exploitation balancing, LLMs can explicitly reason about remaining experiment budget to adjust their recommendation strategies accordingly.

A particularly promising direction involves integration into agentic systems capable of performing computational tasks and gathering molecular descriptors that dynamically empower the LLM optimizer. Descriptor effectiveness across the benchmarks has been demonstrated to depend critically on problem complexity and parameter structure — an agentic LLM system could analyze different molecular features in response to experimental findings. This represents a fundamental shift toward adaptive computation-experiment integration, moving beyond the current paradigm where computational analysis either precedes experimental batches or follows large experimental datasets in isolated cycles and the static descriptor paradigm that limits BO performance. An agentic LLM system could dynamically decide which computational tools to employ (whether calculating electronic properties, predicting solubility, analyzing steric effects, etc.) based on emerging experimental patterns, creating a truly responsive optimization framework that mirrors how expert chemists iteratively refine their understanding throughout experimental campaigns.

## Validation and Community Engagement

Establishing trustworthiness in LLM-GO requires validation across multiple dimensions. While our benchmarking results demonstrate impressive performance across diverse reaction types and complexity levels, these evaluations rely on retrospective analysis of literature datasets where optimal outcomes are already established. Prospective validation in real-world optimization campaigns with unknown optima represents a crucial next step for establishing practical utility. Additionally, performance alone is insufficient for building confidence in AI-driven experimental design. The results suggest that LLMs perform knowledge application rather than traditional optimization patterns, making it crucial to validate the reasoning processes underlying these decisions.

To address these validation challenges, we call upon the scientific community to engage with our web-based platform to provide human expert optimization strategies that enable systematic comparison of reasoning trajectories between human and LLM approaches. By gathering human reasoning data, we can systematically compare rationale, parameter selection patterns, and hypothesis-outcome relationships using text embeddings and semantic similarity analysis to assess consistency between human and LLM decision-making processes. This analysis will reveal whether LLMs arrive at similar experimental choices through reasoning pathways that align with expert chemical intuition, or whether their success stems from alternative decision-making processes that, while effective, may be less trustworthy. Such validation will establish a foundation for responsible integration of LLM guidance in experimental chemistry, ensuring that these systems not only perform well but do so through chemically sound reasoning that experts can interpret, critique, and build upon.

To illustrate the type of human reasoning data we seek, we provide example human optimization campaigns in the supporting information (**Section SL**) for an N-alkylation/deprotection dataset. While this dataset was evaluated in an earlier version[61] of this work and is not included in our primary benchmark comparisons, it serves as a valuable demonstration of expert and non-expert decision-making patterns. Each batch (batch size = 1) outlines a specific hypothesis regarding the electrophile, core, or base, based on chemical principles such as steric hindrance, partial charges, and pKa values. This cognitive approach builds a knowledge base that guides future experiments and leads to high yields (100% in batch 5) through iterative refinement based on prior observations. This sample campaign, while primarily a demonstration, clearly illustrates how human reasoning provides data to, first, further evaluate LLM reasoning and potentially enhance AI model sophistication. Complete dataset details and reaction scheme are available in **Section SL**. Corresponding LLM and BO optimization results are made available on our web-based platform (https://gomes.andrew.cmu.edu/iron-mind).

# Conclusion

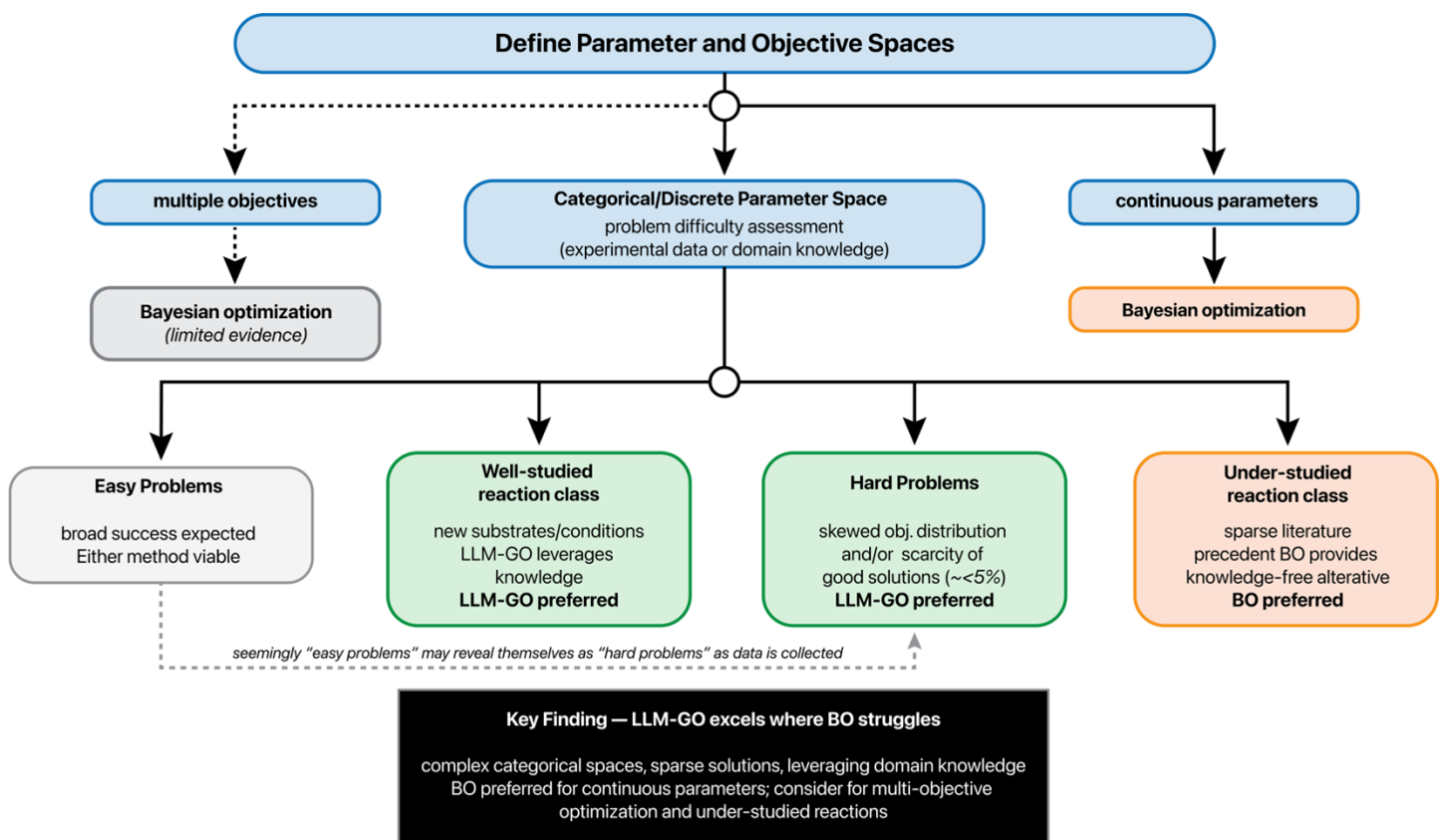

**Figure 8 — Decision framework for optimization method selection in reaction optimization campaigns.** Practitioners begin by defining their parameter and objective spaces, then follow decision pathways based on problem characteristics. For categorical/discrete parameter spaces, problem difficulty can be assessed through initial experiments or domain expertise. LLM-GO is preferred for well-studied reaction classes and hard problems with sparse solutions and/or skewed objective distributions. For under-studied reaction classes with limited literature precedent, BO provides a knowledge-free alternative, though LLMs may be viable when provided with domain context (e.g., key papers or procedures). BO is preferred for continuous parameter spaces; for multi-objective optimization, consider BO though evidence is limited to a single dataset in this study. Problem difficulty may reveal itself as data accumulates — seemingly easy problems can transition to hard problems as scarcity of high-performing conditions becomes apparent.

Our systematic benchmarking reveals that LLMs excel on complex reaction optimization problems where traditional BO struggles. We present a decision framework for practitioners in **Figure 8**. While BO performs well on simple parameter landscapes and multi-objective scenarios, LLMs demonstrate superior performance and remarkable consistency across challenging single-objective datasets. Our entropy analysis reveals that LLMs maintain a consistent exploratory bias relative to BO methods across diverse datasets, enabling effective parameter selection without the dataset-dependent strategy constraints that limit BO performance. This fundamental difference, combined with LLMs' robust handling of categorical parameters, positions LLM-GO as a powerful tool for reaction optimization.

However, transitioning from promising research tool to trusted laboratory infrastructure requires community validation. We call upon the community to engage with our platform to not only to benefit from AI-guided optimization, but to validate whether LLM reasoning aligns with expert chemical intuition. This collaborative effort will establish the foundation for responsible AI integration in experimental chemistry, ensuring these powerful capabilities serve the field with appropriate understanding of their trustworthiness and limitations.

## Data Availability
The Iron Mind web application can be found at: https://gomes.andrew.cmu.edu/iron-mind/. Code for generation of all figures in this manuscript can be found at GitHub: https://github.com/gomesgroup/iron-mind-public. Generated optimization trajectories can be found on HuggingFace: https://huggingface.co/datasets/gomesgroup/iron-mind-data. If you use this dataset, please cite its Zenodo repository DOI: 10.5281/zenodo.16762946.

## Author Contributions
R.M. and G.G. co-developed the main concepts of this project. R.M. developed methods for LLM-GO, complexity analysis, and ran all method evaluations. R.M. and J.E.R developed the entropy analysis and implemented the BO strategies using Atlas. R.M. developed the no code web application. L.A.B. and J.G.E. conducted example human optimization campaigns and contributed with thorough suggestions throughout the development of this project. G.G. coordinated the project direction and secured funding. All authors participated in the writing of this manuscript.

## Acknowledgements
The authors thank Daniil Boiko (CMU Chemical Engineering, now at Onepot AI) for the many discussions about reaction optimization.

We gratefully acknowledge the partial financial support by the National Science Foundation Center for Computer-Assisted Synthesis (NSF C-CAS, grant no. 2202693). This work was partially supported by the Air Force Office of Scientific Research (AFOSR) and the Air Force Research Laboratory (AFRL) Materials and Manufacturing Directorate through the Data-Driven Discovery Of Optimized Multifunctional Material Systems (D3OM2S) Center of Excellence under award number FA8650-19-2-5209. R.M. thanks CMU and its benefactors for the 2024–25 Tata Consulting Services Presidential Fellowship. J.E.R. thanks NSF and C-CAS for the 2024–25 DARE Fellowship.

We thank Anthropic, PBC (anthropic.com) for their support in API credits for our efforts on autonomous chemical research.

Any opinions, findings, and conclusions or recommendations expressed in this material are those of the authors and do not necessarily reflect the views of the National Science Foundation, Air Force Office of Scientific Research, Air Force Research Laboratory, Anthropic or any of the funding entities.

## Competing Interests
R.M. and G.G. are co-founders of $e^{vals}$, a consultancy firm for scientific evaluations of frontier AI models. The other authors declare no competing interests.

# Pre-trained knowledge elevates large language models beyond traditional chemical reaction optimizers

Robert MacKnight,[1] Jose Emilio Regio,[2] Jeffrey G. Ethier,[3] Luke A. Baldwin,[3] Gabe Gomes*[1,2,4,5]

1. Department of Chemical Engineering, Carnegie Mellon University, Pittsburgh, PA 15213, USA
2. Department of Chemistry, Carnegie Mellon University, Pittsburgh, PA 15213, USA
3. Materials and Manufacturing Directorate, Air Force Research Laboratory, Wright-Patterson AFB, OH 45433, USA
4. Machine Learning Department, Carnegie Mellon University, Pittsburgh, PA 15213, USA
5. Wilton E. Scott Institute for Energy Innovation, Carnegie Mellon University, Pittsburgh, PA 15213, USA

* corresponding author: gabegomes@cmu.edu

# Supporting Information

A. Statistical Significance and Effect Size Analysis for Method Performance

Statistical Significance (**p-values**) for LLM vs. BO Method Comparisons

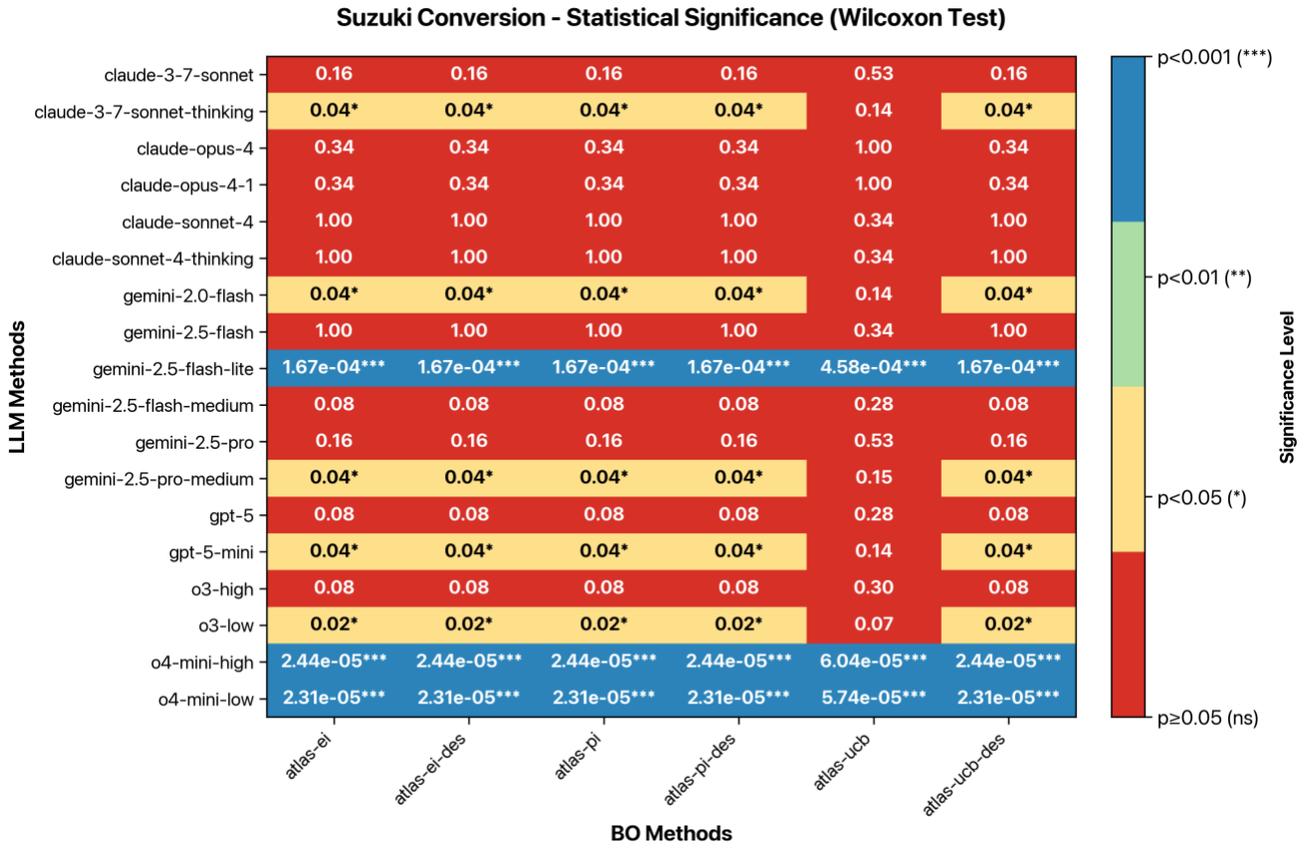

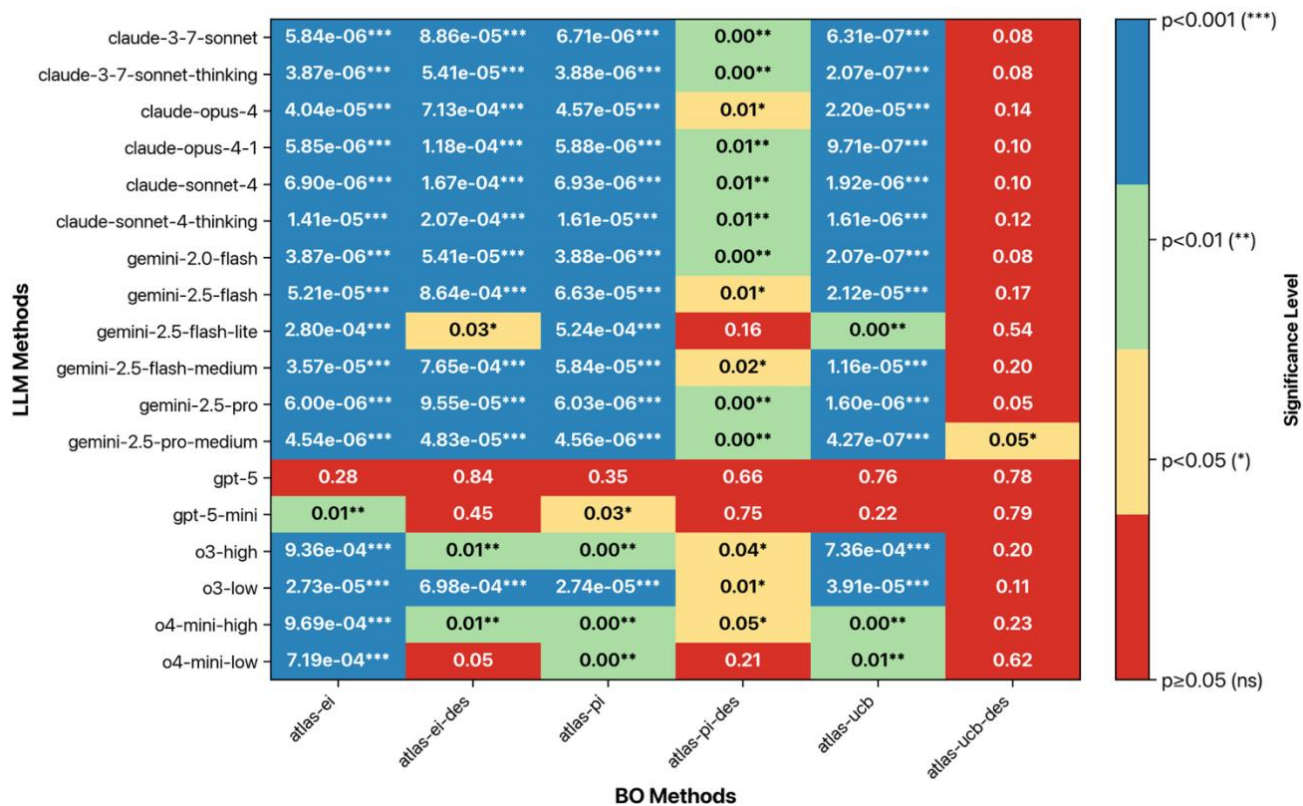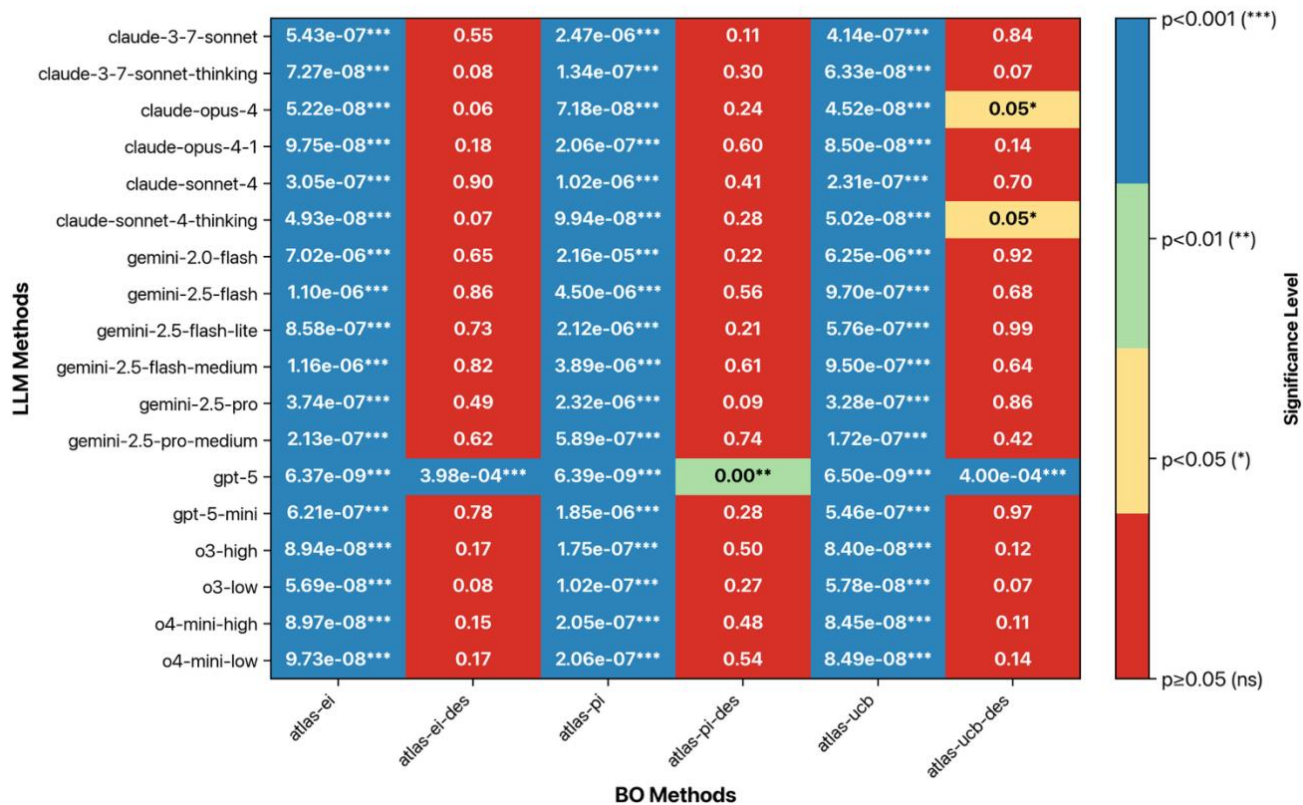

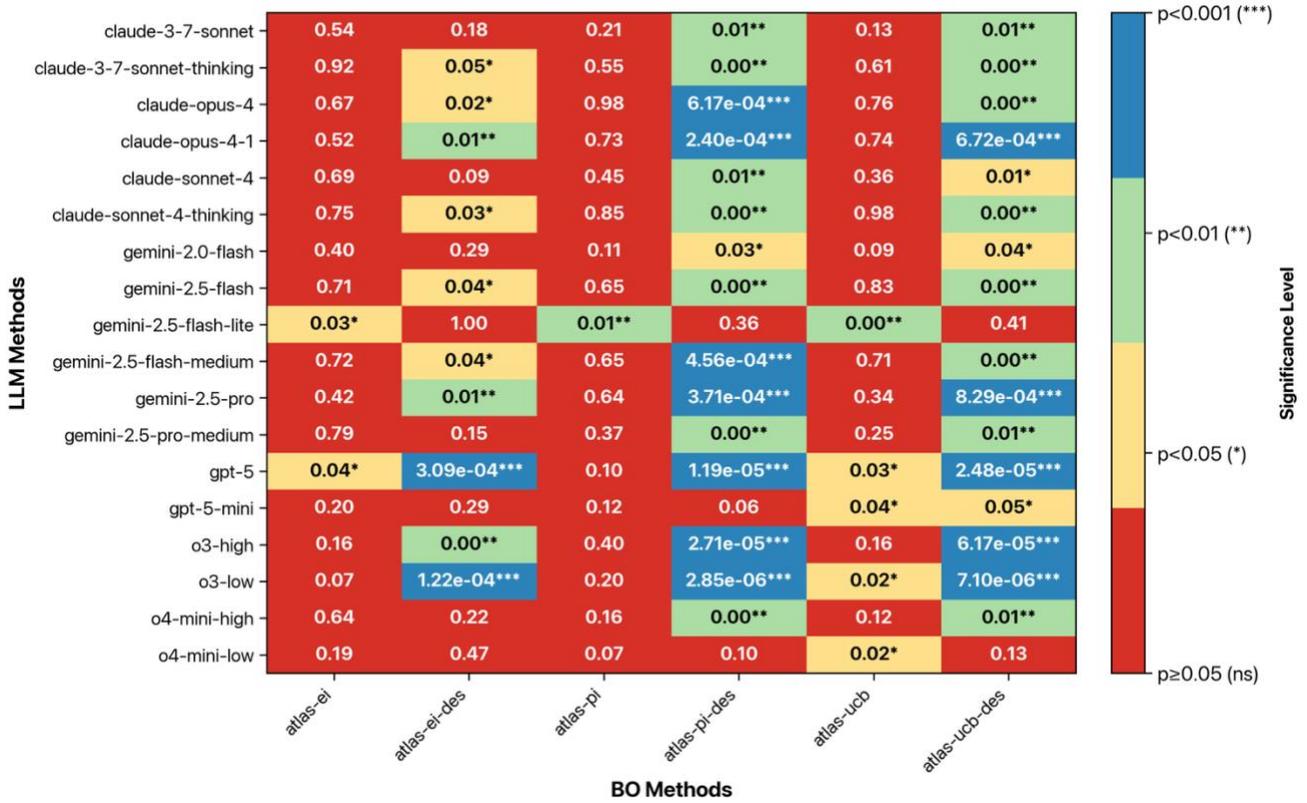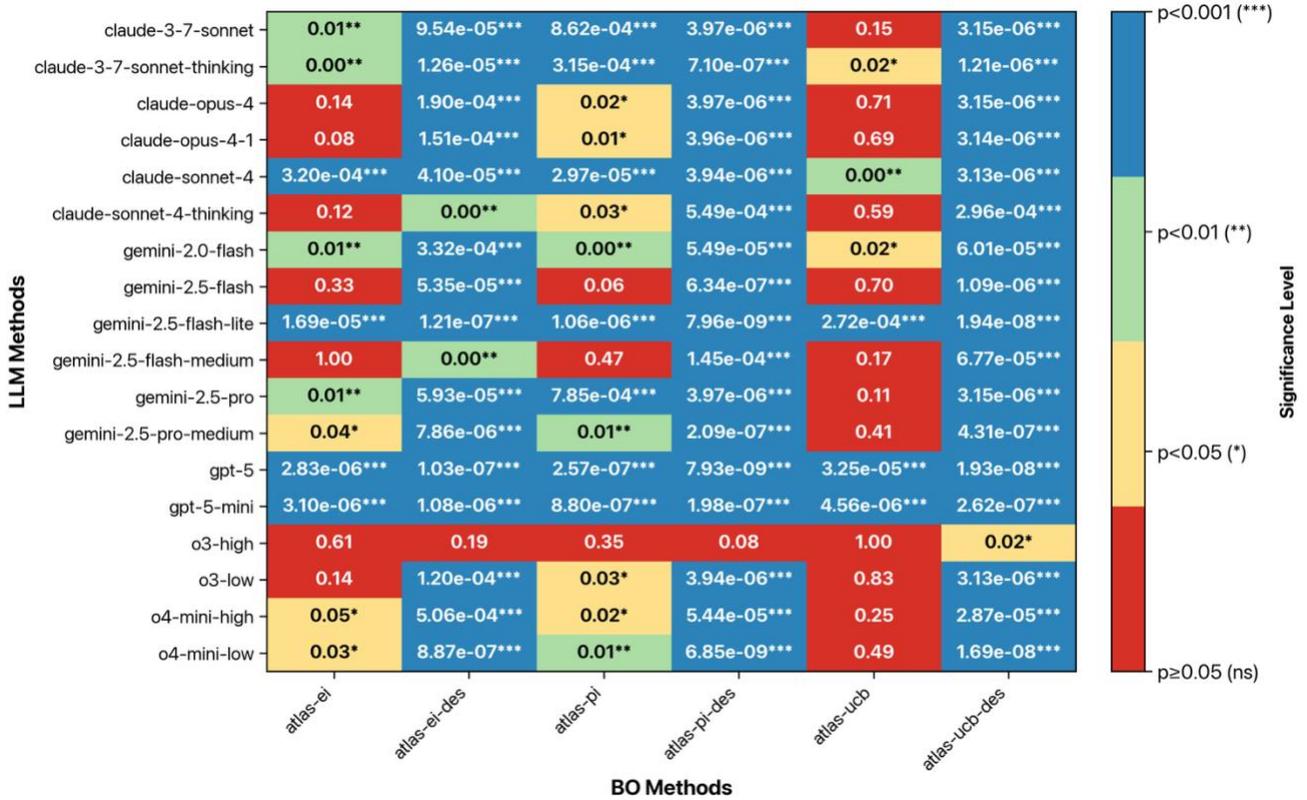

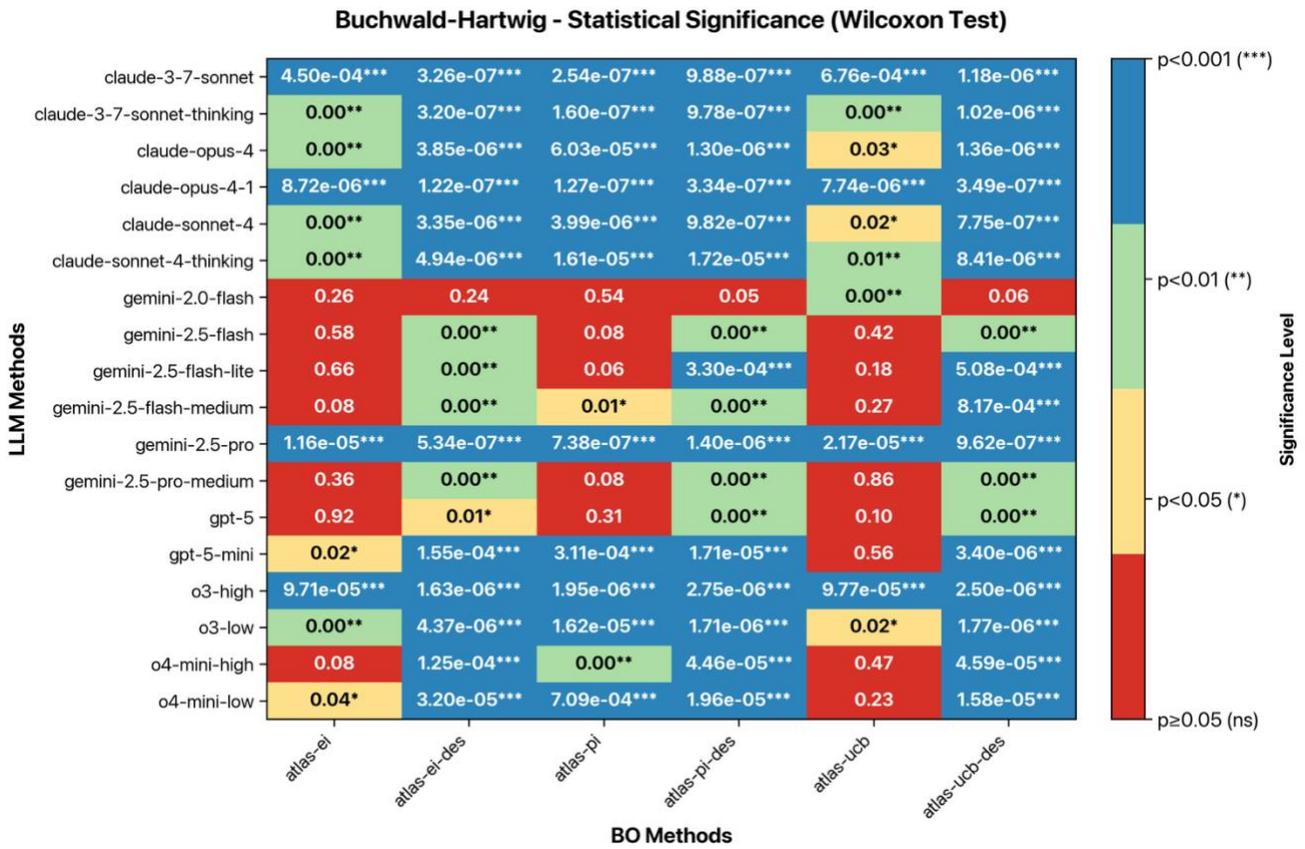

**Figure S1 — Statistical significance tests for method performance.** For each dataset, we conduct Wilcoxon tests to compare the performance distributions of all LLM-GO methods with all BO methods. Red indicates the performance distribution is not significantly different. All other colors indicate a varying level of significance, indicated by the color bar on the right.

# Effect Sizes (**Cliff's Delta**) for LLM-GO vs. BO Method Comparisons

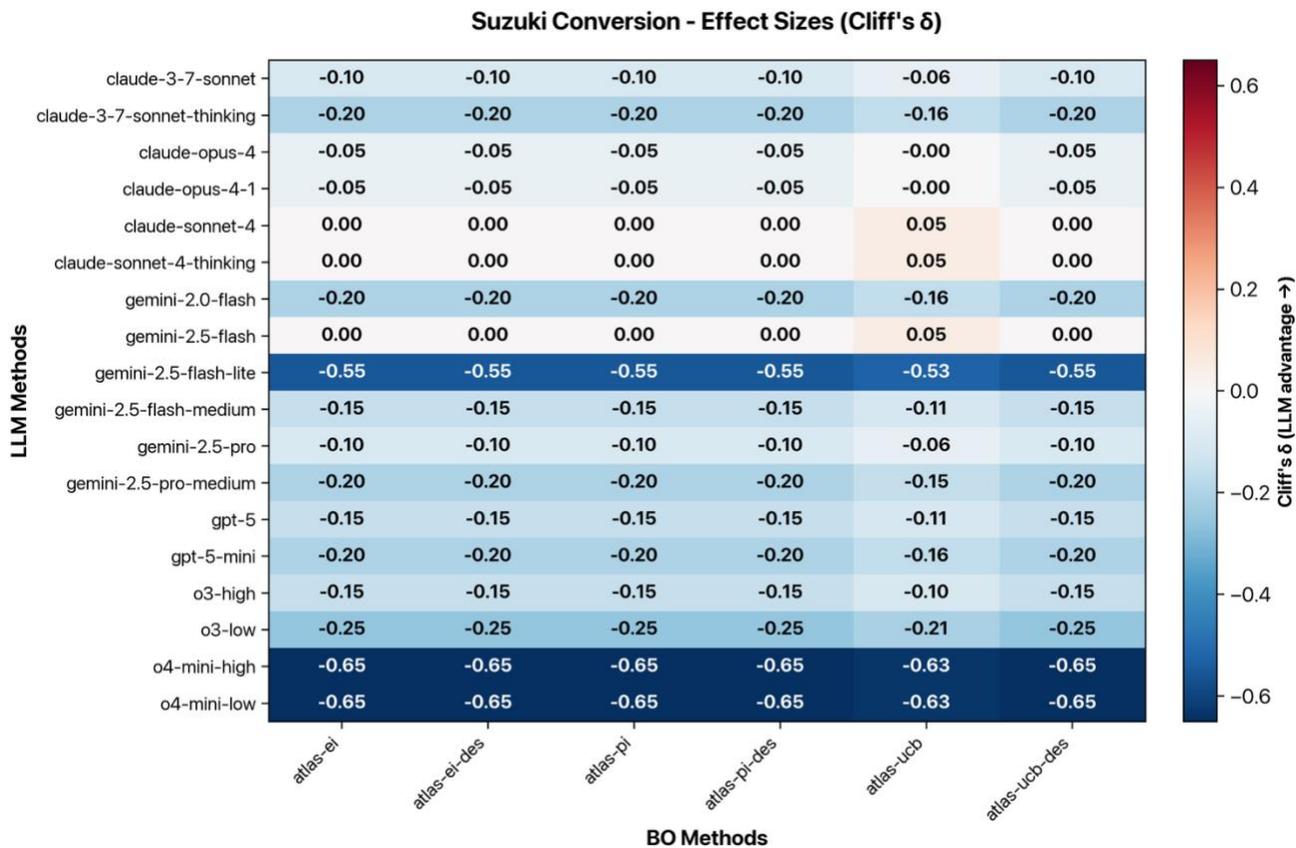

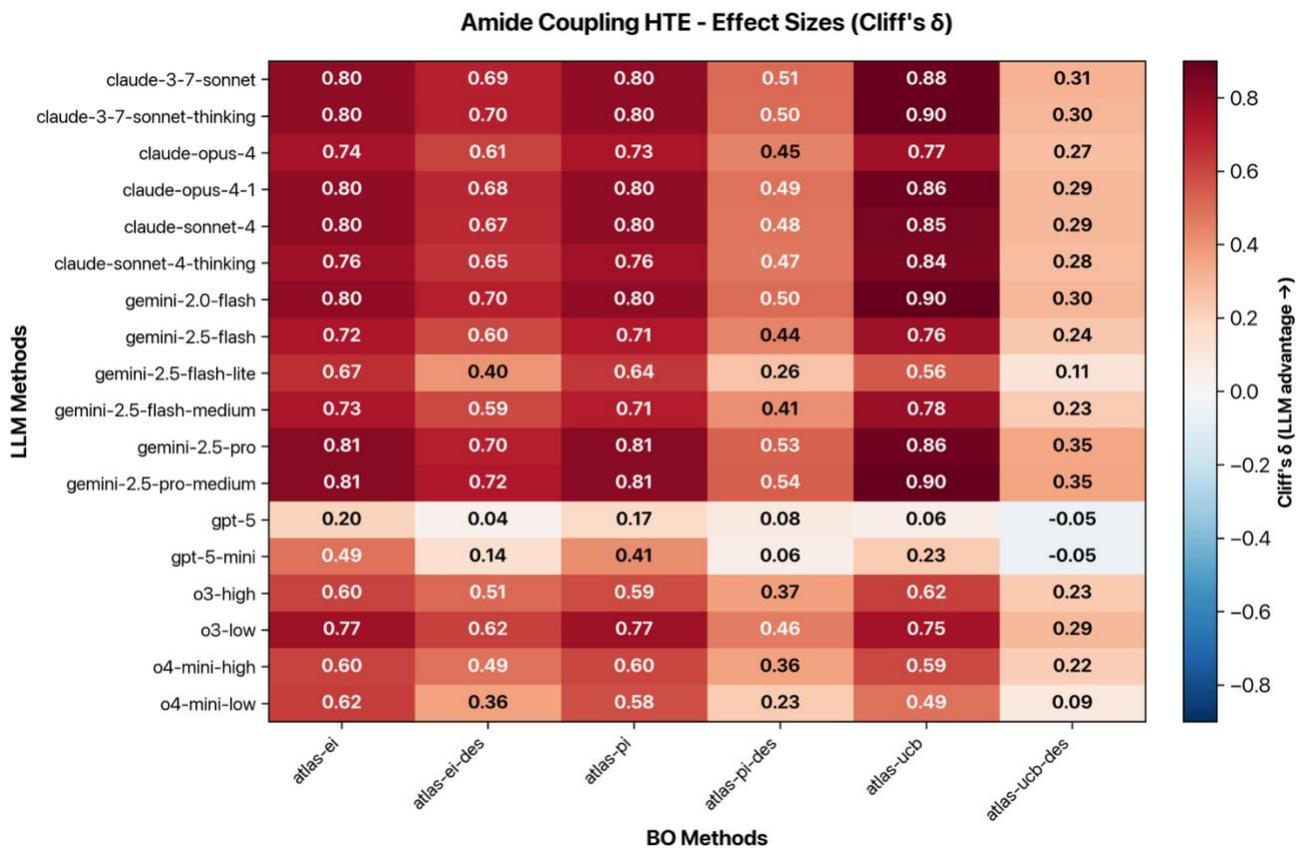

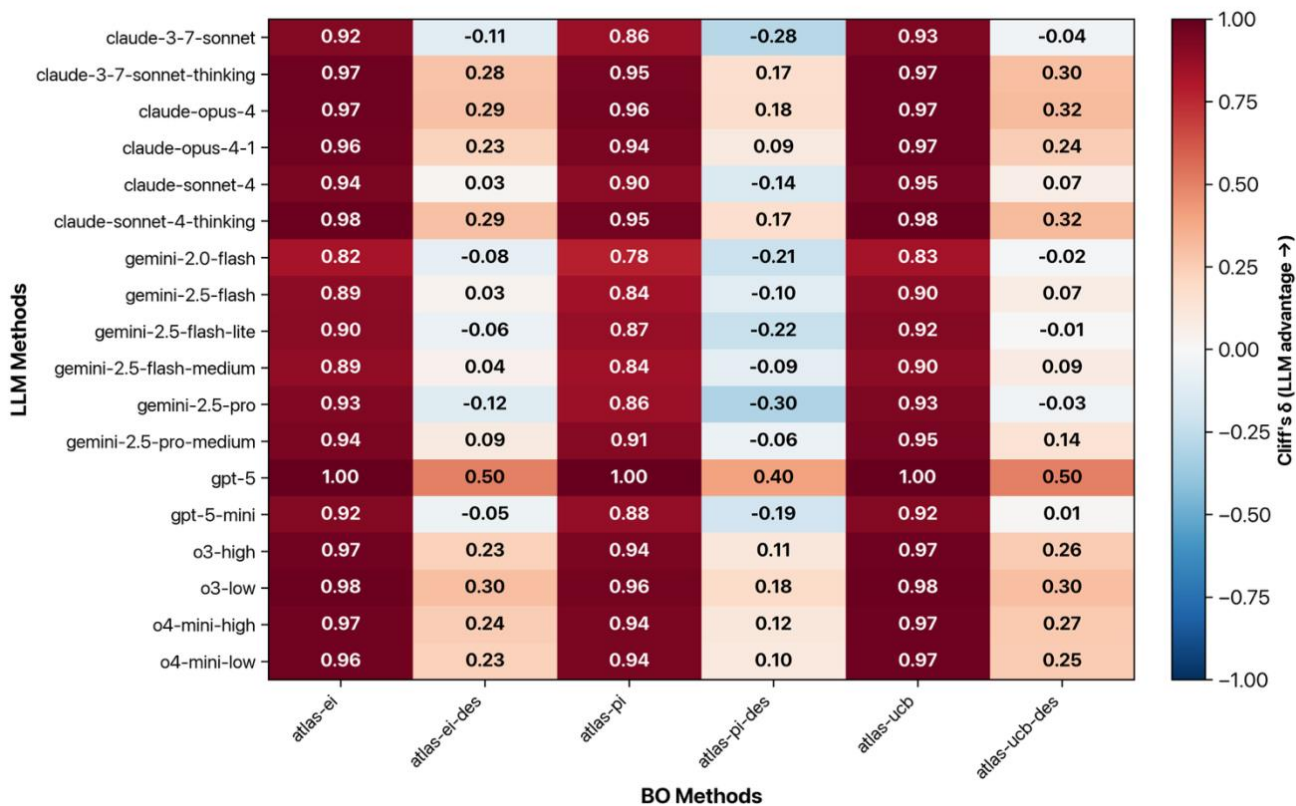

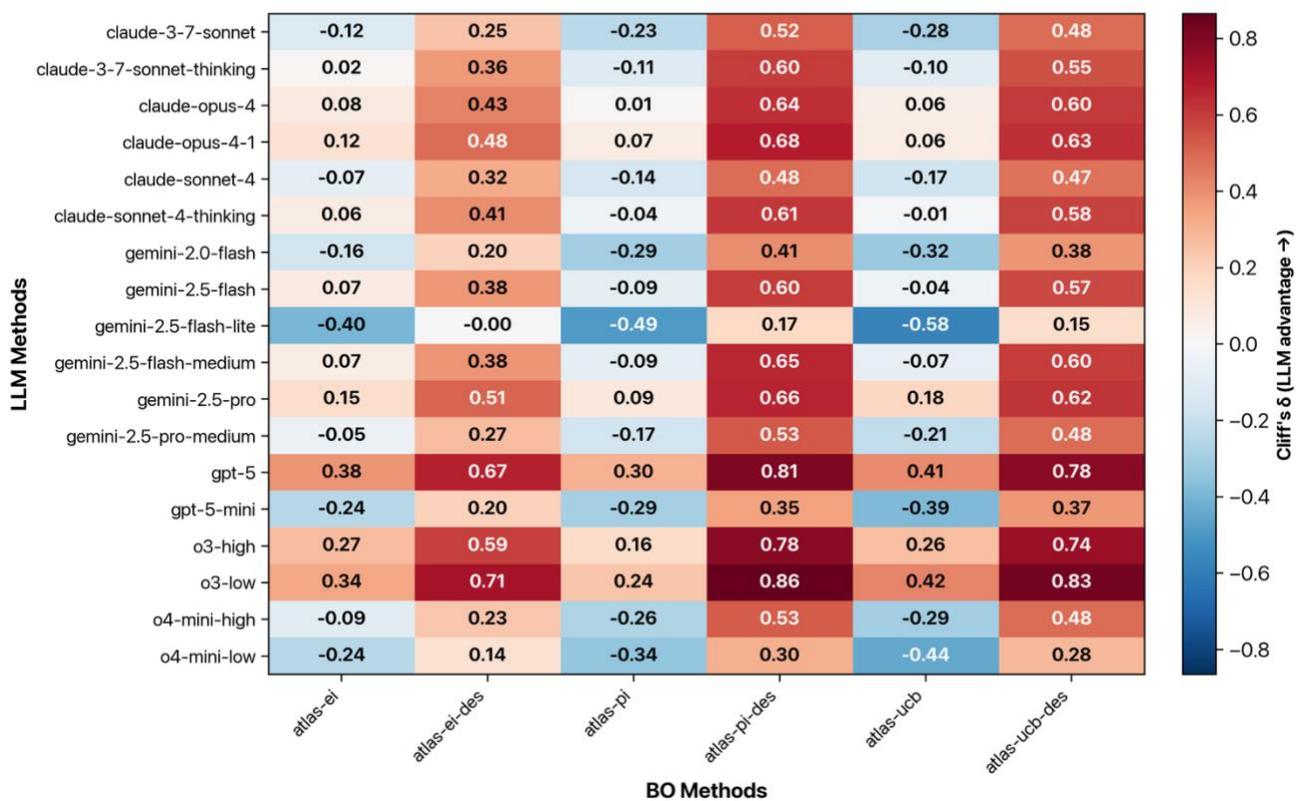

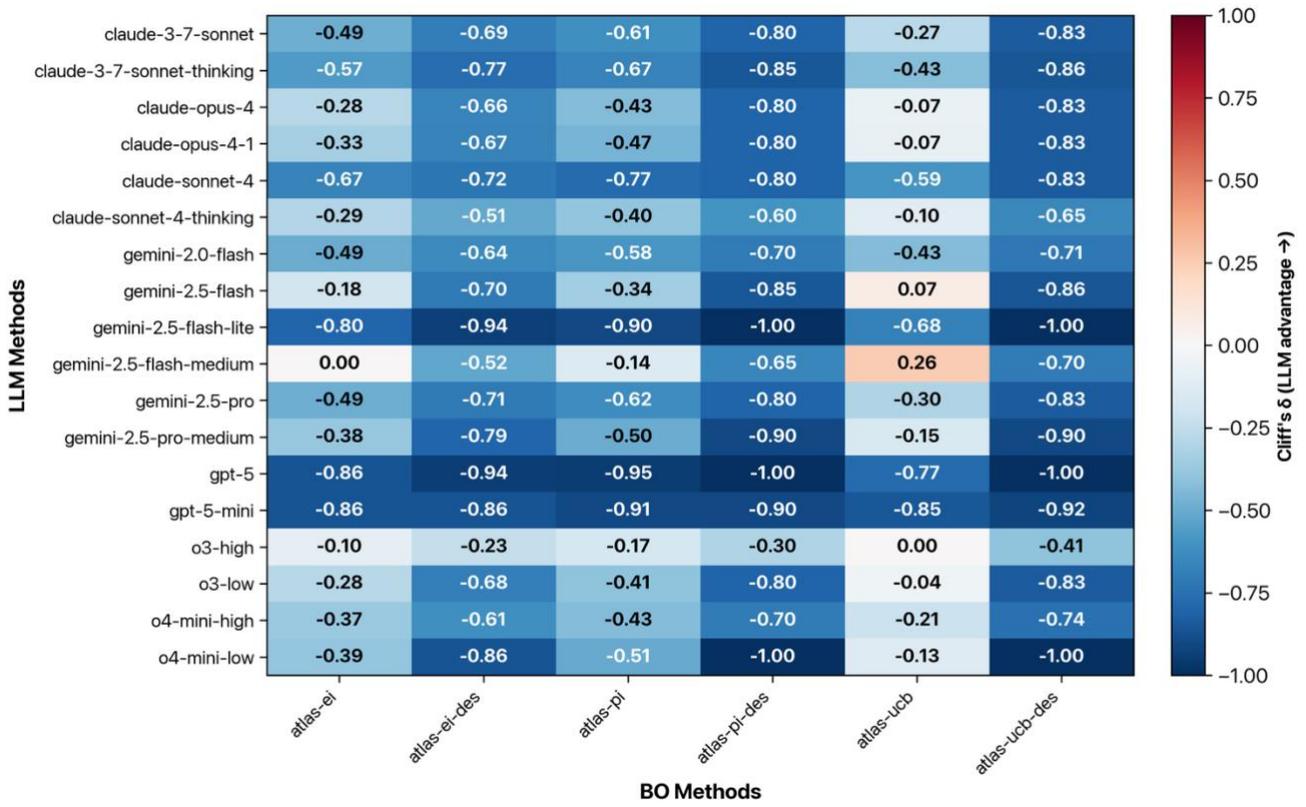
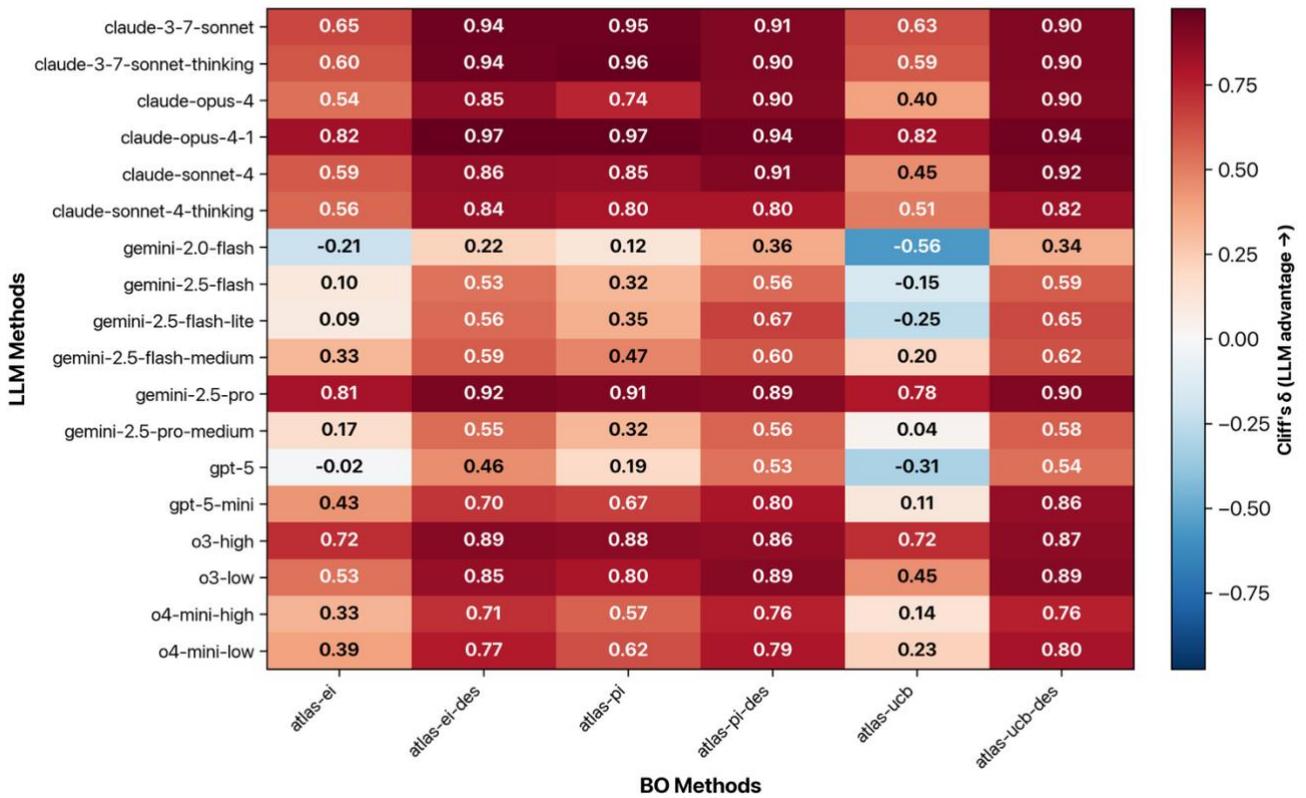

**Figure S2 — Effect sizes for method performance.** For each dataset, we compute Cliff's delta to measure the advantage of all LLM-GO methods compared to all BO methods. Here, advantage measures the performance gain/loss of using an LLM-GO method on a given dataset. Red indicates performance gain, while blue indicates a performance loss.

## B. Bootstrapped Confidence Intervals for Median Method Performance

### Suzuki Conversion

| Method | Median | CI Lower | CI Upper |
|---|---|---|---|
| **Anthropic** | | | |
| claude-3-7-sonnet | 100.0% | 100.0% | 100.0% |
| claude-3-7-sonnet-thinking | 100.0% | 100.0% | 100.0% |
| claude-opus-4 | 100.0% | 100.0% | 100.0% |
| claude-opus-4-1 | 100.0% | 100.0% | 100.0% |
| claude-sonnet-4 | 100.0% | 100.0% | 100.0% |
| claude-sonnet-4-thinking | 100.0% | 100.0% | 100.0% |
| **Google** | | | |
| gemini-2.0-flash | 100.0% | 100.0% | 100.0% |
| gemini-2.5-flash | 100.0% | 100.0% | 100.0% |
| gemini-2.5-flash-lite | 96.0% | 70.8% | 100.0% |
| gemini-2.5-flash-medium | 100.0% | 100.0% | 100.0% |
| gemini-2.5-pro | 100.0% | 100.0% | 100.0% |
| gemini-2.5-pro-medium | 100.0% | 100.0% | 100.0% |
| **OpenAI** | | | |
| gpt-5 | 100.0% | 100.0% | 100.0% |
| gpt-5-mini | 100.0% | 100.0% | 100.0% |
| o3-high | 100.0% | 100.0% | 100.0% |
| o3-low | 100.0% | 100.0% | 100.0% |
| o4-mini-high | 87.2% | 75.0% | 100.0% |
| o4-mini-low | 87.6% | 75.0% | 100.0% |
| **Atlas** | | | |
| atlas-ei | 100.0% | 100.0% | 100.0% |
| atlas-ei-des | 100.0% | 100.0% | 100.0% |
| atlas-pi | 100.0% | 100.0% | 100.0% |
| atlas-pi-des | 100.0% | 100.0% | 100.0% |
| atlas-ucb | 100.0% | 100.0% | 100.0% |
| atlas-ucb-des | 100.0% | 100.0% | 100.0% |
| **Random** | | | |
| random | 100.0% | 100.0% | 100.0% |

### Amide Coupling HTE

| Method | Median | CI Lower | CI Upper |
|---|---|---|---|
| **Anthropic** | | | |
| claude-3-7-sonnet | 98.8% | 98.8% | 98.8% |
| claude-3-7-sonnet-thinking | 98.8% | 98.8% | 98.8% |
| claude-opus-4 | 98.8% | 98.8% | 98.8% |
| claude-opus-4-1 | 98.8% | 98.8% | 98.8% |
| claude-sonnet-4 | 98.8% | 98.8% | 98.8% |
| claude-sonnet-4-thinking | 98.8% | 98.8% | 98.8% |
| **Google** | | | |
| gemini-2.0-flash | 98.8% | 98.8% | 98.8% |
| gemini-2.5-flash | 98.8% | 98.8% | 98.8% |
| gemini-2.5-flash-lite | 98.8% | 97.9% | 98.8% |
| gemini-2.5-flash-medium | 98.8% | 98.8% | 98.8% |
| gemini-2.5-pro | 98.8% | 98.8% | 98.8% |
| gemini-2.5-pro-medium | 98.8% | 98.8% | 98.8% |
| **OpenAI** | | | |
| gpt-5 | 92.7% | 92.7% | 92.7% |
| gpt-5-mini | 97.9% | 89.6% | 98.8% |
| o3-high | 98.8% | 95.8% | 98.8% |
| o3-low | 98.8% | 97.9% | 98.8% |
| o4-mini-high | 98.8% | 97.9% | 98.8% |
| o4-mini-low | 98.8% | 97.9% | 98.8% |
| **Atlas** | | | |
| atlas-ei | 88.5% | 46.1% | 93.5% |
| atlas-ei-des | 92.7% | 90.2% | 95.9% |
| atlas-pi | 92.2% | 81.5% | 93.9% |
| atlas-pi-des | 92.0% | 88.0% | 96.2% |
| atlas-ucb | 92.8% | 87.8% | 93.9% |
| atlas-ucb-des | 92.7% | 91.1% | 99.0% |
| **Random** | | | |
| random | 91.0% | 89.7% | 92.0% |

### Reductive Amination

| Method | Median | CI Lower | CI Upper |
|---|---|---|---|
| **Anthropic** | | | |
| claude-3-7-sonnet | 97.0% | 97.0% | 99.5% |
| claude-3-7-sonnet-thinking | 100.0% | 99.5% | 100.0% |
| claude-opus-4 | 100.0% | 100.0% | 100.0% |
| claude-opus-4-1 | 100.0% | 99.0% | 100.0% |
| claude-sonnet-4 | 98.5% | 96.5% | 100.0% |
| claude-sonnet-4-thinking | 100.0% | 100.0% | 100.0% |
| **Google** | | | |
| gemini-2.0-flash | 98.0% | 97.5% | 100.0% |
| gemini-2.5-flash | 99.0% | 98.0% | 100.0% |
| gemini-2.5-flash-lite | 98.0% | 94.0% | 100.0% |
| gemini-2.5-flash-medium | 99.0% | 98.0% | 100.0% |
| gemini-2.5-pro | 97.0% | 97.0% | 99.0% |
| gemini-2.5-pro-medium | 100.0% | 97.0% | 100.0% |
| **OpenAI** | | | |
| gpt-5 | 100.0% | 100.0% | 100.0% |
| gpt-5-mini | 99.0% | 97.5% | 100.0% |
| o3-high | 100.0% | 99.0% | 100.0% |
| o3-low | 100.0% | 99.0% | 100.0% |
| o4-mini-high | 100.0% | 99.0% | 100.0% |
| o4-mini-low | 100.0% | 99.0% | 100.0% |
| **Atlas** | | | |
| atlas-ei | 76.0% | 48.0% | 76.5% |
| atlas-ei-des | 99.5% | 94.5% | 100.0% |
| atlas-pi | 76.0% | 48.0% | 76.0% |
| atlas-pi-des | 98.0% | 98.0% | 100.0% |
| atlas-ucb | 72.5% | 48.0% | 76.0% |
| atlas-ucb-des | 99.5% | 92.5% | 100.0% |
| **Random** | | | |
| random | 99.0% | 91.0% | 100.0% |

### Suzuki Yield

| Method | Median | CI Lower | CI Upper |
|---|---|---|---|
| **Anthropic** | | | |
| claude-3-7-sonnet | 94.1% | 93.1% | 95.5% |
| claude-3-7-sonnet-thinking | 94.3% | 93.1% | 97.0% |
| claude-opus-4 | 96.3% | 94.0% | 97.0% |
| claude-opus-4-1 | 95.6% | 94.5% | 97.3% |
| claude-sonnet-4 | 94.0% | 91.9% | 97.3% |
| claude-sonnet-4-thinking | 95.2% | 93.7% | 97.3% |
| **Google** | | | |
| gemini-2.0-flash | 93.5% | 92.3% | 95.7% |
| gemini-2.5-flash | 95.2% | 94.3% | 96.2% |
| gemini-2.5-flash-lite | 92.9% | 86.8% | 94.1% |
| gemini-2.5-flash-medium | 94.8% | 94.0% | 96.5% |
| gemini-2.5-pro | 96.2% | 95.0% | 97.3% |
| gemini-2.5-pro-medium | 95.3% | 94.5% | 95.5% |
| **OpenAI** | | | |
| gpt-5 | 97.0% | 96.0% | 99.2% |
| gpt-5-mini | 93.0% | 91.1% | 95.8% |
| o3-high | 96.5% | 96.0% | 97.6% |
| o3-low | 97.0% | 96.5% | 97.0% |
| o4-mini-high | 93.7% | 93.4% | 95.0% |
| o4-mini-low | 92.8% | 90.6% | 94.5% |
| **Atlas** | | | |
| atlas-ei | 95.7% | 91.5% | 96.8% |
| atlas-ei-des | 91.5% | 76.5% | 96.2% |
| atlas-pi | 96.1% | 94.7% | 96.7% |
| atlas-pi-des | 90.7% | 75.0% | 93.1% |
| atlas-ucb | 96.2% | 94.6% | 96.5% |
| atlas-ucb-des | 90.1% | 77.9% | 94.4% |
| **Random** | | | |
| random | 94.2% | 91.4% | 95.3% |

### Chan-Lam

| Method | Median | CI Lower | CI Upper |
|---|---|---|---|
| **Anthropic** | | | |
| claude-3-7-sonnet | 58.0% | 51.9% | 62.9% |
| claude-3-7-sonnet-thinking | 51.4% | 44.3% | 58.0% |
| claude-opus-4 | 62.1% | 58.1% | 68.2% |
| claude-opus-4-1 | 62.9% | 56.4% | 68.2% |
| claude-sonnet-4 | 48.9% | 47.9% | 51.8% |
| claude-sonnet-4-thinking | 59.4% | 51.9% | 72.4% |
| **Google** | | | |
| gemini-2.0-flash | 46.1% | 36.5% | 61.6% |
| gemini-2.5-flash | 68.2% | 56.1% | 69.3% |
| gemini-2.5-flash-lite | 47.0% | 23.9% | 57.2% |
| gemini-2.5-flash-medium | 69.3% | 63.9% | 70.5% |
| gemini-2.5-pro | 53.7% | 50.3% | 65.0% |
| gemini-2.5-pro-medium | 61.7% | 44.1% | 69.3% |
| **OpenAI** | | | |
| gpt-5 | 46.6% | 39.7% | 50.9% |
| gpt-5-mini | 40.6% | 40.6% | 46.2% |
| o3-high | 67.5% | 54.7% | 75.1% |
| o3-low | 63.3% | 54.3% | 72.4% |
| o4-mini-high | 54.9% | 40.6% | 72.4% |
| o4-mini-low | 63.3% | 48.9% | 73.1% |
| **Atlas** | | | |
| atlas-ei | 64.4% | 62.9% | 74.7% |
| atlas-ei-des | 74.9% | 74.9% | 74.9% |
| atlas-pi | 73.5% | 62.9% | 74.7% |
| atlas-pi-des | 74.9% | 74.9% | 74.9% |
| atlas-ucb | 65.0% | 58.0% | 67.7% |
| atlas-ucb-des | 74.9% | 74.9% | 74.9% |
| **Random** | | | |
| random | 58.5% | 52.2% | 62.9% |

### Buchwald-Hartwig

| Method | Median | CI Lower | CI Upper |
|---|---|---|---|
| **Anthropic** | | | |
| claude-3-7-sonnet | 95.8% | 94.4% | 96.9% |
| claude-3-7-sonnet-thinking | 95.4% | 95.1% | 95.8% |
| claude-opus-4 | 92.4% | 90.6% | 97.5% |
| claude-opus-4-1 | 98.0% | 96.9% | 99.7% |
| claude-sonnet-4 | 94.7% | 91.3% | 97.2% |
| claude-sonnet-4-thinking | 96.9% | 94.1% | 98.0% |
| **Google** | | | |
| gemini-2.0-flash | 81.0% | 74.2% | 82.5% |
| gemini-2.5-flash | 84.8% | 77.5% | 91.7% |
| gemini-2.5-flash-lite | 85.1% | 81.6% | 88.8% |
| gemini-2.5-flash-medium | 94.9% | 83.3% | 98.9% |
| gemini-2.5-pro | 98.9% | 96.9% | 99.7% |
| gemini-2.5-pro-medium | 85.8% | 79.6% | 98.0% |
| **OpenAI** | | | |
| gpt-5 | 83.0% | 79.8% | 85.5% |
| gpt-5-mini | 85.2% | 84.3% | 93.6% |
| o3-high | 98.0% | 96.7% | 99.7% |
| o3-low | 94.8% | 92.8% | 98.0% |
| o4-mini-high | 87.1% | 83.5% | 97.5% |
| o4-mini-low | 90.4% | 85.7% | 96.9% |
| **Atlas** | | | |
| atlas-ei | 81.2% | 80.5% | 91.4% |
| atlas-ei-des | 76.0% | 62.7% | 78.7% |
| atlas-pi | 80.5% | 77.0% | 80.5% |
| atlas-pi-des | 76.0% | 67.6% | 76.3% |
| atlas-ucb | 89.4% | 81.2% | 92.8% |
| atlas-ucb-des | 75.0% | 63.4% | 79.0% |
| **Random** | | | |
| random | 83.9% | 81.5% | 84.6% |

**Figure S3 — Bootstrapped CIs for median method performance.** For each dataset-method combination, the confidence interval lower and upper bounds are shown for median performance. More red text indicates better performance, while more blue text indicates worse performance.

## C. Convergence Analysis

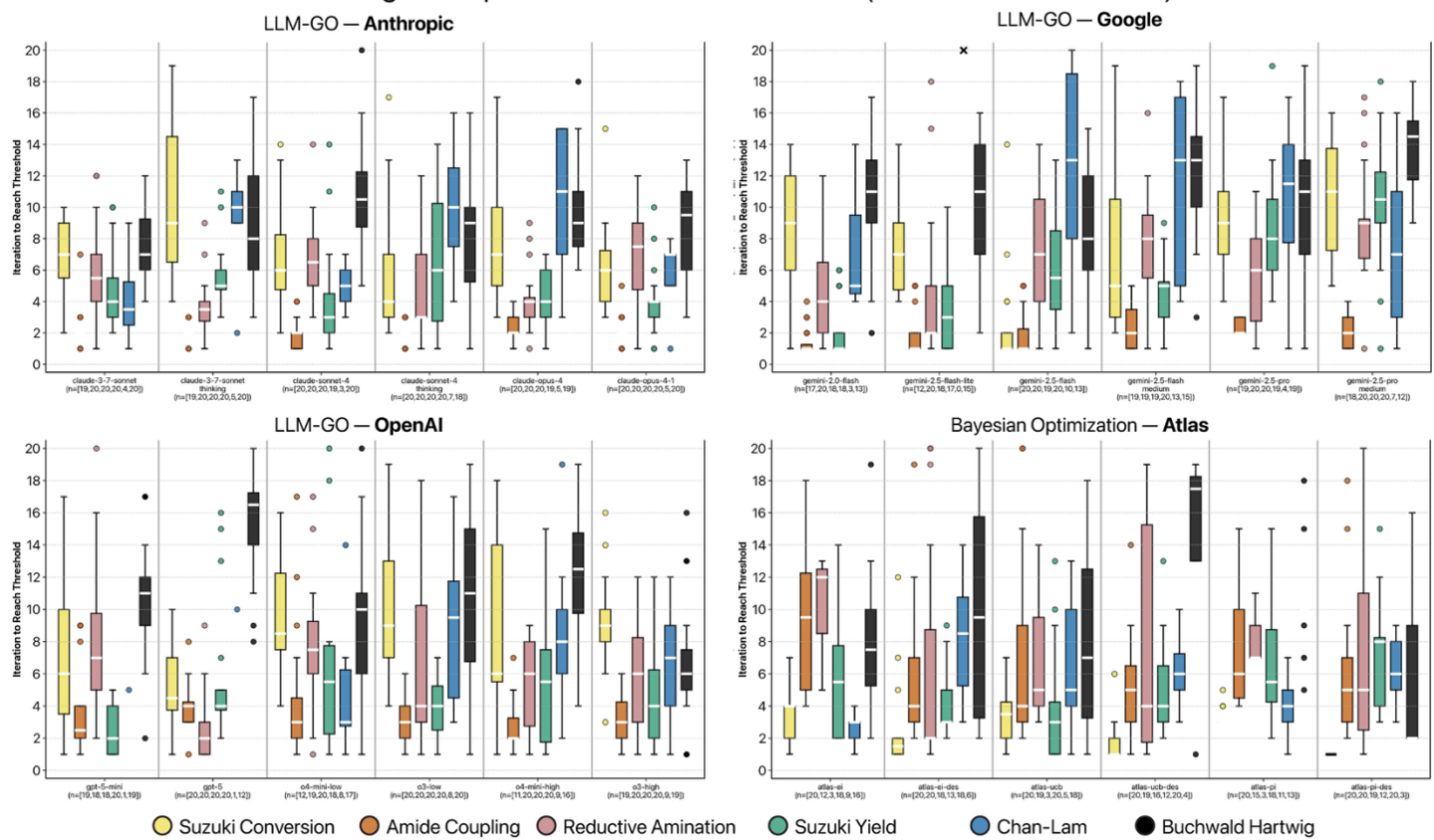

**Figure S4 — Convergence speed to 80% of the maximum.** Each individual boxplots shows the distribution of iterations at which the first objective value greater than 80% of the maximal value was found. Boxplots are plotted with varying sample sizes. Underneath each method shows the number of points making up the distribution for each dataset (left to right). This accounts for individual campaigns where a suitable objective value was not obtained.

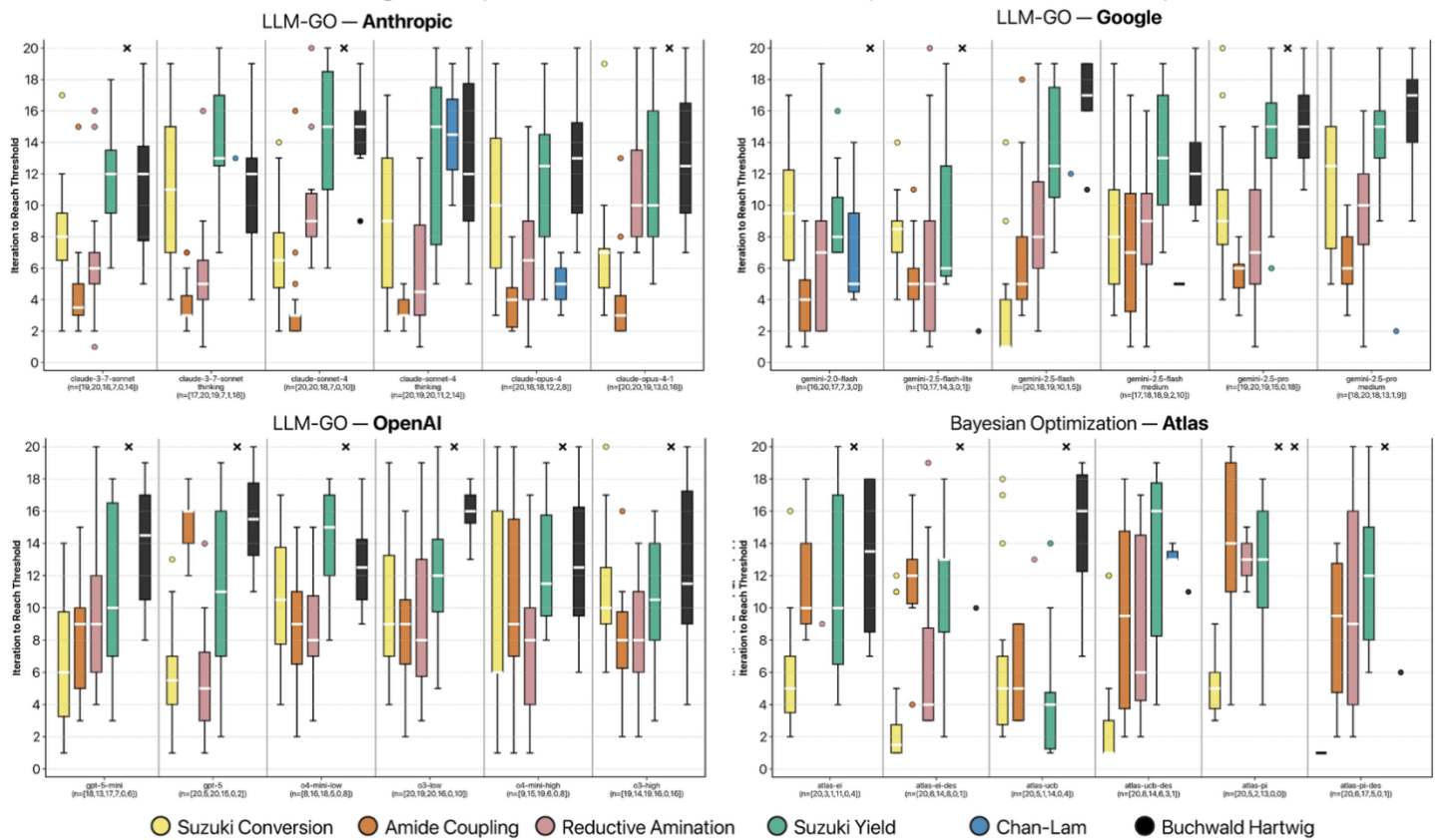

**Figure S5 — Convergence speed to 95% of the maximum.** Each individual boxplots shows the distribution of iterations at which the first objective value greater than 95% of the maximal value was found. Boxplots are plotted with varying sample sizes. Underneath each method shows the number of points making up the distribution for each dataset (left to right). This accounts for individual campaigns where a suitable objective value was not obtained.

# D. Statistical Significance and Effect Size Analysis for Method Cumulative Entropy

## Statistical Significance (**p-values**) for LLM vs. BO Method Comparisons

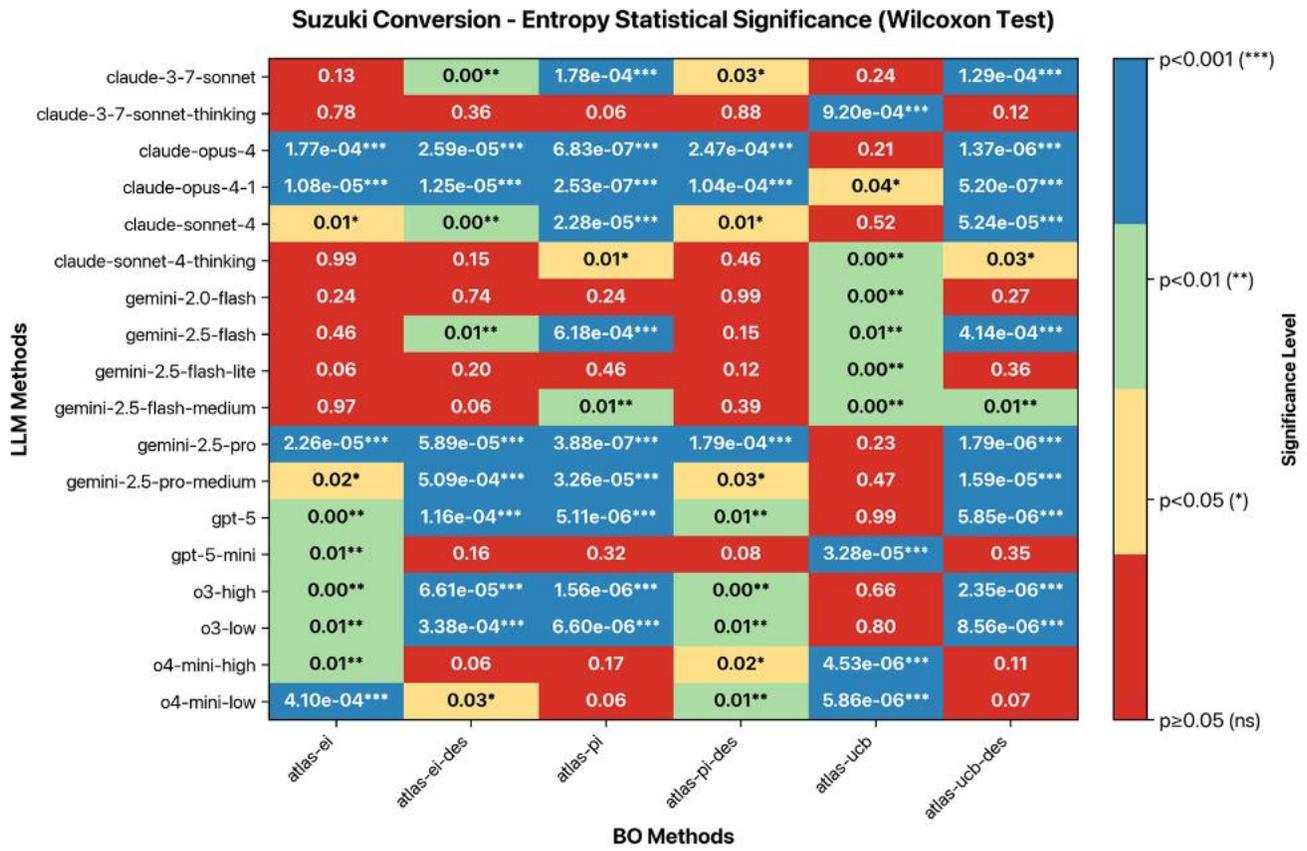

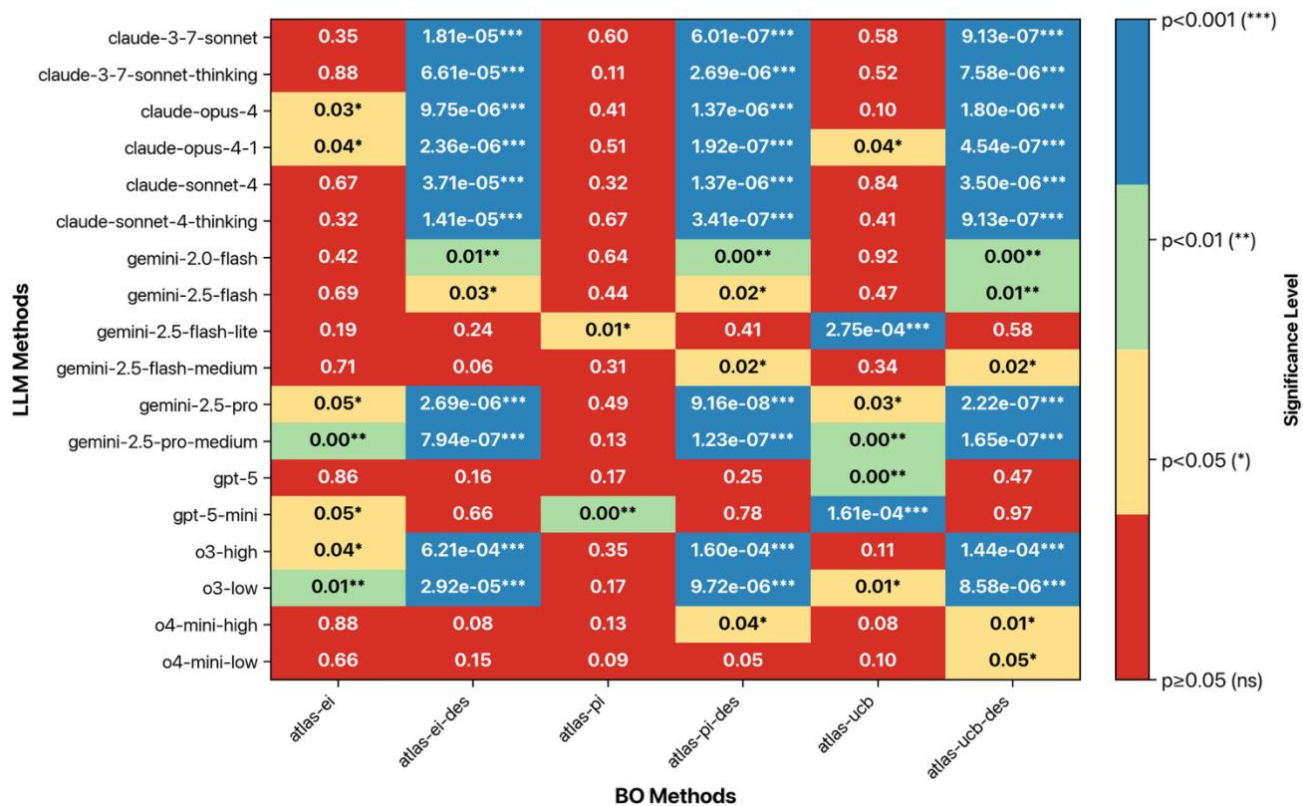
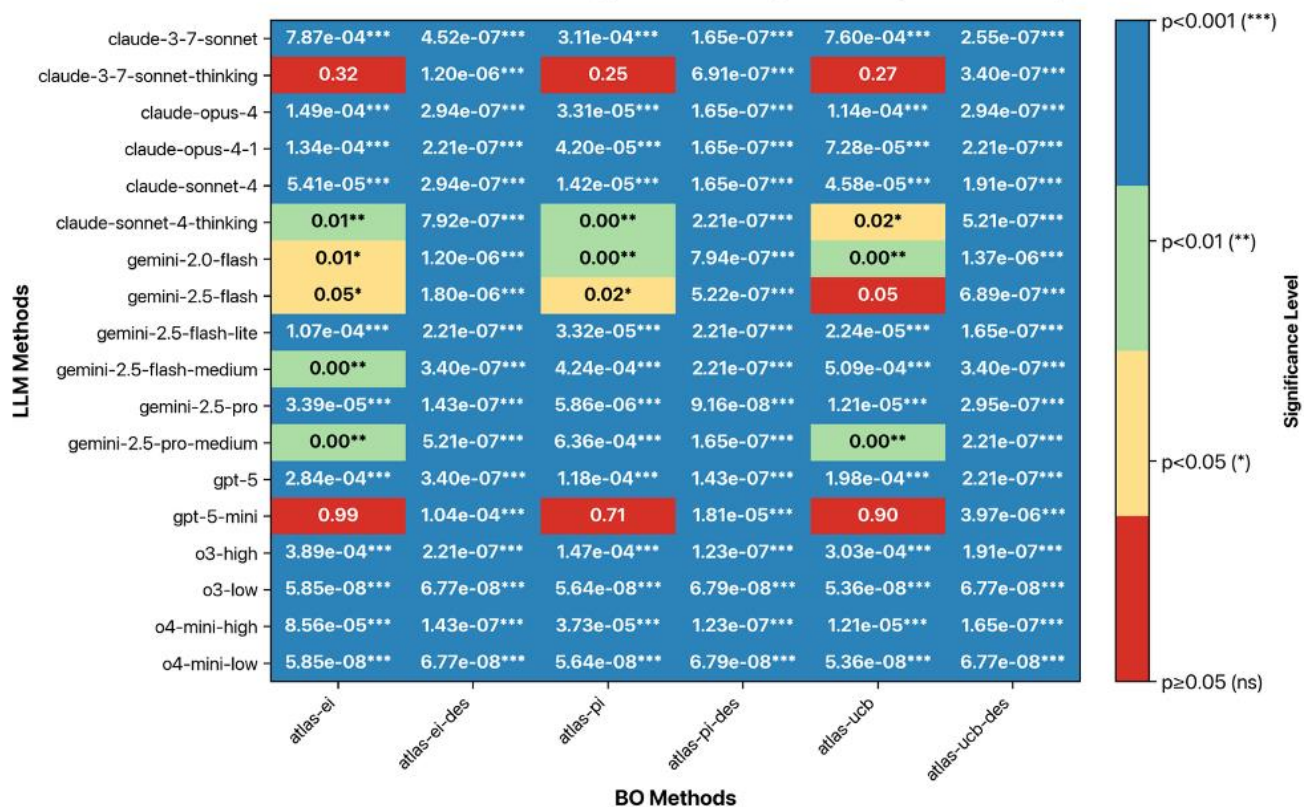

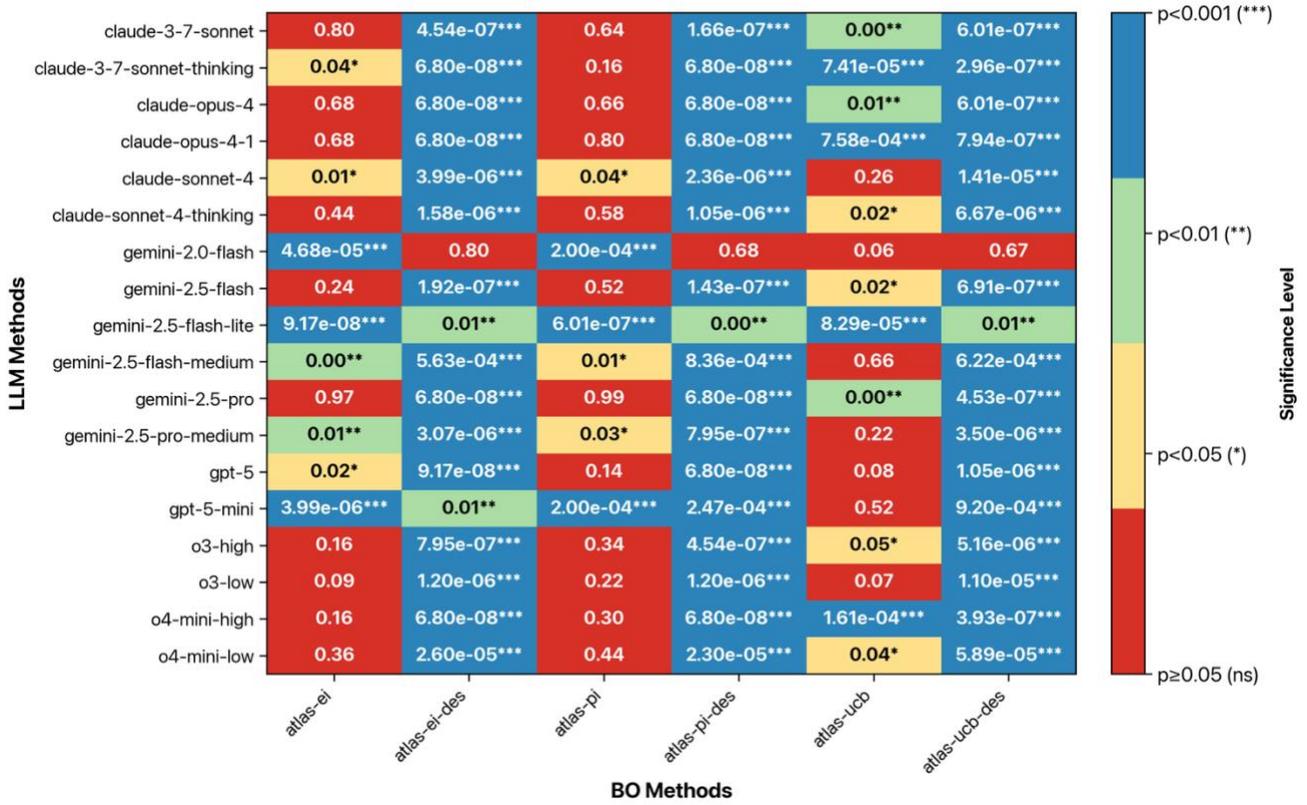

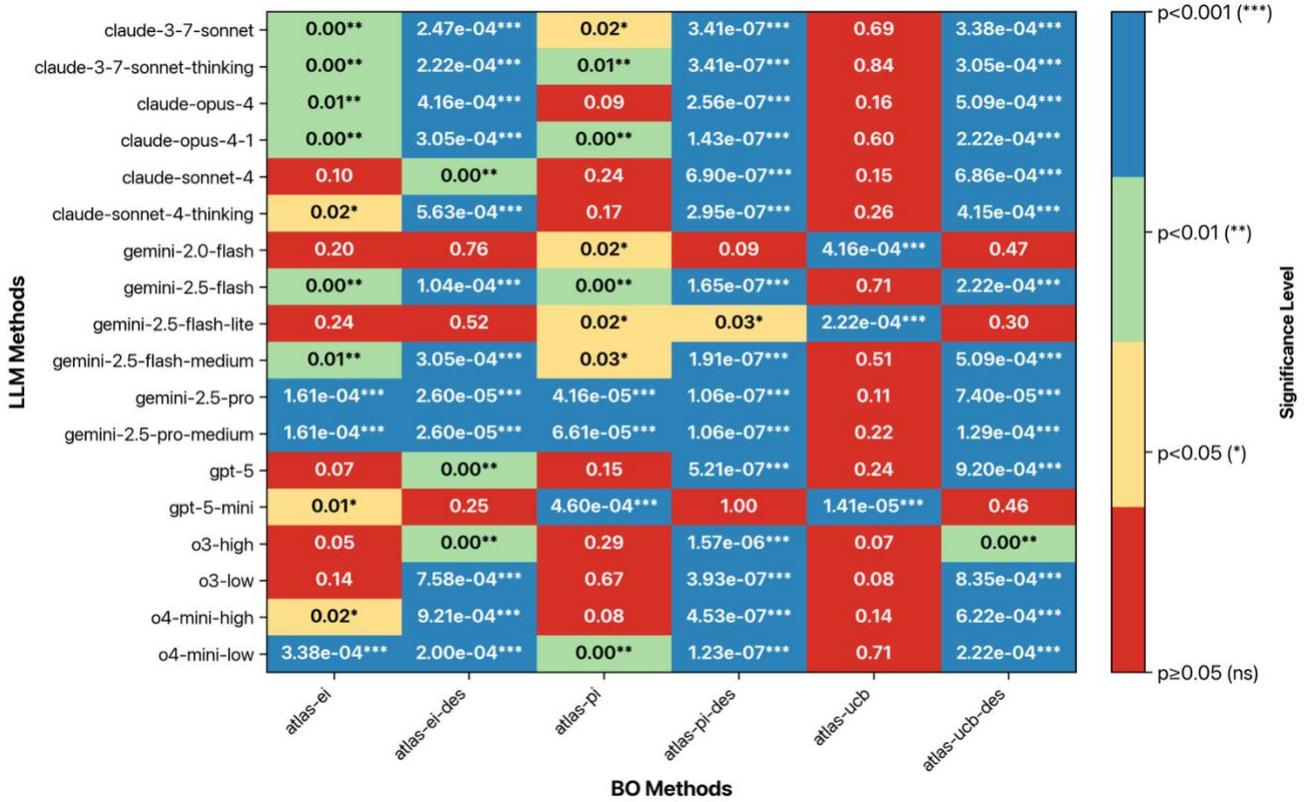

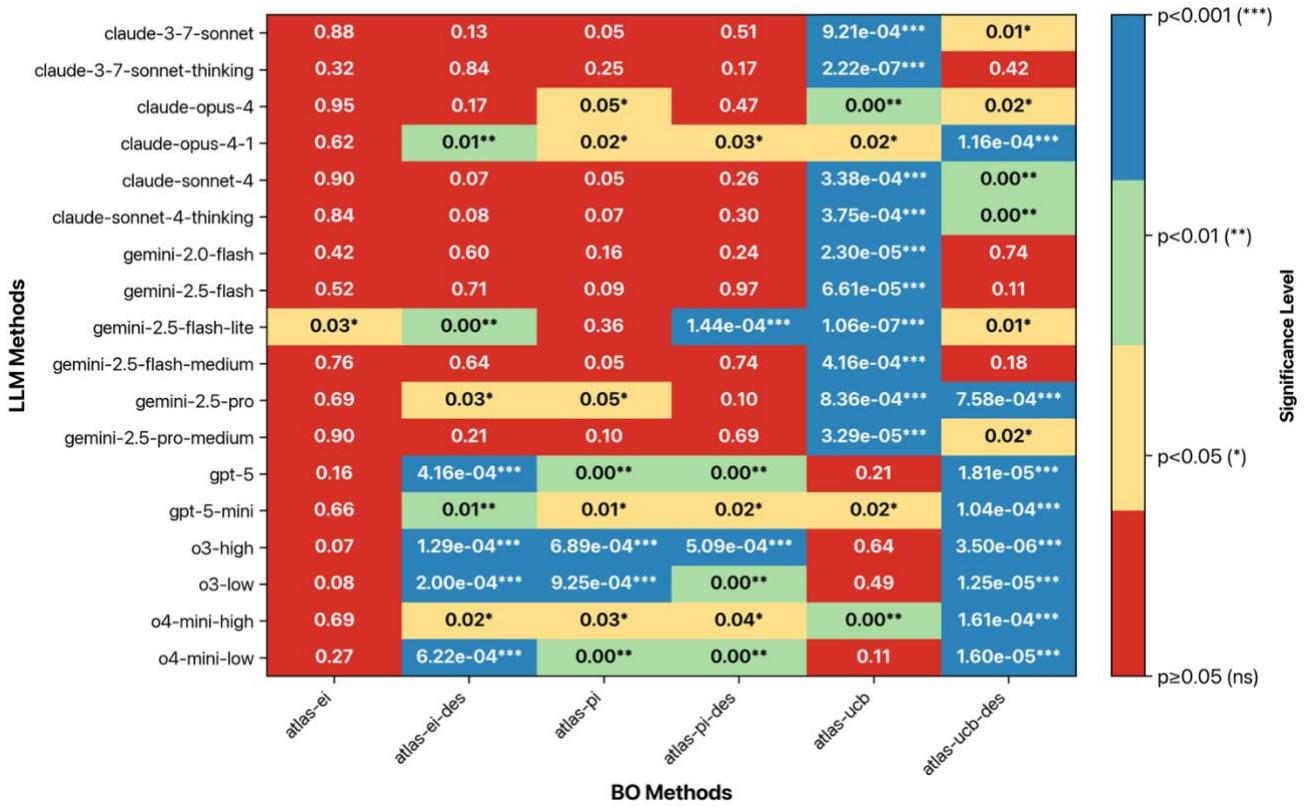

**Figure S6 — Statistical significance tests for method cumulative entropy.** For each dataset, we conduct Wilcoxon tests to compare the cumulative entropy distributions of all LLM-GO methods with all BO methods. Red indicates the cumulative entropy distribution is not significantly different. All other colors indicate a varying level of significance, indicated by the color bar on the right.

# Effect Sizes (**Cliff's Delta**) for LLM-GO vs. BO Method Comparisons

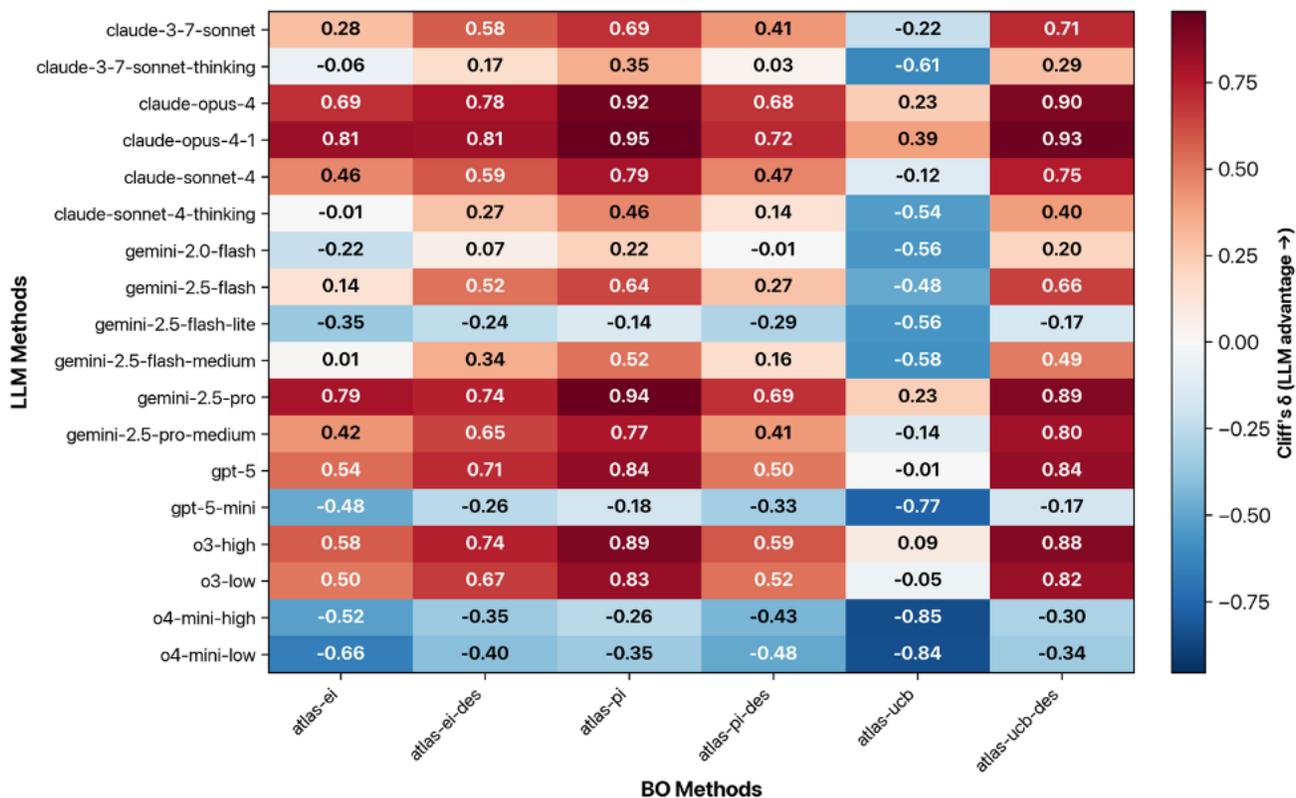

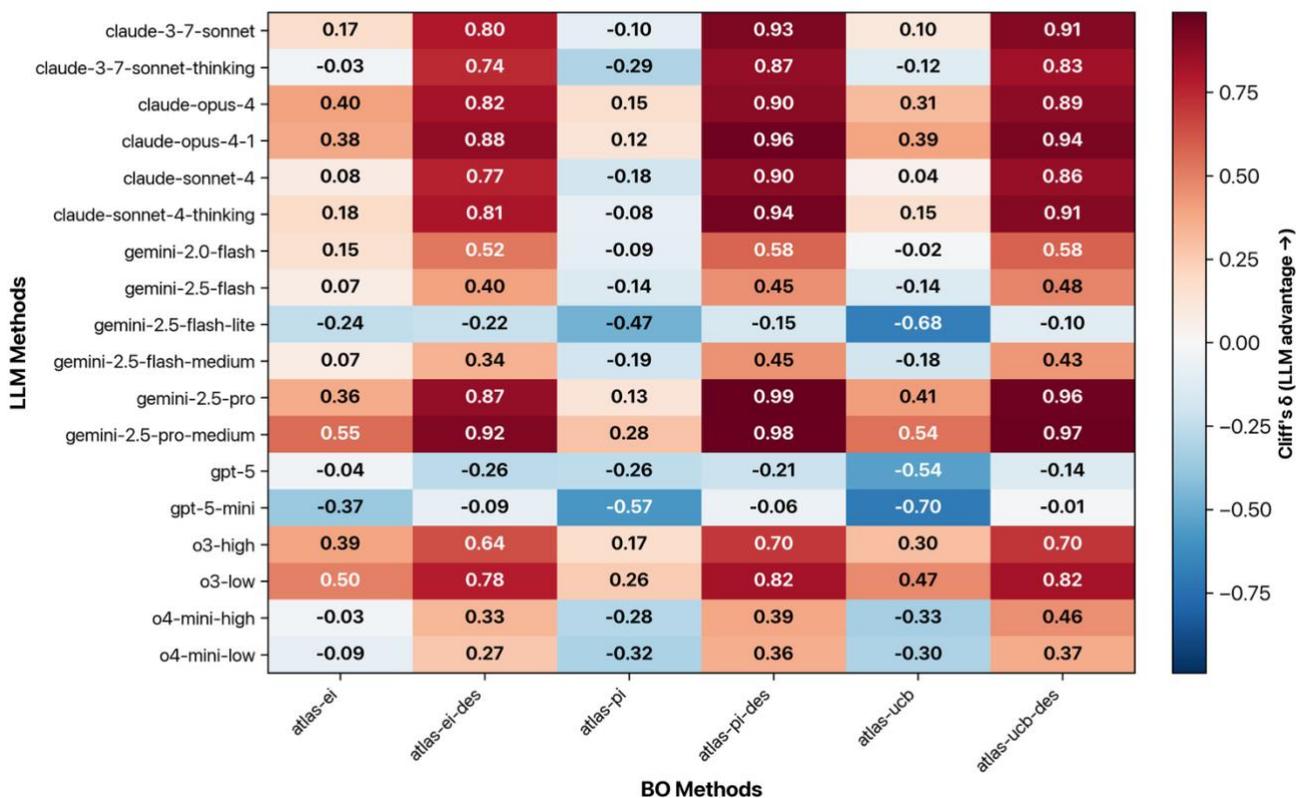

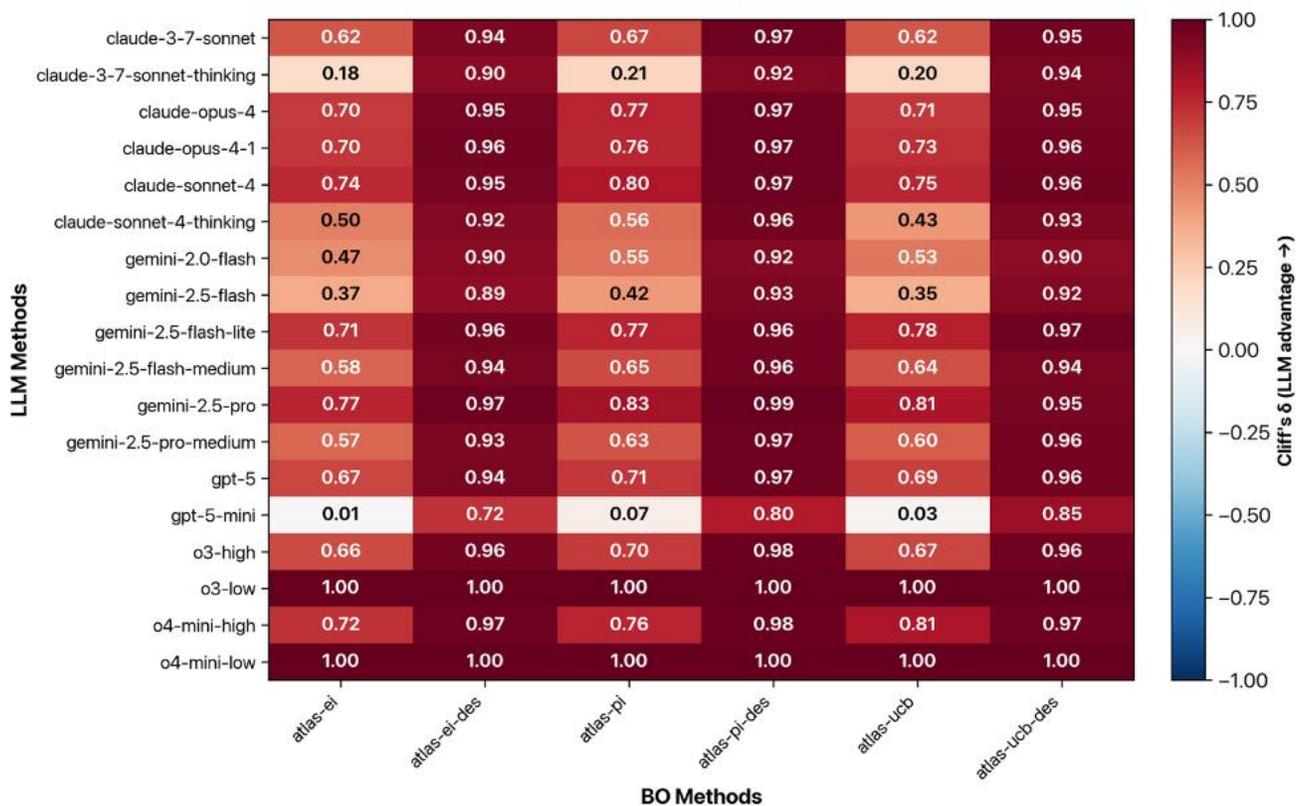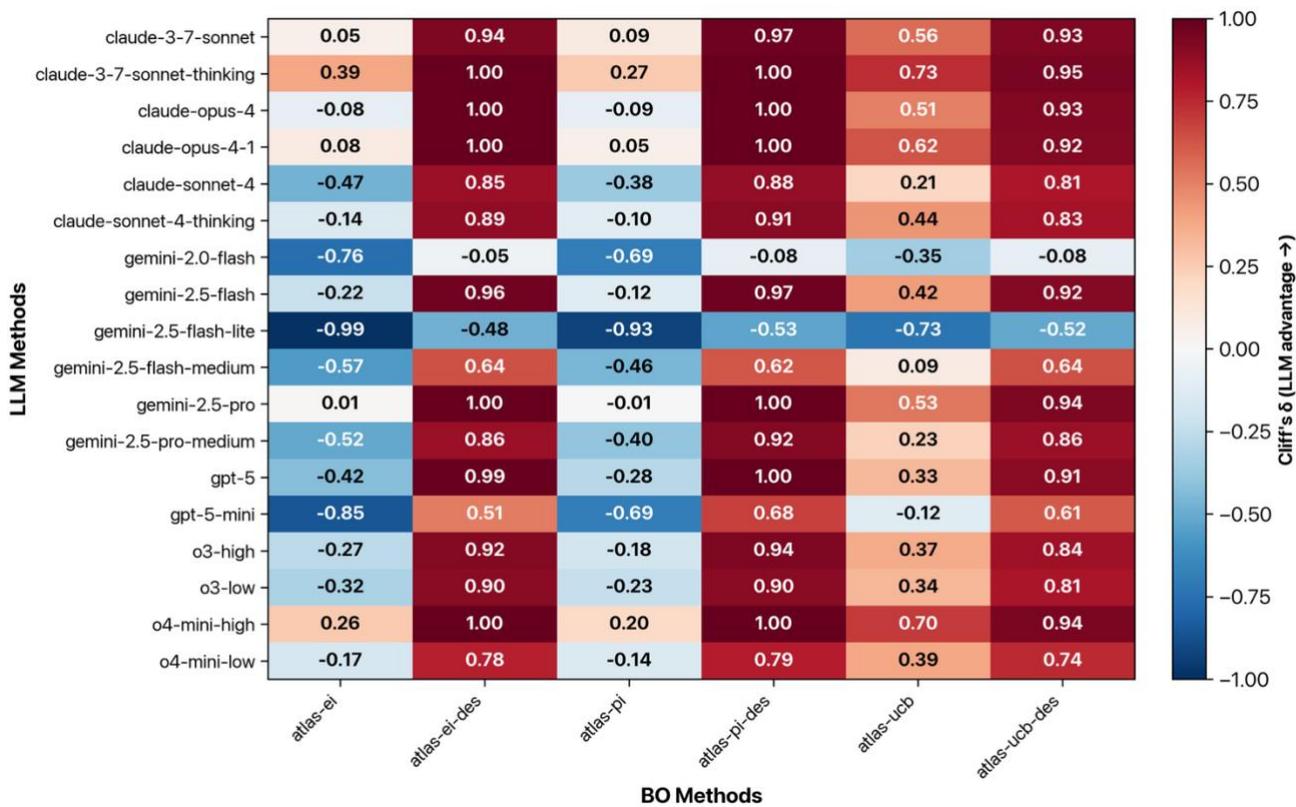

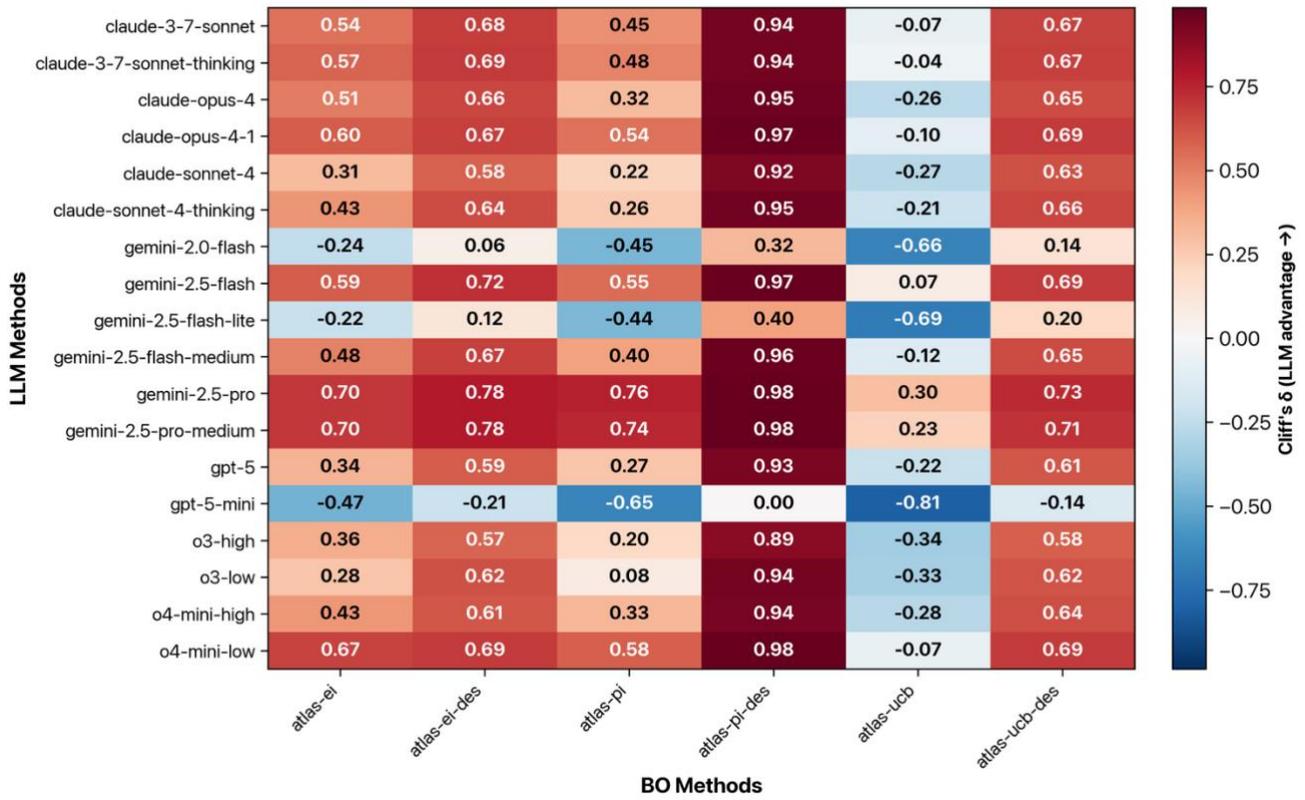
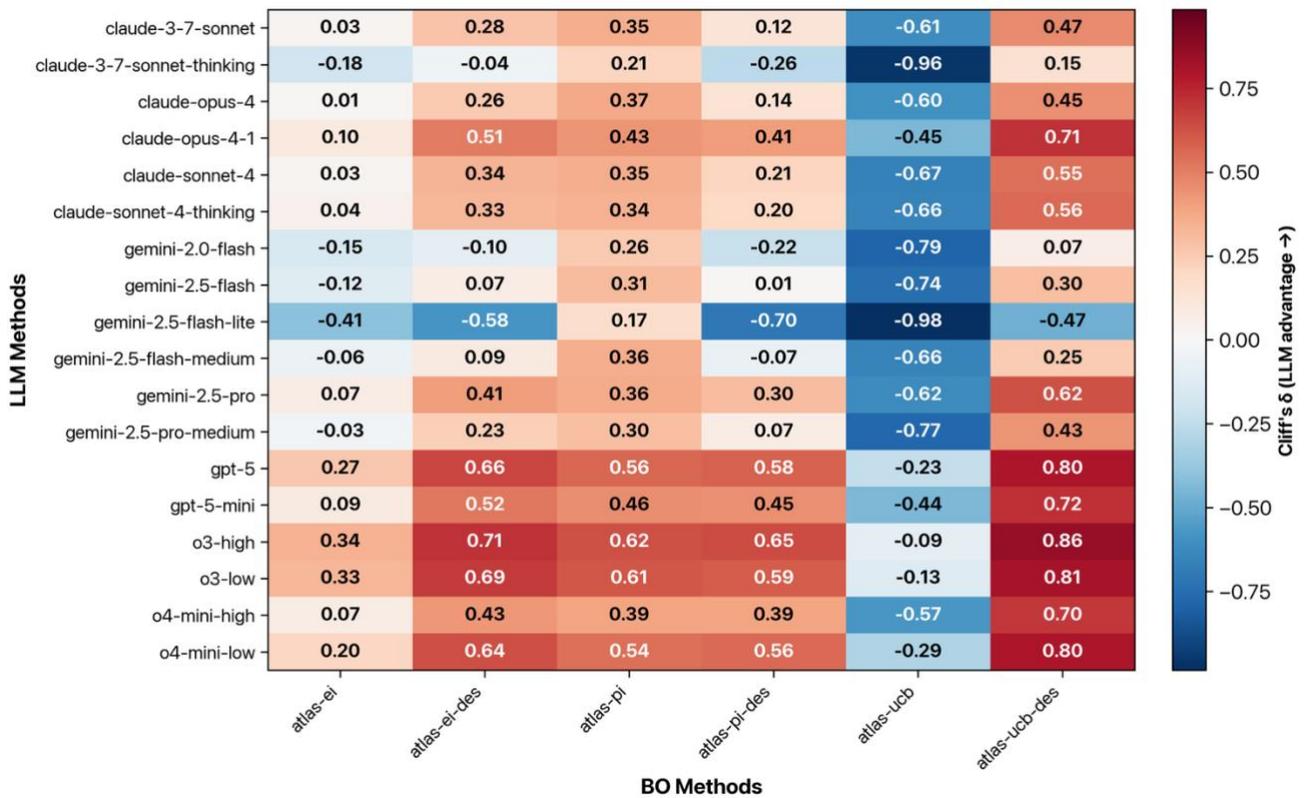

**Figure S7 — Effect sizes for method cumulative entropy.** For each dataset, we compute Cliff's delta to measure the advantage of all LLM-GO methods compared to all BO methods. Here, advantage measures the exploration gain/loss of using an LLM-GO method on a given dataset. Red indicates exploration gain, while blue indicates an exploration loss.

# E. Bootstrapped Confidence Intervals for Median Cumulative Entropy

### Suzuki Conversion

| Method | Median | CI Lower | CI Upper |
|---|---|---|---|
| **Anthropic** | | | |
| claude-3-7-sonnet | 0.568 | 0.543 | 0.589 |
| claude-3-7-sonnet-thinking | 0.534 | 0.512 | 0.553 |
| claude-opus-4 | 0.600 | 0.586 | 0.648 |
| claude-opus-4-1 | 0.613 | 0.591 | 0.644 |
| claude-sonnet-4 | 0.578 | 0.545 | 0.602 |
| claude-sonnet-4-thinking | 0.525 | 0.515 | 0.556 |
| **Google** | | | |
| gemini-2.0-flash | 0.510 | 0.486 | 0.572 |
| gemini-2.5-flash | 0.540 | 0.535 | 0.560 |
| gemini-2.5-flash-lite | 0.478 | 0.358 | 0.540 |
| gemini-2.5-flash-medium | 0.543 | 0.525 | 0.557 |
| gemini-2.5-pro | 0.603 | 0.590 | 0.612 |
| gemini-2.5-pro-medium | 0.562 | 0.549 | 0.608 |
| **OpenAI** | | | |
| gpt-5 | 0.577 | 0.566 | 0.604 |
| gpt-5-mini | 0.474 | 0.440 | 0.530 |
| o3-high | 0.587 | 0.573 | 0.618 |
| o3-low | 0.586 | 0.568 | 0.603 |
| o4-mini-high | 0.444 | 0.336 | 0.519 |
| o4-mini-low | 0.450 | 0.407 | 0.508 |
| **Atlas** | | | |
| atlas-ei | 0.525 | 0.502 | 0.580 |
| atlas-ei-des | 0.503 | 0.475 | 0.530 |
| atlas-pi | 0.494 | 0.477 | 0.501 |
| atlas-pi-des | 0.504 | 0.469 | 0.571 |
| atlas-ucb | 0.592 | 0.567 | 0.609 |
| atlas-ucb-des | 0.485 | 0.463 | 0.522 |

### Amide Coupling HTE

| Method | Median | CI Lower | CI Upper |
|---|---|---|---|
| **Anthropic** | | | |
| claude-3-7-sonnet | 0.728 | 0.700 | 0.740 |
| claude-3-7-sonnet-thinking | 0.701 | 0.681 | 0.719 |
| claude-opus-4 | 0.734 | 0.715 | 0.766 |
| claude-opus-4-1 | 0.735 | 0.715 | 0.761 |
| claude-sonnet-4 | 0.715 | 0.704 | 0.732 |
| claude-sonnet-4-thinking | 0.728 | 0.702 | 0.739 |
| **Google** | | | |
| gemini-2.0-flash | 0.722 | 0.645 | 0.742 |
| gemini-2.5-flash | 0.689 | 0.603 | 0.743 |
| gemini-2.5-flash-lite | 0.547 | 0.423 | 0.635 |
| gemini-2.5-flash-medium | 0.695 | 0.626 | 0.735 |
| gemini-2.5-pro | 0.742 | 0.729 | 0.749 |
| gemini-2.5-pro-medium | 0.752 | 0.737 | 0.764 |
| **OpenAI** | | | |
| gpt-5 | 0.524 | 0.423 | 0.597 |
| gpt-5-mini | 0.609 | 0.396 | 0.639 |
| o3-high | 0.742 | 0.710 | 0.762 |
| o3-low | 0.757 | 0.723 | 0.769 |
| o4-mini-high | 0.647 | 0.592 | 0.718 |
| o4-mini-low | 0.658 | 0.576 | 0.715 |
| **Atlas** | | | |
| atlas-ei | 0.713 | 0.349 | 0.736 |
| atlas-ei-des | 0.577 | 0.551 | 0.617 |
| atlas-pi | 0.731 | 0.688 | 0.755 |
| atlas-pi-des | 0.566 | 0.538 | 0.618 |
| atlas-ucb | 0.704 | 0.661 | 0.742 |
| atlas-ucb-des | 0.552 | 0.489 | 0.611 |

### Reductive Amination

| Method | Median | CI Lower | CI Upper |
|---|---|---|---|
| **Anthropic** | | | |
| claude-3-7-sonnet | 0.270 | 0.235 | 0.301 |
| claude-3-7-sonnet-thinking | 0.237 | 0.196 | 0.317 |
| claude-opus-4 | 0.268 | 0.230 | 0.342 |
| claude-opus-4-1 | 0.317 | 0.248 | 0.340 |
| claude-sonnet-4 | 0.343 | 0.232 | 0.353 |
| claude-sonnet-4-thinking | 0.226 | 0.218 | 0.266 |
| **Google** | | | |
| gemini-2.0-flash | 0.243 | 0.235 | 0.266 |
| gemini-2.5-flash | 0.236 | 0.220 | 0.261 |
| gemini-2.5-flash-lite | 0.295 | 0.260 | 0.316 |
| gemini-2.5-flash-medium | 0.255 | 0.233 | 0.283 |
| gemini-2.5-pro | 0.267 | 0.251 | 0.318 |
| gemini-2.5-pro-medium | 0.273 | 0.222 | 0.293 |
| **OpenAI** | | | |
| gpt-5 | 0.298 | 0.243 | 0.312 |
| gpt-5-mini | 0.219 | 0.190 | 0.251 |
| o3-high | 0.291 | 0.236 | 0.307 |
| o3-low | 0.641 | 0.595 | 0.688 |
| o4-mini-high | 0.292 | 0.274 | 0.304 |
| o4-mini-low | 0.654 | 0.632 | 0.714 |
| **Atlas** | | | |
| atlas-ei | 0.209 | 0.200 | 0.213 |
| atlas-ei-des | 0.106 | 0.087 | 0.135 |
| atlas-pi | 0.200 | 0.200 | 0.213 |
| atlas-pi-des | 0.119 | 0.063 | 0.130 |
| atlas-ucb | 0.200 | 0.200 | 0.225 |
| atlas-ucb-des | 0.067 | 0.051 | 0.094 |

### Suzuki Yield

| Method | Median | CI Lower | CI Upper |
|---|---|---|---|
| **Anthropic** | | | |
| claude-3-7-sonnet | 0.556 | 0.520 | 0.597 |
| claude-3-7-sonnet-thinking | 0.579 | 0.541 | 0.615 |
| claude-opus-4 | 0.544 | 0.514 | 0.573 |
| claude-opus-4-1 | 0.551 | 0.539 | 0.569 |
| claude-sonnet-4 | 0.521 | 0.477 | 0.537 |
| claude-sonnet-4-thinking | 0.539 | 0.520 | 0.556 |
| **Google** | | | |
| gemini-2.0-flash | 0.410 | 0.284 | 0.501 |
| gemini-2.5-flash | 0.528 | 0.510 | 0.554 |
| gemini-2.5-flash-lite | 0.287 | 0.268 | 0.385 |
| gemini-2.5-flash-medium | 0.484 | 0.448 | 0.519 |
| gemini-2.5-pro | 0.540 | 0.521 | 0.583 |
| gemini-2.5-pro-medium | 0.507 | 0.491 | 0.530 |
| **OpenAI** | | | |
| gpt-5 | 0.519 | 0.494 | 0.542 |
| gpt-5-mini | 0.453 | 0.443 | 0.492 |
| o3-high | 0.535 | 0.498 | 0.553 |
| o3-low | 0.522 | 0.499 | 0.549 |
| o4-mini-high | 0.562 | 0.554 | 0.588 |
| o4-mini-low | 0.529 | 0.496 | 0.576 |
| **Atlas** | | | |
| atlas-ei | 0.552 | 0.531 | 0.565 |
| atlas-ei-des | 0.427 | 0.407 | 0.446 |
| atlas-pi | 0.548 | 0.522 | 0.580 |
| atlas-pi-des | 0.424 | 0.408 | 0.434 |
| atlas-ucb | 0.489 | 0.416 | 0.541 |
| atlas-ucb-des | 0.417 | 0.411 | 0.425 |

### Chan-Lam

| Method | Median | CI Lower | CI Upper |
|---|---|---|---|
| **Anthropic** | | | |
| claude-3-7-sonnet | 0.586 | 0.543 | 0.633 |
| claude-3-7-sonnet-thinking | 0.587 | 0.551 | 0.640 |
| claude-opus-4 | 0.555 | 0.531 | 0.595 |
| claude-opus-4-1 | 0.581 | 0.571 | 0.607 |
| claude-sonnet-4 | 0.565 | 0.500 | 0.594 |
| claude-sonnet-4-thinking | 0.563 | 0.523 | 0.596 |
| **Google** | | | |
| gemini-2.0-flash | 0.457 | 0.381 | 0.528 |
| gemini-2.5-flash | 0.595 | 0.564 | 0.659 |
| gemini-2.5-flash-lite | 0.443 | 0.410 | 0.522 |
| gemini-2.5-flash-medium | 0.587 | 0.534 | 0.603 |
| gemini-2.5-pro | 0.630 | 0.608 | 0.675 |
| gemini-2.5-pro-medium | 0.625 | 0.608 | 0.666 |
| **OpenAI** | | | |
| gpt-5 | 0.582 | 0.489 | 0.605 |
| gpt-5-mini | 0.384 | 0.307 | 0.457 |
| o3-high | 0.558 | 0.504 | 0.581 |
| o3-low | 0.528 | 0.502 | 0.567 |
| o4-mini-high | 0.567 | 0.535 | 0.594 |
| o4-mini-low | 0.584 | 0.558 | 0.625 |
| **Atlas** | | | |
| atlas-ei | 0.519 | 0.436 | 0.538 |
| atlas-ei-des | 0.394 | 0.377 | 0.472 |
| atlas-pi | 0.544 | 0.505 | 0.563 |
| atlas-pi-des | 0.387 | 0.368 | 0.406 |
| atlas-ucb | 0.597 | 0.564 | 0.643 |
| atlas-ucb-des | 0.389 | 0.359 | 0.408 |

### Buchwald-Hartwig

| Method | Median | CI Lower | CI Upper |
|---|---|---|---|
| **Anthropic** | | | |
| claude-3-7-sonnet | 0.479 | 0.449 | 0.516 |
| claude-3-7-sonnet-thinking | 0.450 | 0.436 | 0.457 |
| claude-opus-4 | 0.497 | 0.454 | 0.514 |
| claude-opus-4-1 | 0.525 | 0.503 | 0.535 |
| claude-sonnet-4 | 0.492 | 0.464 | 0.520 |
| claude-sonnet-4-thinking | 0.488 | 0.471 | 0.504 |
| **Google** | | | |
| gemini-2.0-flash | 0.455 | 0.394 | 0.489 |
| gemini-2.5-flash | 0.490 | 0.445 | 0.512 |
| gemini-2.5-flash-lite | 0.379 | 0.326 | 0.405 |
| gemini-2.5-flash-medium | 0.466 | 0.417 | 0.522 |
| gemini-2.5-pro | 0.495 | 0.477 | 0.532 |
| gemini-2.5-pro-medium | 0.477 | 0.465 | 0.494 |
| **OpenAI** | | | |
| gpt-5 | 0.556 | 0.524 | 0.567 |
| gpt-5-mini | 0.533 | 0.510 | 0.546 |
| o3-high | 0.554 | 0.537 | 0.573 |
| o3-low | 0.555 | 0.534 | 0.572 |
| o4-mini-high | 0.501 | 0.490 | 0.542 |
| o4-mini-low | 0.539 | 0.524 | 0.554 |
| **Atlas** | | | |
| atlas-ei | 0.499 | 0.352 | 0.567 |
| atlas-ei-des | 0.449 | 0.398 | 0.504 |
| atlas-pi | 0.251 | 0.236 | 0.526 |
| atlas-pi-des | 0.474 | 0.434 | 0.498 |
| atlas-ucb | 0.572 | 0.528 | 0.609 |
| atlas-ucb-des | 0.422 | 0.393 | 0.470 |

**Figure S8 — Bootstrapped CIs for median cumulative entropy.** For each dataset-method combination, the confidence interval lower and upper bounds are shown for median cumulative entropy. Text with more red indicates more exploratory methods, while text with more blue indicates more exploitative methods.

## F. Performance-Cumulative Entropy Correlation Analysis for all Methods

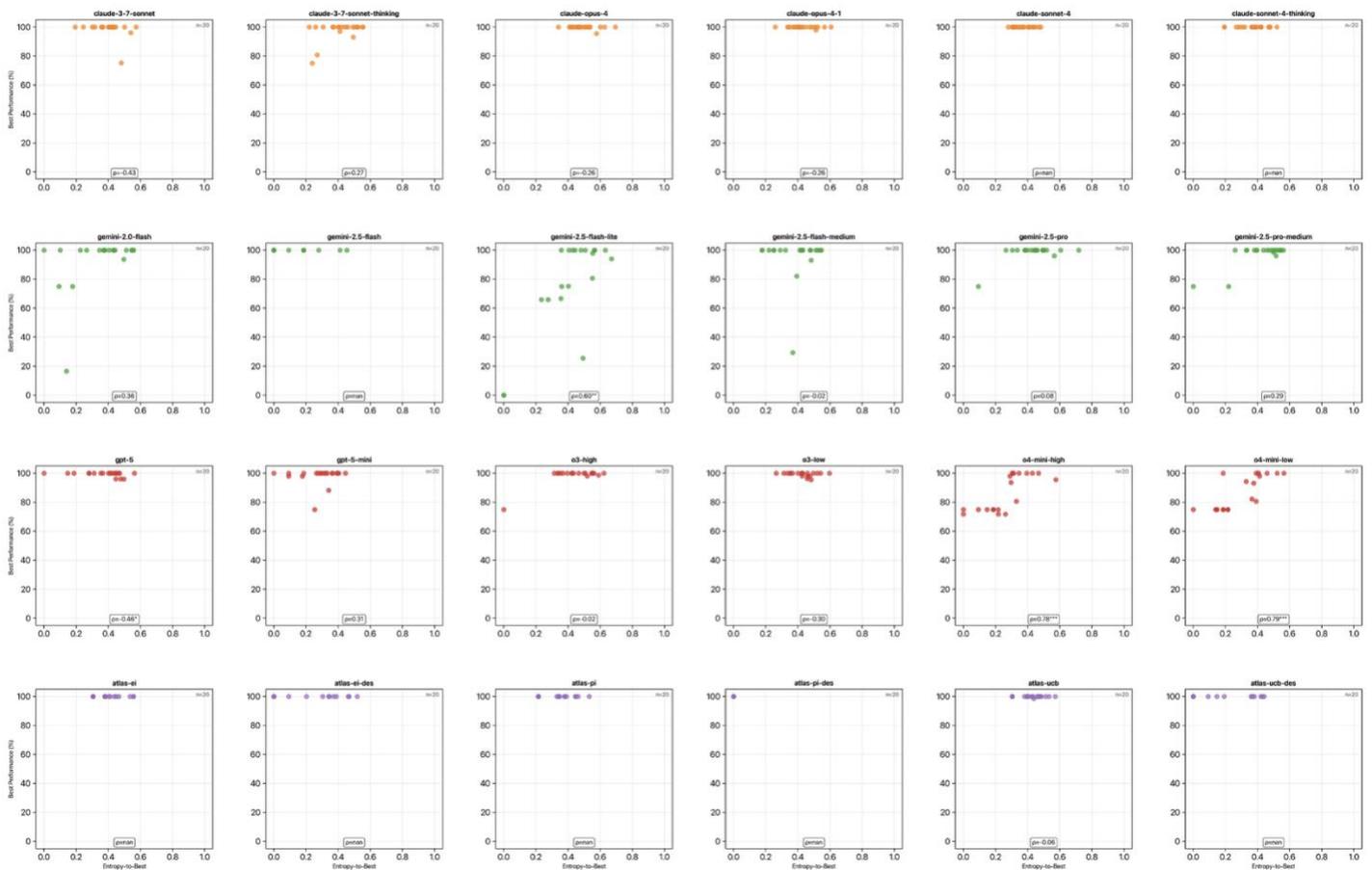

**Figure S9A — Cumulative entropy-to-best vs. performance for Suzuki Conversion.** For each method, the cumulative entropy-to-best observation is plotted on the x-axis, with the best performance plotted on the y-axis.

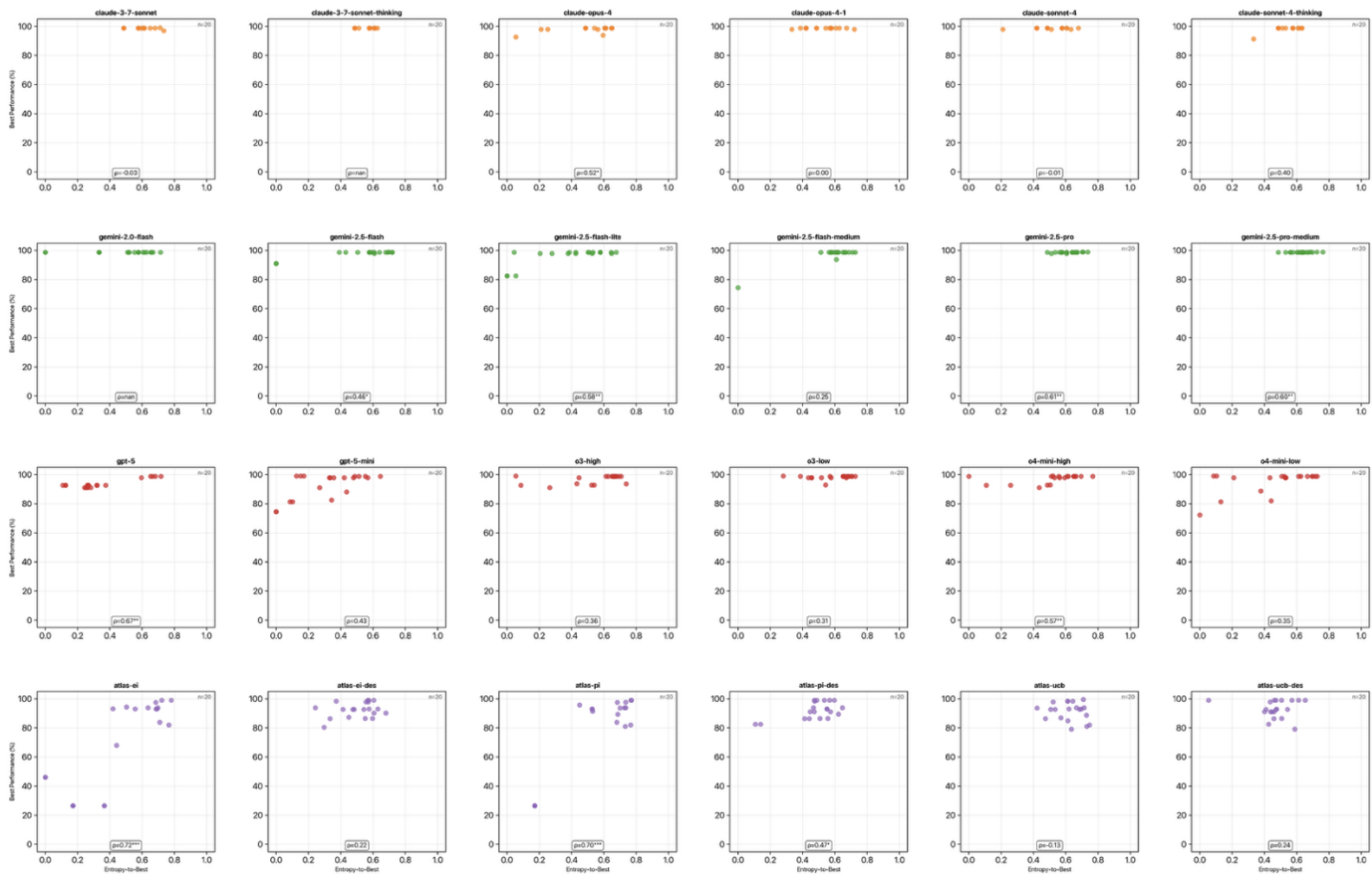

**Figure S9B — Cumulative entropy-to-best vs. performance for Amide Coupling HTE.** For each method, the cumulative entropy-to-best observation is plotted on the x-axis, with the best performance plotted on the y-axis.

# Reductive Amination
## Method-Specific Cumulative Entropy-to-best vs. Performance Correlation

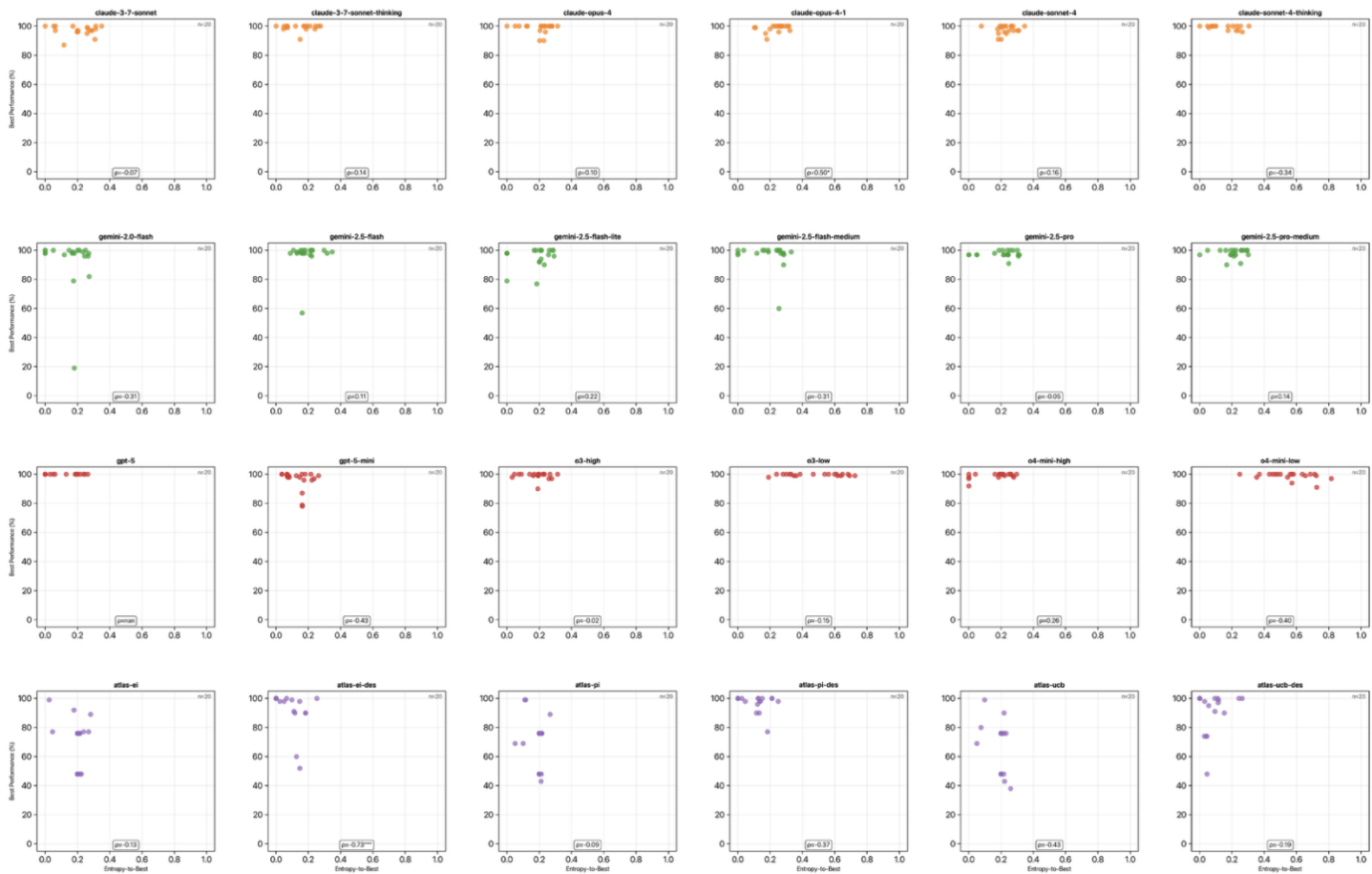

**Figure S9C — Cumulative entropy-to-best vs. performance for Reductive Amination.** For each method, the cumulative entropy-to-best observation is plotted on the x-axis, with the best performance plotted on the y-axis.

# Suzuki Yield
## Method-Specific Cumulative Entropy-to-best vs. Performance Correlation

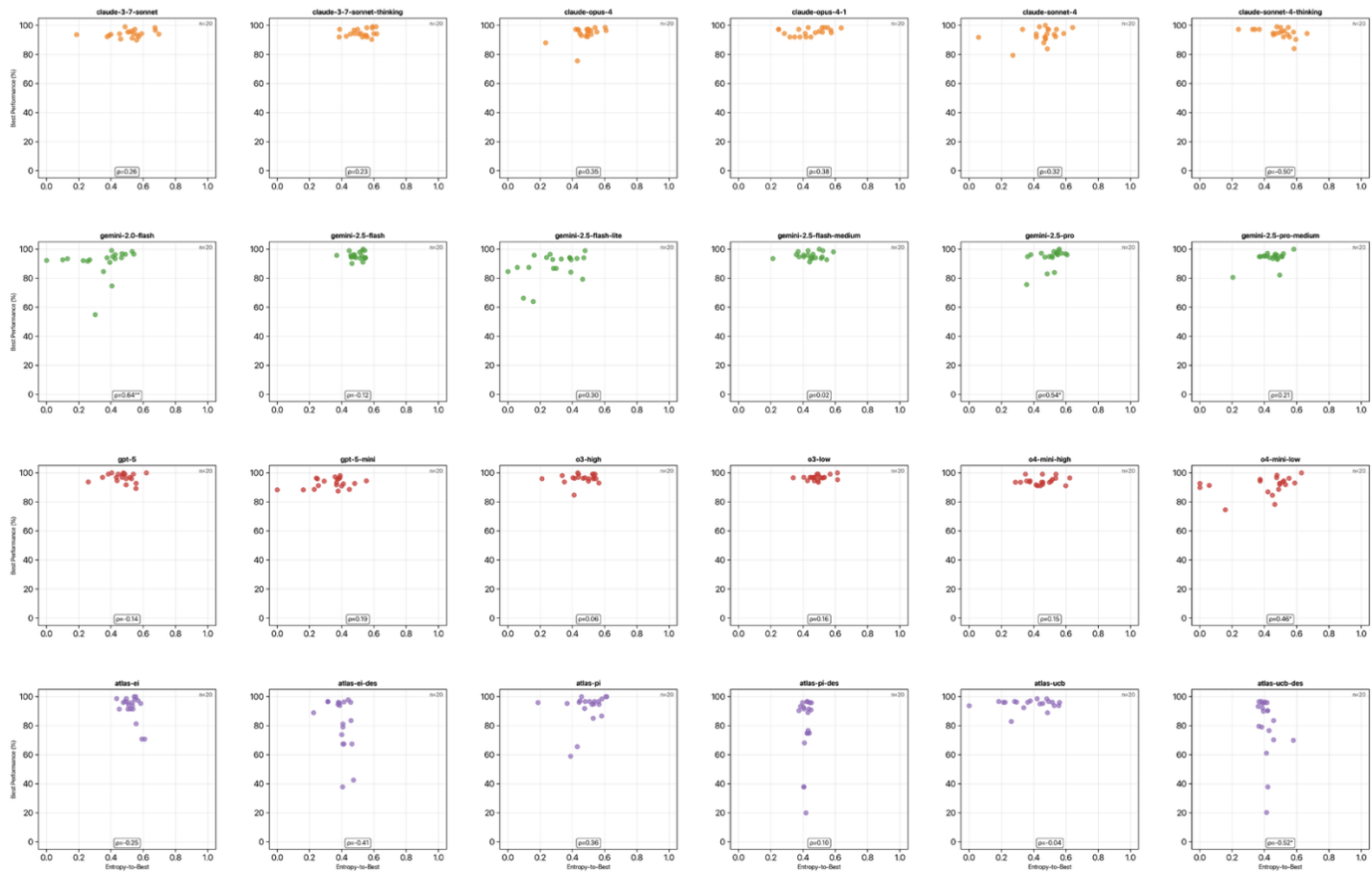

**Figure S9D — Cumulative entropy-to-best vs. performance for Suzuki Yield.** For each method, the cumulative entropy-to-best observation is plotted on the x-axis, with the best performance plotted on the y-axis.

# Chan-Lam
## Method-Specific Cumulative Entropy-to-best vs. Performance Correlation

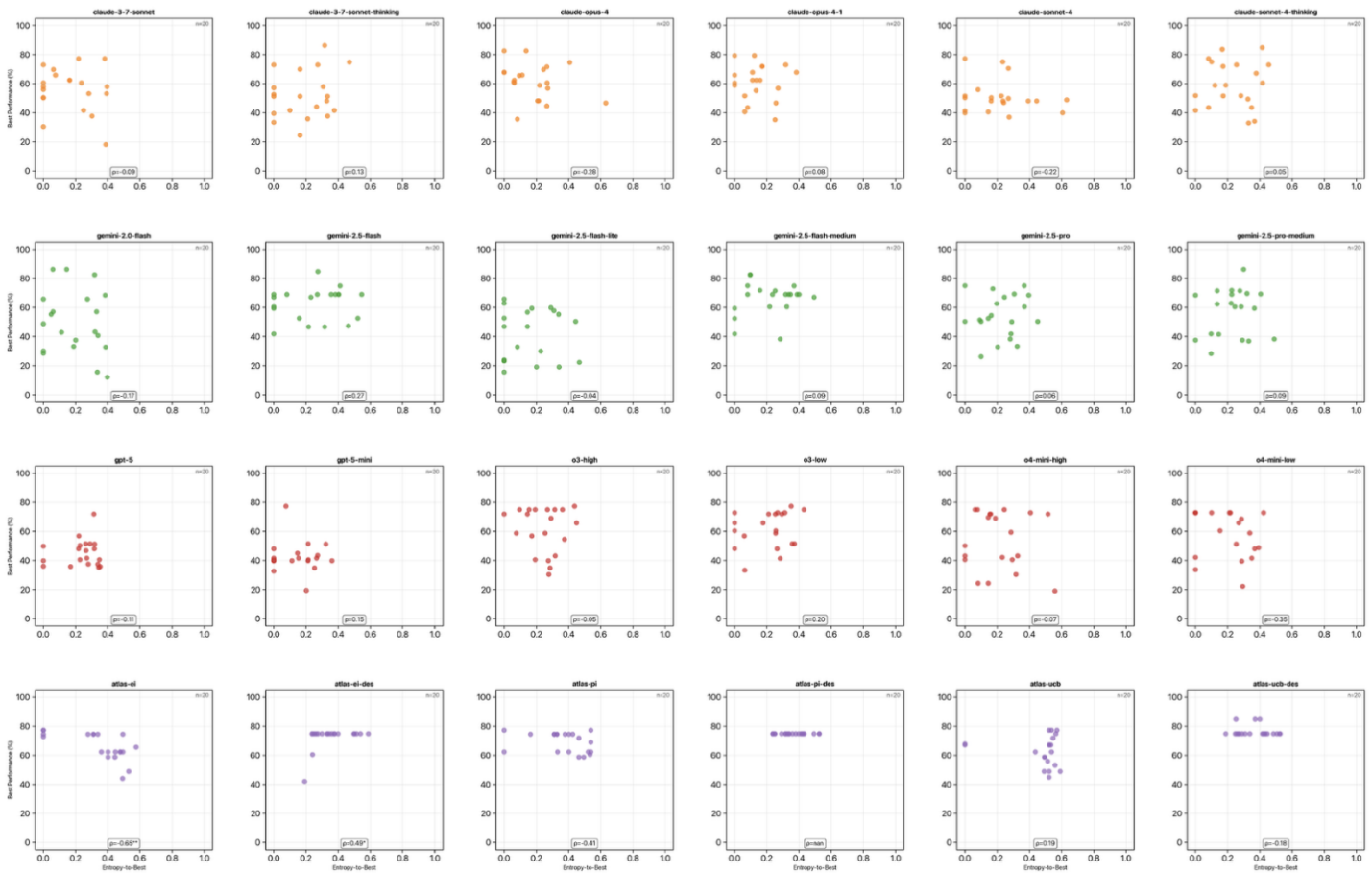

**Figure S9E — Cumulative entropy-to-best vs. performance for Chan-Lam.** For each method, the cumulative entropy-to-best observation is plotted on the x-axis, with the best performance plotted on the y-axis.

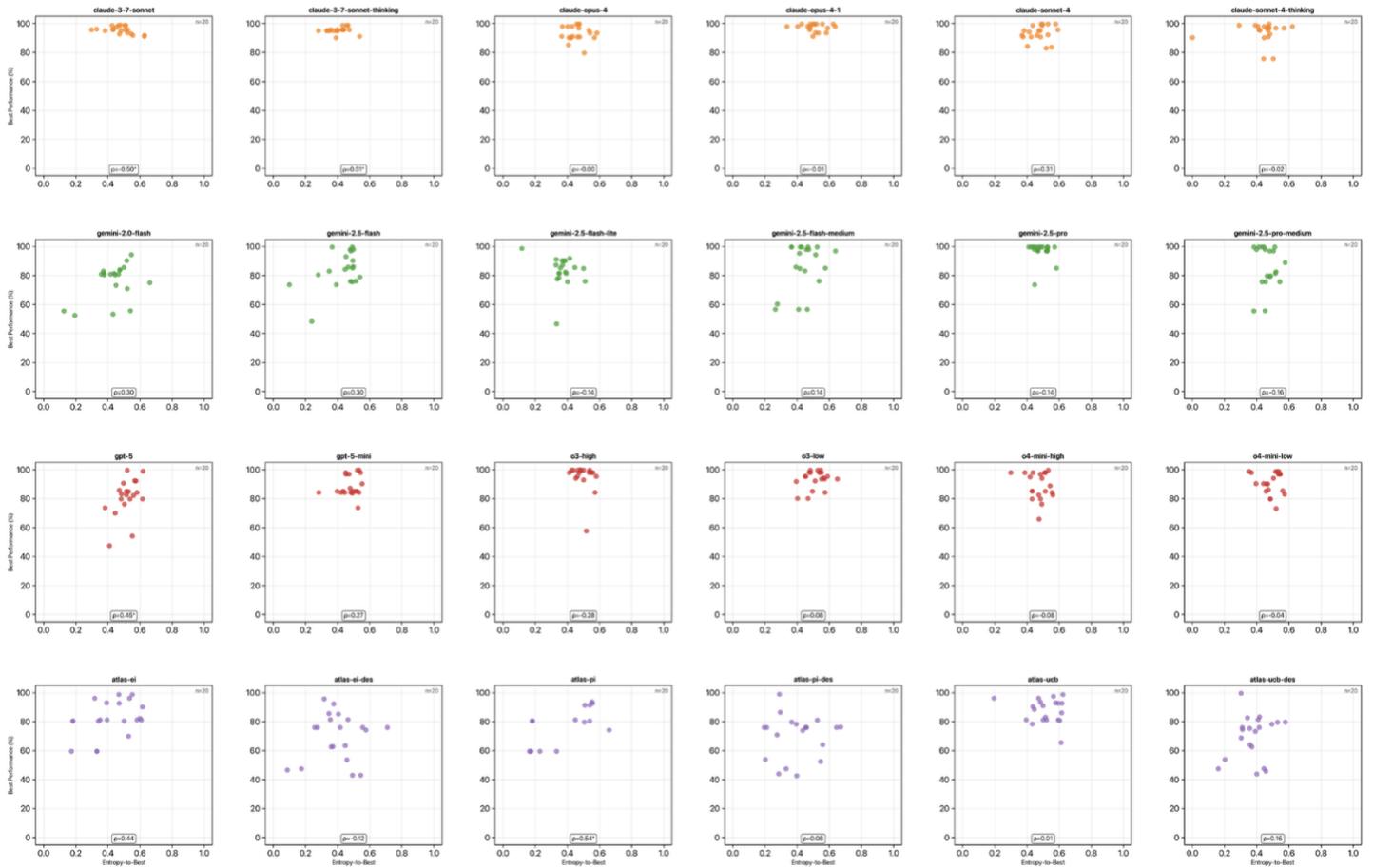

**Figure S9F — Cumulative entropy-to-best vs. performance for Buchwald-Hartwig.** For each method, the cumulative entropy-to-best observation is plotted on the x-axis, with the best performance plotted on the y-axis.

# G. Per-Parameter Cumulative Entropy Analysis

## Suzuki Conversion

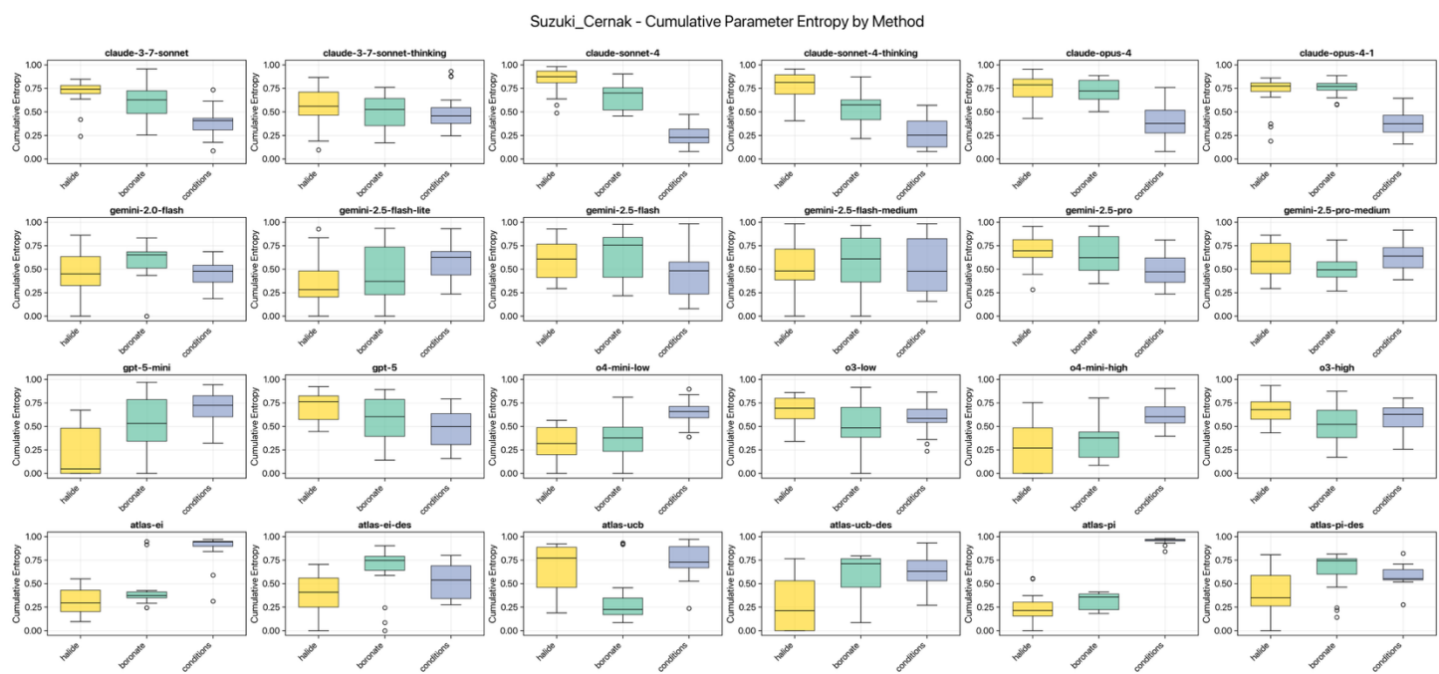

**Figure S10A — Per-parameter cumulative entropy for Suzuki Conversion.** For each method, the distribution of cumulative entropies across all runs is shown for each individual parameter.

## Amide Coupling HTE

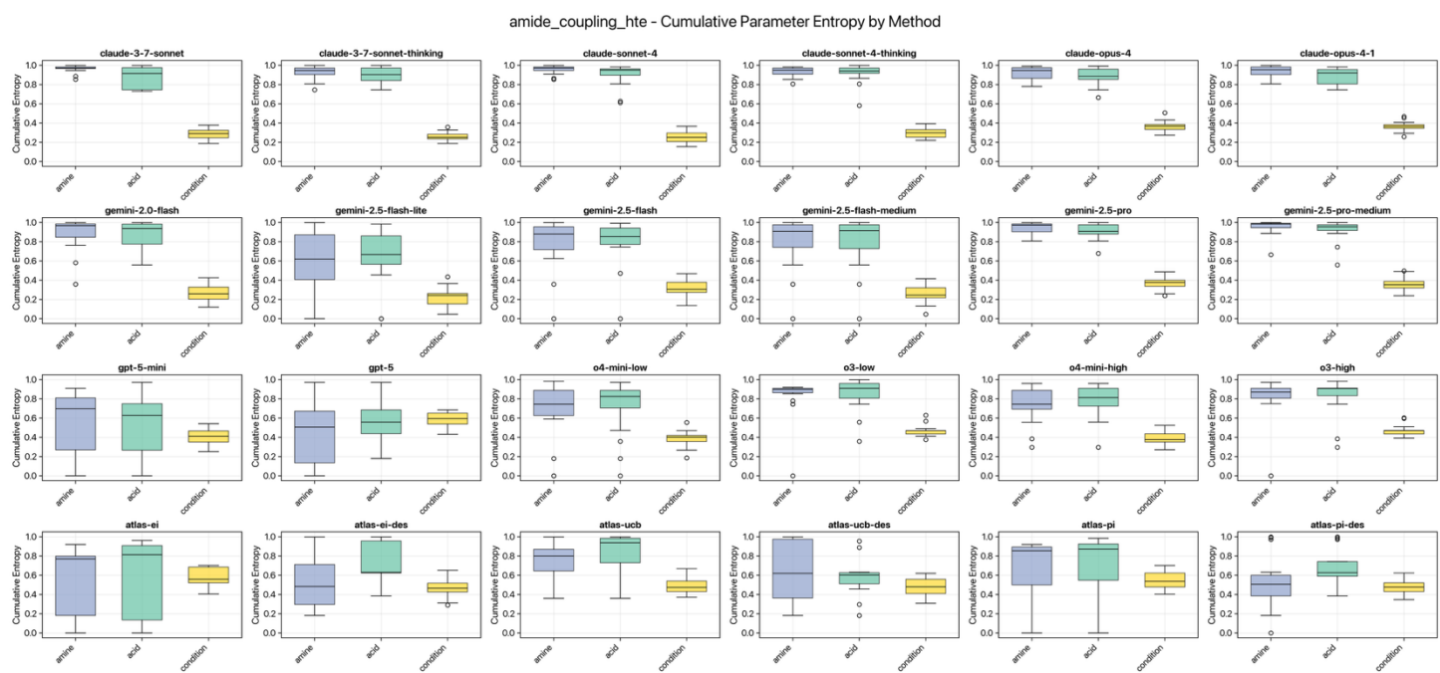

**Figure S10B — Per-parameter cumulative entropy for Amide Coupling HTE.** For each method, the distribution of cumulative entropies across all runs is shown for each individual parameter.

# Reductive Amination

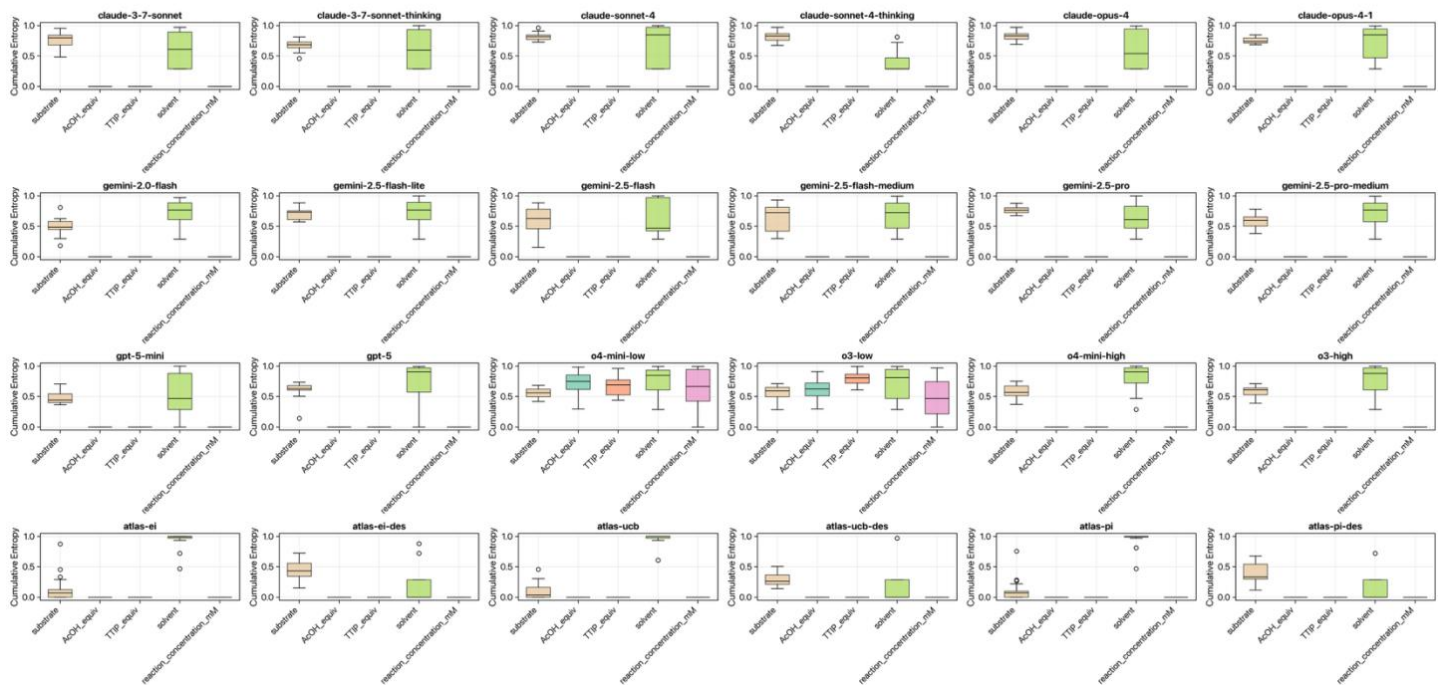

**Figure S10C — Per-parameter cumulative entropy for Reductive Amination.** For each method, the distribution of cumulative entropies across all runs is shown for each individual parameter.

# Suzuki Yield

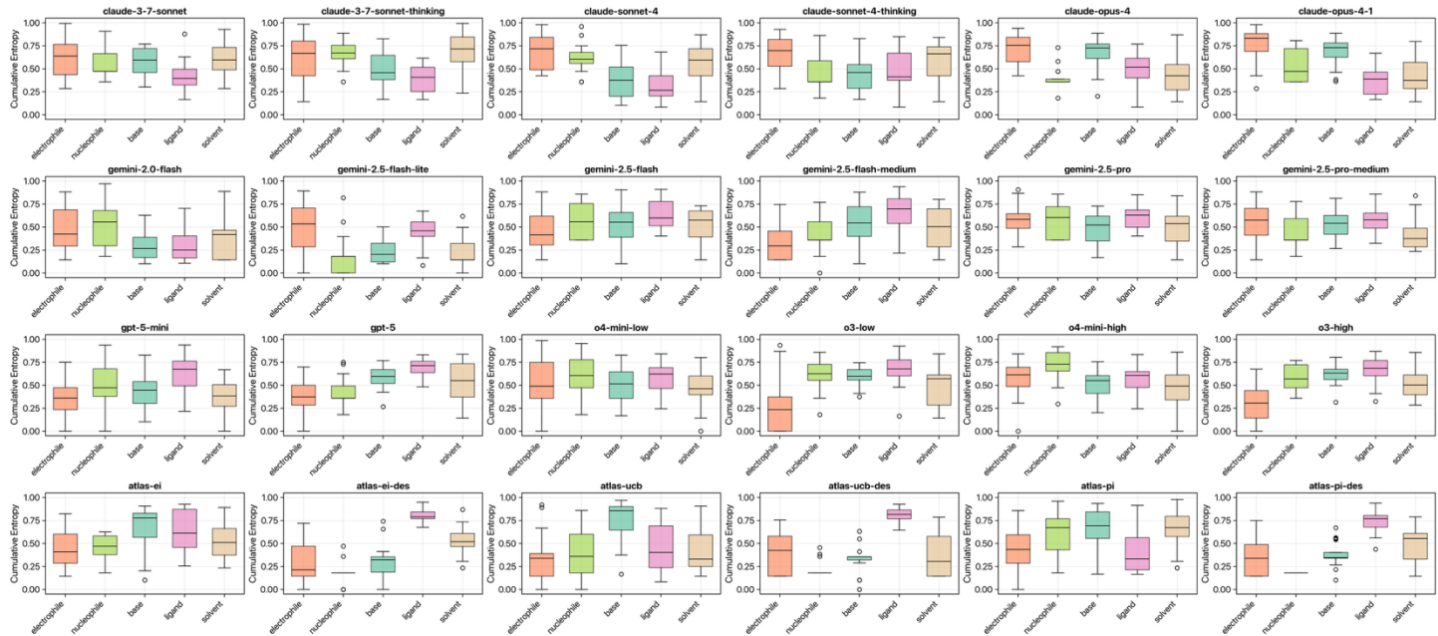

**Figure S10D — Per-parameter cumulative entropy for Suzuki Yield.** For each method, the distribution of cumulative entropies across all runs is shown for each individual parameter.

# Chan-Lam

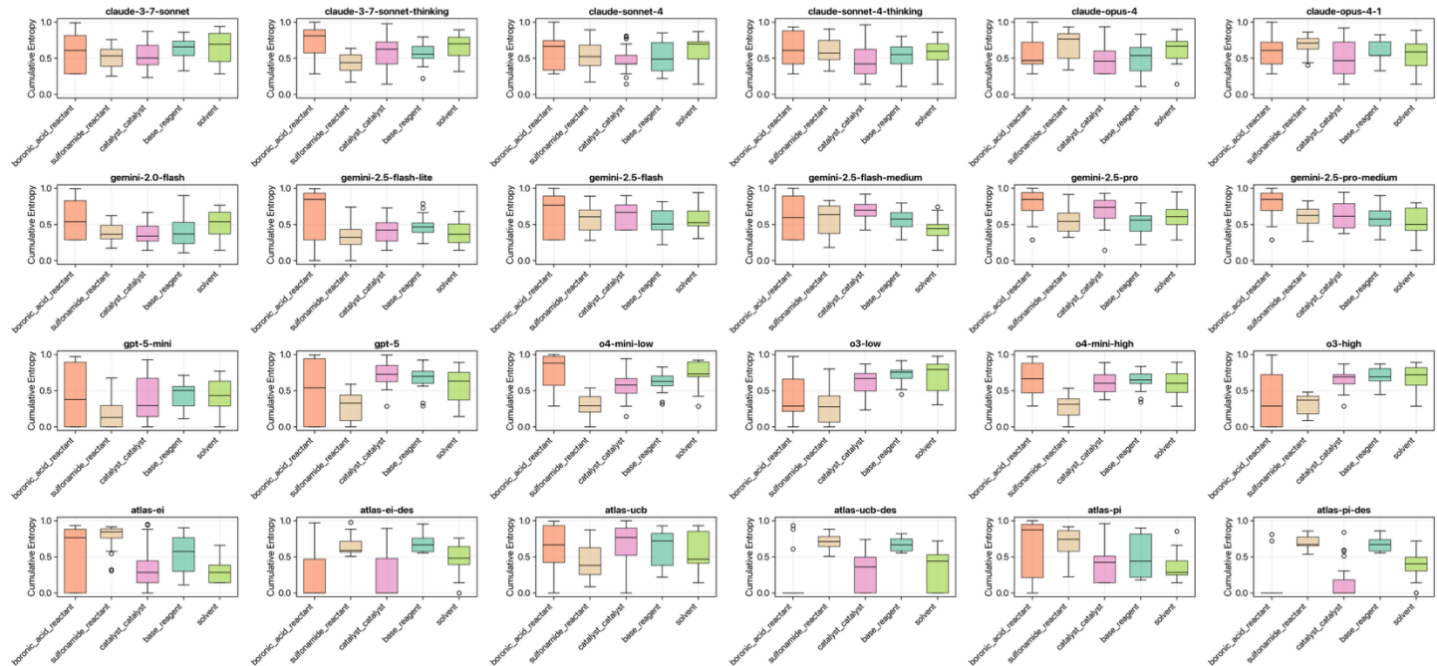

**Figure S10E — Per-parameter cumulative entropy for Chan-Lam.** For each method, the distribution of cumulative entropies across all runs is shown for each individual parameter.

# Buchwald-Hartwig

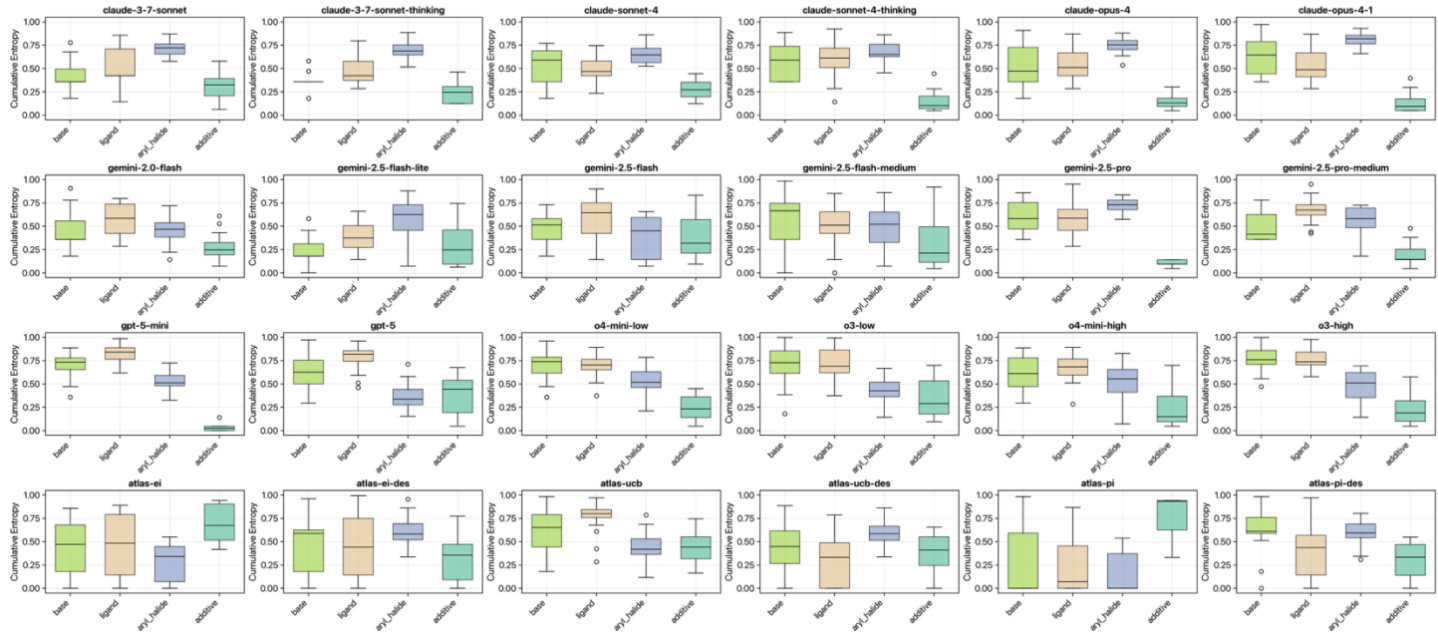

**Figure S10F — Per-parameter cumulative entropy for Buchwald-Hartwig.** For each method, the distribution of cumulative entropies across all runs is shown for each individual parameter.

# H. Table of All Evaluated Large Language Models

**Table S1 — Mapping between model names used in figures and API names used.**

| Model | API Name |
|---|---|
| claude-3-7-sonnet | claude-3-7-sonnet-latest |
| claude-3-7-sonnet-thinking | claude-3-7-sonnet-latest |
| claude-sonnet-4 | claude-sonnet-4-20250514 |
| claude-sonnet-4-thinking | claude-sonnet-4-20250514 |
| claude-opus-4 | claude-opus-4-20250514 |
| claude-opus-4-1 | claude-opus-4-20250805 |
| gemini-2.0-flash | gemini-2.0-flash |
| gemini-2.5-flash-lite | gemini-2.5-flash-lite-preview-06-17 |
| gemini-2.5-flash | gemini-2.5-flash-preview-04-17 |
| gemini-2.5-flash-medium | gemini-2.5-flash-preview-04-17 |
| gemini-2.5-pro | gemini-2.5-pro-preview-03-25 |
| gemini-2.5-pro-medium | gemini-2.5-preview-03-25 |
| gpt-5-mini | gpt-5-mini |
| gpt-5 | gpt-5 |
| o4-mini-low | o4-mini |
| o3-low | o3 |
| o4-mini-high | o4-mini |
| o3-high | o3 |

I. Invalid Suggestion Rates for LLM-GO Methods

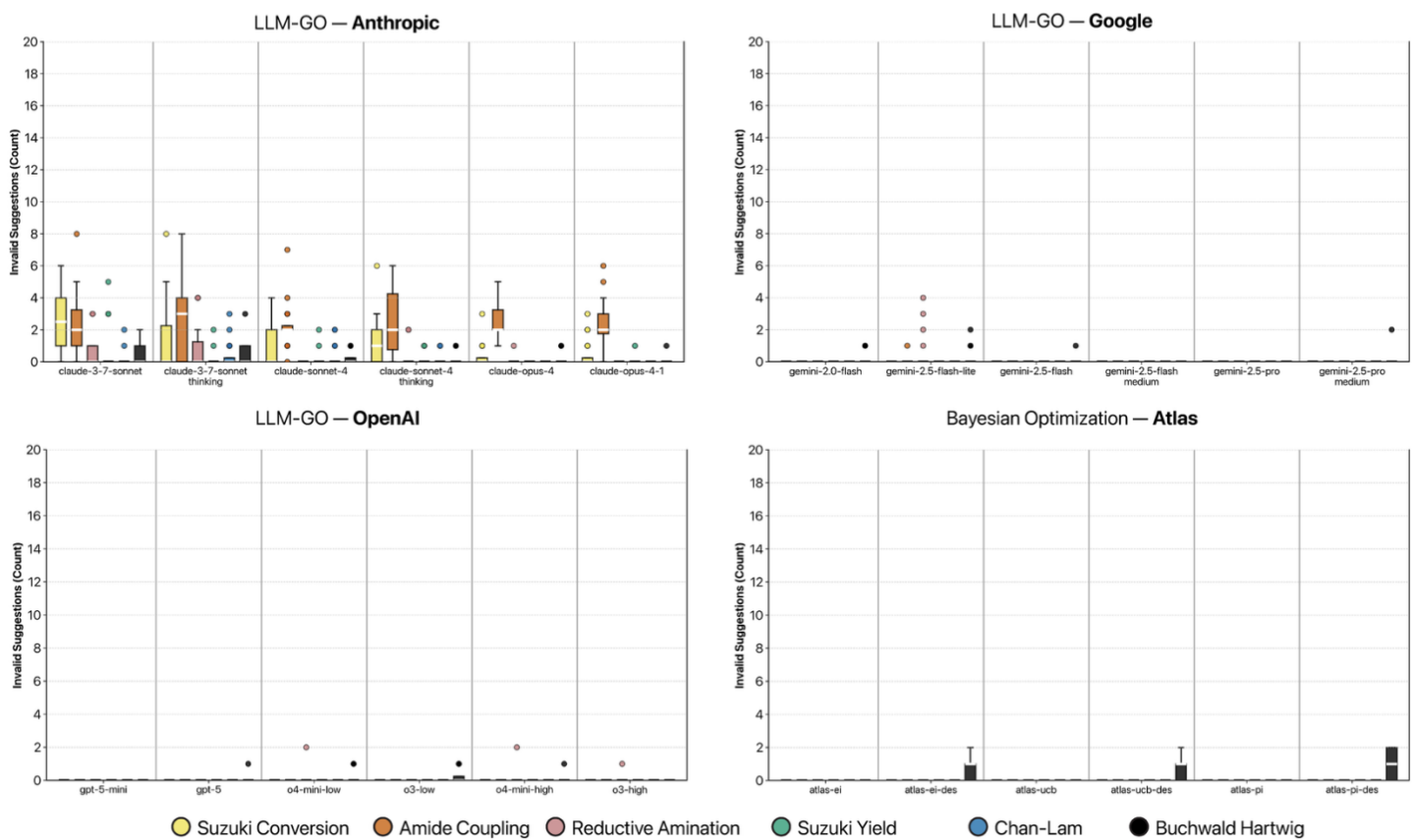

**Figure S11 — Invalid suggestion rates grouped by provider.** For each method-dataset pair, the distribution of invalid suggestions (objective value(s) is not available) across 20 independent campaigns is shown.

## J. Baseline Methods

### Random Sampling Baseline

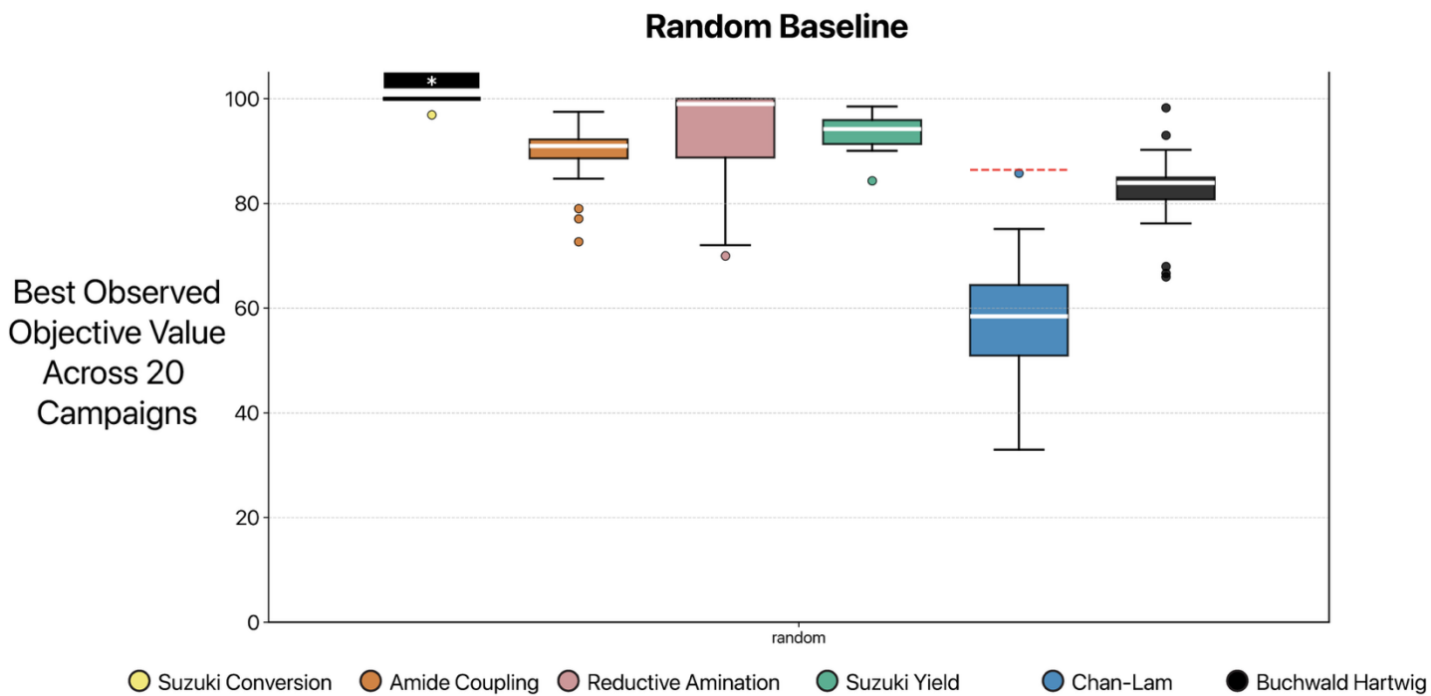

**Figure S12 — Random sampling baseline performance.** Boxplots (**n=20**) display the distribution of best objective values achieved via random sampling across 20 independent optimization campaigns for each dataset. Colors are used to represent the six different optimization datasets. Asterisks (*) indicate methods where the median equals the maximum objective value and the interquartile range is zero. Objective values are the raw values from each dataset's objective function. Individual points beyond the whiskers indicate outliers in the performance distribution.

# Statistical Significance (**p-values**) and Effect Sizes (**Cliff's Delta**) for LLM/BO vs. Random Performance

**Suzuki Conversion - Methods vs Random Baseline Comparison**
**Statistical Significance and Effect Sizes**

| Method | Median % | p-value | Effect Size (δ) | Practical Sig. | vs Random |
|---|---|---|---|---|---|
| **Anthropic** | | | | | |
| claude-sonnet-4 | 100.0% | 0.342ns | 0.050 | Negligible | 0.0% |
| claude-3-7-sonnet | 100.0% | 0.534ns | -0.055 | Negligible | 0.0% |
| claude-3-7-sonnet-thinking | 100.0% | 0.151ns | -0.155 | Small | 0.0% |
| claude-opus-4 | 100.0% | 1.000ns | -0.003 | Negligible | 0.0% |
| claude-sonnet-4-thinking | 100.0% | 0.342ns | 0.050 | Negligible | 0.0% |
| claude-opus-4-1 | 100.0% | 1.000ns | 0.003 | Negligible | 0.0% |
| **Google** | | | | | |
| gemini-2.5-flash-medium | 100.0% | 0.275ns | -0.107 | Negligible | 0.0% |
| gemini-2.5-pro-medium | 100.0% | 0.151ns | -0.155 | Small | 0.0% |
| gemini-2.5-pro | 100.0% | 0.534ns | -0.055 | Negligible | 0.0% |
| gemini-2.0-flash | 100.0% | 0.138ns | -0.160 | Small | 0.0% |
| gemini-2.5-flash | 100.0% | 0.342ns | 0.050 | Negligible | 0.0% |
| gemini-2.5-flash-lite | 96.0% | 5.19e-04*** | -0.522 | Large | -4.0% |
| **OpenAI** | | | | | |
| o3-high | 100.0% | 0.324ns | -0.098 | Negligible | 0.0% |
| gpt-5-mini | 100.0% | 0.165ns | -0.150 | Small | 0.0% |
| o3-low | 100.0% | 0.089ns | -0.198 | Small | 0.0% |
| gpt-5 | 100.0% | 0.275ns | -0.107 | Negligible | 0.0% |
| o4-mini-low | 87.6% | 6.58e-05*** | -0.627 | Large | -12.4% |
| o4-mini-high | 87.2% | 6.91e-05*** | -0.627 | Large | -12.8% |
| **Atlas** | | | | | |
| atlas-ei-des | 100.0% | 0.342ns | 0.050 | Negligible | 0.0% |
| atlas-ei | 100.0% | 0.342ns | 0.050 | Negligible | 0.0% |
| atlas-ucb | 100.0% | 1.000ns | 0.003 | Negligible | 0.0% |
| atlas-pi-des | 100.0% | 0.342ns | 0.050 | Negligible | 0.0% |
| atlas-pi | 100.0% | 0.342ns | 0.050 | Negligible | 0.0% |
| atlas-ucb-des | 100.0% | 0.342ns | 0.050 | Negligible | 0.0% |

## Amide Coupling Hte - Methods vs Random Baseline Comparison
### Statistical Significance and Effect Sizes

| Method | Median % | p-value | Effect Size (δ) | Practical Sig. | vs Random |
|---|---|---|---|---|---|
| **Anthropic** | | | | | |
| claude-sonnet-4 | 98.8% | 1.94e-08*** | 1.000 | Large | +7.8% |
| claude-3-7-sonnet | 98.8% | 2.08e-08*** | 0.990 | Large | +7.8% |
| claude-3-7-sonnet-thinking | 98.8% | 7.96e-09*** | 1.000 | Large | +7.8% |
| claude-opus-4 | 98.8% | 1.18e-07*** | 0.955 | Large | +7.8% |
| claude-sonnet-4-thinking | 98.8% | 4.18e-08*** | 0.960 | Large | +7.8% |
| claude-opus-4-1 | 98.8% | 1.50e-08*** | 1.000 | Large | +7.8% |
| **Google** | | | | | |
| gemini-2.5-pro-medium | 98.8% | 1.94e-08*** | 1.000 | Large | +7.8% |
| gemini-2.5-flash-lite | 98.8% | 4.76e-05*** | 0.745 | Large | +7.8% |
| gemini-2.5-pro | 98.8% | 2.92e-08*** | 1.000 | Large | +7.8% |
| gemini-2.5-flash-medium | 98.8% | 5.50e-07*** | 0.885 | Large | +7.8% |
| gemini-2.5-flash | 98.8% | 4.47e-07*** | 0.905 | Large | +7.8% |
| gemini-2.0-flash | 98.8% | 7.96e-09*** | 1.000 | Large | +7.8% |
| **OpenAI** | | | | | |
| o4-mini-high | 98.8% | 3.34e-06*** | 0.853 | Large | +7.8% |
| o3-high | 98.8% | 4.64e-06*** | 0.838 | Large | +7.8% |
| o4-mini-low | 98.8% | 3.79e-04*** | 0.655 | Large | +7.8% |
| o3-low | 98.8% | 9.14e-08*** | 0.975 | Large | +7.8% |
| gpt-5-mini | 97.9% | 0.020* | 0.432 | Medium | +6.9% |
| gpt-5 | 92.7% | 5.32e-03** | 0.512 | Large | +1.7% |
| **Atlas** | | | | | |
| atlas-ucb | 92.8% | 0.172ns | 0.255 | Small | +1.8% |
| atlas-ucb-des | 92.7% | 0.039* | 0.383 | Medium | +1.7% |
| atlas-ei-des | 92.7% | 0.090ns | 0.315 | Small | +1.7% |
| atlas-pi | 92.2% | 0.957ns | 0.013 | Negligible | +1.2% |
| atlas-pi-des | 92.0% | 0.244ns | 0.217 | Small | +1.0% |
| atlas-ei | 88.5% | 0.524ns | -0.120 | Negligible | -2.5% |

**Suzuki Yield - Methods vs Random Baseline Comparison**
Statistical Significance and Effect Sizes

| Method | Median % | p-value | Effect Size (δ) | Practical Sig. | vs Random |
|---|---|---|---|---|---|
| **Anthropic** | | | | | |
| claude-opus-4 | 96.3% | 0.140ns | 0.275 | Small | +2.1% |
| claude-opus-4-1 | 95.6% | 0.039* | 0.383 | Medium | +1.4% |
| claude-sonnet-4-thinking | 95.2% | 0.133ns | 0.280 | Small | +1.0% |
| claude-3-7-sonnet-thinking | 94.3% | 0.323ns | 0.185 | Small | +0.1% |
| claude-3-7-sonnet | 94.1% | 0.695ns | 0.075 | Negligible | -0.1% |
| claude-sonnet-4 | 94.0% | 0.946ns | 0.015 | Negligible | -0.2% |
| **Google** | | | | | |
| gemini-2.5-pro | 96.2% | 0.056ns | 0.355 | Medium | +2.0% |
| gemini-2.5-pro-medium | 95.3% | 0.194ns | 0.242 | Small | +1.1% |
| gemini-2.5-flash | 95.2% | 0.083ns | 0.323 | Small | +1.0% |
| gemini-2.5-flash-medium | 94.8% | 0.064ns | 0.345 | Medium | +0.6% |
| gemini-2.0-flash | 93.5% | 0.968ns | -0.010 | Negligible | -0.7% |
| gemini-2.5-flash-lite | 92.9% | 0.068ns | -0.340 | Medium | -1.3% |
| **OpenAI** | | | | | |
| o3-low | 97.0% | 1.03e-03** | 0.608 | Large | +2.8% |
| gpt-5 | 97.0% | 1.74e-03** | 0.580 | Large | +2.8% |
| o3-high | 96.5% | 5.76e-03** | 0.512 | Large | +2.3% |
| o4-mini-high | 93.7% | 0.645ns | 0.087 | Negligible | -0.5% |
| gpt-5-mini | 93.0% | 0.552ns | -0.113 | Negligible | -1.2% |
| o4-mini-low | 92.8% | 0.358ns | -0.172 | Small | -1.3% |
| **Atlas** | | | | | |
| atlas-ucb | 96.2% | 0.120ns | 0.290 | Small | +2.0% |
| atlas-pi | 96.1% | 0.163ns | 0.260 | Small | +1.9% |
| atlas-ei | 95.7% | 0.291ns | 0.198 | Small | +1.5% |
| atlas-ei-des | 91.5% | 0.159ns | -0.263 | Small | -2.7% |
| atlas-pi-des | 90.7% | 0.024* | -0.420 | Medium | -3.5% |
| atlas-ucb-des | 90.1% | 0.021* | -0.427 | Medium | -4.1% |

**Reductive Amination - Methods vs Random Baseline Comparison**
**Statistical Significance and Effect Sizes**

| Method | Median % | p-value | Effect Size (δ) | Practical Sig. | vs Random |
|---|---|---|---|---|---|
| **Anthropic** | | | | | |
| claude-sonnet-4-thinking | 100.0% | 0.035* | 0.347 | Medium | +1.0% |
| claude-3-7-sonnet-thinking | 100.0% | 0.041* | 0.343 | Medium | +1.0% |
| claude-opus-4-1 | 100.0% | 0.106ns | 0.275 | Small | +1.0% |
| claude-opus-4 | 100.0% | 0.026* | 0.360 | Medium | +1.0% |
| claude-sonnet-4 | 98.5% | 0.659ns | 0.080 | Negligible | -0.5% |
| claude-3-7-sonnet | 97.0% | 0.845ns | -0.037 | Negligible | -2.0% |
| **Google** | | | | | |
| gemini-2.5-pro-medium | 100.0% | 0.369ns | 0.158 | Small | +1.0% |
| gemini-2.5-flash | 99.0% | 0.621ns | 0.090 | Negligible | 0.0% |
| gemini-2.5-flash-medium | 99.0% | 0.550ns | 0.107 | Negligible | 0.0% |
| gemini-2.0-flash | 98.0% | 0.933ns | -0.018 | Negligible | -1.0% |
| gemini-2.5-flash-lite | 98.0% | 0.989ns | -0.005 | Negligible | -1.0% |
| gemini-2.5-pro | 97.0% | 0.799ns | -0.048 | Negligible | -2.0% |
| **OpenAI** | | | | | |
| o3-low | 100.0% | 0.039* | 0.350 | Medium | +1.0% |
| o3-high | 100.0% | 0.080ns | 0.297 | Small | +1.0% |
| gpt-5 | 100.0% | 1.67e-04*** | 0.550 | Large | +1.0% |
| o4-mini-low | 100.0% | 0.091ns | 0.287 | Small | +1.0% |
| o4-mini-high | 100.0% | 0.078ns | 0.300 | Small | +1.0% |
| gpt-5-mini | 99.0% | 0.955ns | 0.013 | Negligible | 0.0% |
| **Atlas** | | | | | |
| atlas-pi-des | 100.0% | 0.260ns | 0.195 | Small | +1.0% |
| atlas-ei-des | 99.5% | 0.720ns | 0.065 | Negligible | +0.5% |
| atlas-ucb-des | 99.5% | 0.920ns | 0.020 | Negligible | +0.5% |
| atlas-pi | 76.0% | 1.43e-05*** | -0.795 | Large | -23.0% |
| atlas-ei | 76.0% | 1.12e-05*** | -0.805 | Large | -23.0% |
| atlas-ucb | 72.5% | 5.23e-06*** | -0.835 | Large | -26.5% |

## Chan-Lam - Methods vs Random Baseline Comparison
### Statistical Significance and Effect Sizes

| Method | Median % | p-value | Effect Size (δ) | Practical Sig. | vs Random |
|---|---|---|---|---|---|
| **Anthropic** | | | | | |
| claude-opus-4-1 | 62.5% | 0.425ns | 0.150 | Small | +4.1% |
| claude-opus-4 | 61.8% | 0.323ns | 0.185 | Small | +3.5% |
| claude-sonnet-4-thinking | 59.0% | 0.797ns | 0.050 | Negligible | +0.6% |
| claude-3-7-sonnet | 57.9% | 0.882ns | -0.030 | Negligible | -0.5% |
| claude-3-7-sonnet-thinking | 51.4% | 0.159ns | -0.263 | Small | -7.0% |
| claude-sonnet-4 | 48.5% | 0.029* | -0.405 | Medium | -9.9% |
| **Google** | | | | | |
| gemini-2.5-flash-medium | 69.0% | 0.023* | 0.420 | Medium | +10.7% |
| gemini-2.5-flash | 68.1% | 0.187ns | 0.245 | Small | +9.7% |
| gemini-2.5-pro-medium | 61.5% | 0.818ns | 0.045 | Negligible | +3.1% |
| gemini-2.5-pro | 52.1% | 0.449ns | -0.142 | Negligible | -6.3% |
| gemini-2.5-flash-lite | 47.0% | 6.03e-03** | -0.510 | Large | -11.4% |
| gemini-2.0-flash | 46.1% | 0.126ns | -0.285 | Small | -12.3% |
| **OpenAI** | | | | | |
| o3-high | 67.5% | 0.278ns | 0.203 | Small | +9.1% |
| o3-low | 63.2% | 0.279ns | 0.203 | Small | +4.8% |
| o4-mini-low | 63.2% | 0.674ns | 0.080 | Negligible | +4.8% |
| o4-mini-high | 54.9% | 0.735ns | -0.065 | Negligible | -3.5% |
| gpt-5 | 44.3% | 9.19e-04*** | -0.615 | Large | -14.1% |
| gpt-5-mini | 40.6% | 1.01e-04*** | -0.720 | Large | -17.8% |
| **Atlas** | | | | | |
| atlas-ei-des | 74.9% | 1.02e-04*** | 0.688 | Large | +16.5% |
| atlas-ucb-des | 74.9% | 4.76e-06*** | 0.815 | Large | +16.5% |
| atlas-pi-des | 74.9% | 3.99e-06*** | 0.800 | Large | +16.5% |
| atlas-pi | 73.3% | 1.60e-03** | 0.583 | Large | +14.9% |
| atlas-ucb | 64.8% | 0.203ns | 0.237 | Small | +6.4% |
| atlas-ei | 64.1% | 0.015* | 0.450 | Medium | +5.7% |

**Buchwald-Hartwig - Methods vs Random Baseline Comparison**
**Statistical Significance and Effect Sizes**

| Method | Median % | p-value | Effect Size (δ) | Practical Sig. | vs Random |
|---|---|---|---|---|---|
| **Anthropic** | | | | | |
| claude-opus-4-1 | 98.0% | 3.75e-07*** | 0.938 | Large | +14.1% |
| claude-sonnet-4-thinking | 96.9% | 7.70e-05*** | 0.733 | Large | +13.0% |
| claude-3-7-sonnet | 95.8% | 1.18e-06*** | 0.900 | Large | +11.8% |
| claude-3-7-sonnet-thinking | 95.4% | 9.50e-07*** | 0.902 | Large | +11.5% |
| claude-sonnet-4 | 94.7% | 4.10e-05*** | 0.760 | Large | +10.8% |
| claude-opus-4 | 92.4% | 2.42e-05*** | 0.782 | Large | +8.4% |
| **Google** | | | | | |
| gemini-2.5-pro | 98.9% | 3.72e-06*** | 0.850 | Large | +14.9% |
| gemini-2.5-flash-medium | 94.9% | 0.054ns | 0.357 | Medium | +11.0% |
| gemini-2.5-pro-medium | 85.8% | 0.473ns | 0.135 | Negligible | +1.8% |
| gemini-2.5-flash-lite | 85.1% | 0.516ns | 0.122 | Negligible | +1.2% |
| gemini-2.5-flash | 84.8% | 0.490ns | 0.130 | Negligible | +0.8% |
| gemini-2.0-flash | 81.0% | 0.041* | -0.380 | Medium | -2.9% |
| **OpenAI** | | | | | |
| o3-high | 98.0% | 1.04e-05*** | 0.815 | Large | +14.1% |
| o3-low | 94.8% | 1.08e-04*** | 0.718 | Large | +10.9% |
| o4-mini-low | 90.4% | 2.54e-03** | 0.560 | Large | +6.4% |
| o4-mini-high | 87.1% | 0.053ns | 0.360 | Medium | +3.2% |
| gpt-5-mini | 85.2% | 5.98e-03** | 0.510 | Large | +1.3% |
| gpt-5 | 83.0% | 0.935ns | -0.018 | Negligible | -1.0% |
| **Atlas** | | | | | |
| atlas-ucb | 89.4% | 0.155ns | 0.265 | Small | +5.5% |
| atlas-ei | 81.2% | 0.616ns | -0.095 | Negligible | -2.7% |
| atlas-pi | 80.5% | 0.046* | -0.370 | Medium | -3.4% |
| atlas-pi-des | 76.0% | 9.95e-04*** | -0.610 | Large | -8.0% |
| atlas-ei-des | 76.0% | 4.66e-03** | -0.525 | Large | -8.0% |
| atlas-ucb-des | 75.0% | 4.60e-04*** | -0.650 | Large | -9.0% |

**Figure S13 — Optimization method performance vs. random sampling baseline.** For each dataset-optimization method pair we compare the performance distribution to a random baseline. Whether the performance distribution is significantly different than that of the random baseline is quantified using p-values from a Wilcoxon rank-sum test. Color coding indicates statistical significance: green (p < 0.001), yellow (p<0.01), blue (p < 0.05), red (p > 0.05, not significant). Cliff's delta effect sizes serve to quantify the performance of the optimization method compared to the random baseline. Positive values indicate superior performance to random sampling. Green shading in the effect size column indicates positive effects, orange indicates negative effects.

## Bayesian Optimization Campaign Specifications

Each BO campaign begins with k=1 random initial point followed by 20 acquisition-guided iterations. For descriptor-enabled or continuous/mixed spaces, we use a BoTorch SingleTaskGP with a scaled Matern-5/2 kernel with automatic relevance determination (ARD) lengthscales and Gaussian observation noise; GP hyperparameters are optimized via multi-start L-BFGS-B with automatic retries. For fully categorical spaces without molecular descriptors, we use a categorical GP with a Hamming-distance kernel; for mixed categorical-continuous spaces without descriptors, we use a mixed GP that combines the categorical kernel on categorical dimensions with a Matern-5/2 kernel on continuous dimensions. In all datasets, no parameter is truly continuous. Thus, the acquisition function is computed for the full Cartesian product and the combination with the highest acquisition value is selected. We do not fix a global random seed across runs so that replicate campaigns are statistically independent.

Across all datasets, up to 10 features were selected for each categorical parameter, using either variance-based (few options, applied before standardization) or correlation-based filtering (applied after standardization), since correlation calculations become statistically unreliable across very few parameter options. Lastly, the features were standardized across all options for a categorical parameter. In all cases, any discrete parameters (i.e. concentration) are represented as float values. No objective data was used in feature selection. The exact descriptors can be found by name and value in the GitHub repository (https://github.com/gomesgroup/iron-mind-public/tree/main/computed_descriptors).

## K. Optimization Method Cost Analysis

**Table S2 — Optimizer cost and timetable.** Approximate cost (averaged over 20 independent campaigns) and timetable for a single campaign with batch size = 1 and budget = 20.

| Method | Compute cost | Time | Interpretability |
| --- | --- | --- | --- |
| BO (OHE) | ~$0 | Minutes | Black box |
| BO (Descriptors) | ~$0 | Hours (descriptor calc) | Feature importance |
| LLM-GO (o3) | ~$0.16 | Minutes | Full rationale |
| LLM-GO (claude-4-opus) | ~$2.14 | Minutes | Full rationale |
| LLM-GO (gemini-2.5-pro) | ~$0.20 | Minutes | Full rationale |

## L. Example Human Campaign — N-Alkylation/Deprotection

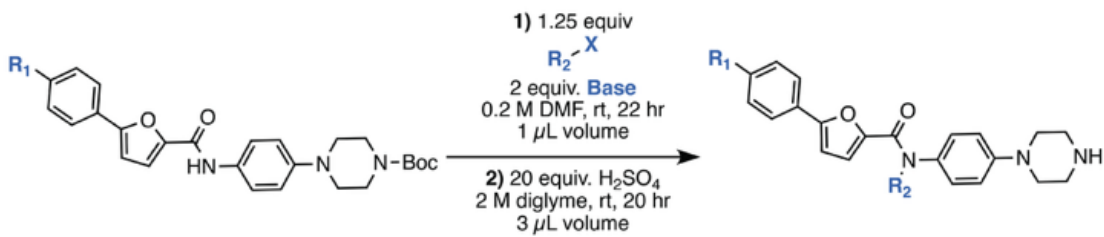

**Scheme S1 — Reaction scheme for N-Alkylation/Deprotection Reaction**. This dataset measures percent yield for a sequential N-alkylation/deprotection reaction on a nanomole scale across three key categorical parameters.[44] The parameters include: 1.) electrophile, 2.) core substrate, 3.) base. A total of 12 electrophiles, two cores, and four bases were considered. The dataset contains 96 measured yields, providing a more focused sampling of the parameter space compared to other datasets.

# Human Optimization Campaign Conducted by Dr. Luke Baldwin

Suggestion 1:

base: LiHMDS
core: tert-butyl 4-(4-(5-(4-chlorophenyl)furan-2-carboxamido)phenyl)piperazine-1-carboxylate
electrophile: iodomethane

Reasoning: No experimental has been acquired and my initial hypothesis is that less bulky groups with iodides will be good due to iodides being a good leaving group. This set of experiments tests reagents that I have experience using, LiHMDS is a non-nucleophilic base that I have hands on experience with so it will be a good place to start for my knowledge.

Yield: 55.0

-----------------------------------------------------------------------------------------------

Suggestion 2:

base: LiHMDS
core: tert-butyl 4-(4-(5-(4-cyanophenyl)furan-2-carboxamido)phenyl)piperazine-1-carboxylate
electrophile: iodomethane

Reasoning: I've reviewed the prior experiment which provided moderate yield and my hypothesis is still valid. Initially I thought that the aryl chloride of the core was more electron withdrawing which could help to make the amide more acidic but the substituent is very far away from the core so early on I'd like to check to see the role of this. I believe that this next experiment will help me understand the core.

Yield: 60.0

-----------------------------------------------------------------------------------------------

Suggestion 3:

base: LiHMDS
core: tert-butyl 4-(4-(5-(4-chlorophenyl)furan-2-carboxamido)phenyl)piperazine-1-carboxylate
electrophile: 1-bromo-3-methylbut-2-ene

Reasoning: Thus far the two cores don't seem to be that different based on data thus far so my hypothesis is that small electrophiles with a good leaving group are best. Allyl bromides are also pretty good electrophiles so testing this will allow me to determine if this is better or works than alkyl iodides.

Yield: 49.0

-----------------------------------------------------------------------------------------------

Suggestion 4:

base: LiHMDS
core: tert-butyl 4-(4-(5-(4-chlorophenyl)furan-2-carboxamido)phenyl)piperazine-1-carboxylate
electrophile: 2-(bromomethyl)pyridine

Reasoning: Thus far there hasn't been much difference in the yields but I think the electrophile is pretty important still because it can be a dominant factor for SN2 reactions, which my hypothesis is that is is that the first step of the reaction falls under that reaction class. For pyridine reagents, the ortho position might be partial positive due to resonance and that would make the carbon on the bromide more electrophilic. Plus the closeness to the nitrogen of the pyridine may stabilize a nucleophile.

Yield: 40.0

-------------------------------------------------------------------------------------------------

Suggestion 5:

base: P2-Et
core: tert-butyl 4-(4-(5-(4-chlorophenyl)furan-2-carboxamido)phenyl)piperazine-1-carboxylate
electrophile: 2-(bromomethyl)pyridine

Reasoning: The reactions thus far has maintained 40-60% yield but we've kept the same base, which is LiHMDS. I still think my hypothesis and rationale is correct but I've not encountered the cyclic nitrogen-phosphorous bases so need to understand the reactivity of these before continuing.

Yield: 100.0

-------------------------------------------------------------------------------------------------

Suggestion 6:

base: BTTP
core: tert-butyl 4-(4-(5-(4-chlorophenyl)furan-2-carboxamido)phenyl)piperazine-1-carboxylate
electrophile: 2-(bromomethyl)pyridine

Reasoning: Well I believe my hypothesis and rationale is correct since I found 100% yield but to confirm that I am checking the reactivity of BTTP since I am unsure what that base does. This would allow me to continue to determine what the primary factors are in this reaction.

Yield: 100.0

-------------------------------------------------------------------------------------------------

Suggestion 7:

base: NaOtBu
core: tert-butyl 4-(4-(5-(4-chlorophenyl)furan-2-carboxamido)phenyl)piperazine-1-carboxylate
electrophile: 4-(bromomethyl)pyridine

Reasoning: I just found another 100% yielding reaction so those nitrogen phosphorus bases must be super bases or strong enough to deprotonate the amide to make it a good nucleophile. Checking the alkoxide now which should be the least powerful.

Yield: 82.0

-------------------------------------------------------------------------------------------------

Suggestion 8:

base: P2-Et
core: tert-butyl 4-(4-(5-(4-chlorophenyl)furan-2-carboxamido)phenyl)piperazine-1-carboxylate
electrophile: bromomethylbenzene

Reasoning: The last reaction with the alkoxide showed that the basicity of the base matters but that it's not the only factor at play because HMDS has a pka of ~28 and the alcohol has a pka of ~16. The base is less intuitive but the benzene ring core should be less electrophilic than the ortho pyridine so the expectation is for this to have a lower yield based on SN2 reactivity.

Yield: 67.0

-------------------------------------------------------------------------------------------------

Suggestion 9:

base: P2-Et
core: tert-butyl 4-(4-(5-(4-chlorophenyl)furan-2-carboxamido)phenyl)piperazine-1-carboxylate
electrophile: 1-(bromomethyl)-2-phenylbenzene

Reasoning: So far the rationale seems to be holding and to reaffirm this the biphenyl should be even worse yield because the slight twist angle of the bi-phenyl should block the incoming nucleophile. This reaction will test that.

Yield: 55.0

-------------------------------------------------------------------------------------------------

Suggestion 10:

base: BTTP
core: tert-butyl 4-(4-(5-(4-chlorophenyl)furan-2-carboxamido)phenyl)piperazine-1-carboxylate
electrophile: 2-(bromomethyl)-1-methylimidazole

Reasoning: The campaign is working as intended and the imidaozle should be less reactive than the pyridine is the imidazole is less partial positive at the ring carbon. I believe that it should be less partial positive due to resonance but this experiment will help to resolve that uncertainty.

Yield: 80.0

-------------------------------------------------------------------------------------------------

Suggestion 11:

base: NaOtBu
core: tert-butyl 4-(4-(5-(4-chlorophenyl)furan-2-carboxamido)phenyl)piperazine-1-carboxylate
electrophile: 2-(bromomethyl)pyridine

Reasoning: We haven't checked the ortho and para effect too much which should be in resonance with each other creating a partial positive on both positions. I've done the para position with NaOtBu so this will allow me to check the ortho position now to ensure that the reactivities are similar that that this resonance hypothesis is true. Also the nitrogen may be assisting. But that is not clear at this time.

Yield: 100.0

-------------------------------------------------------------------------------------------------

Suggestion 12:

base: NaOtBu
core: tert-butyl 4-(4-(5-(4-chlorophenyl)furan-2-carboxamido)phenyl)piperazine-1-carboxylate
electrophile: 2-(bromomethyl)-6-methylpyridine

Reasoning: The last experiment showed that the ortho position plays more of a role than I'd expect. There aren't many pyridines with additional substituents but overall I'd think that electron donating groups will lead to decreased reactivity and the electrophile being selected in this batch will check that.

Yield: 76.0

-------------------------------------------------------------------------------------------------

Suggestion 13:

base: P2-Et
core: tert-butyl 4-(4-(5-(4-chlorophenyl)furan-2-carboxamido)phenyl)piperazine-1-carboxylate
electrophile: iodomethane

Reasoning: Based off the experiment performed thus far I haven't checked the reactivity of the iodomethane with this extra good base now. Origianally I focused on LiHMDS before findings that the P2-Et and other nitrogen-phosphorous base works very well. So rechecking this electrophile. My suspicion is that it has similar reactivity to the pyridines because the methyl group is so small and the iodide is a good leaving group.

Yield: 51.0

-------------------------------------------------------------------------------------------------

# Human Optimization Campaign Conducted by Dr. Jeffrey Ethier

Suggestion 1:

base: NaOtBu
core: tert-butyl 4-(4-(5-(4-cyanophenyl)furan-2-carboxamido)phenyl)piperazine-1-carboxylate
electrophile: 1-iodobutane

Reasoning: My hypothesis is that the core will not matter that much in the yield, and that the electrophile and base will impact the yield more. However, since this is the first round, I do not have any data to go off of. I also do not have chemistry intuition to base my first decision off of, so I am simply choosing the electrophile, core, and base at random to begin.

Yield: 67.0

-------------------------------------------------------------------------------------------------

Suggestion 2:

base: NaOtBu
core: tert-butyl 4-(4-(5-(4-chlorophenyl)furan-2-carboxamido)phenyl)piperazine-1-carboxylate
electrophile: 1-iodobutane

Reasoning: The previous yield was good at 67%. I will now test the previous hypothesis regarding the core to see if it impacts the yield. I am keeping the electrophile and base constant from the previous suggestion to determine if the core will have an effect on yield.

Yield: 52.0

-------------------------------------------------------------------------------------------------

Suggestion 3:

base: NaOtBu
core: tert-butyl 4-(4-(5-(4-cyanophenyl)furan-2-carboxamido)phenyl)piperazine-1-carboxylate
electrophile: 4-(bromomethyl)pyridine

Reasoning: Based on the previous data, the core does have a small effect on the yield (roughly 15% difference in yield). I do not know what part of the core is contributing to the better yield, but it appears that the end functional group affects the reaction yield, with the nitrogen performing better. Let's see whether choosing a brominated methyl pyridine electrophile will perform better than the iodobutane. My hypothesis is that the more bulky electrophiles will perform worse than the hydrocarbons with iodine. I am keeping the base constant for now.

Yield: 93.0

-------------------------------------------------------------------------------------------------

Suggestion 4:

base: NaOtBu
core: tert-butyl 4-(4-(5-(4-cyanophenyl)furan-2-carboxamido)phenyl)piperazine-1-carboxylate
electrophile: 2-(bromomethyl)pyridine

Reasoning: Based on the last result with a yield of 93%, my hypothesis was incorrect and in fact the brominated methyl pyridine improved the reaction yield significantly. Based on all previous data, it seems the nitrogen groups for the electrophile and core are playing an important role in the reaction yield. If so, moving the nitrogen on the pyridine molecule should change the yield significantly. I am keeping the core and base constant, since it appears that this base works well for this reaction.

Yield: 100.0

-------------------------------------------------------------------------------------------------

Suggestion 5:

base: P2-Et
core: tert-butyl 4-(4-(5-(4-chlorophenyl)furan-2-carboxamido)phenyl)piperazine-1-carboxylate
electrophile: 2-(bromomethyl)pyridine

Reasoning: It appears my previous hypothesis was indeed correct and that the position of the nitrogen group impacted yield, improving from 93% to 100%. Since it seems the nitrogen groups are playing a significant role in yield, I would like to explore other bases, keeping the electrophile and core the same as the prior suggestion, which reached 100% yield. If nitrogen on the pyridine and core are playing a role in this reaction, then changing the base to include more nitrogen atoms may also have a high yield, so I am suggesting to use P2-Et as the base.

Yield: 100.0

---

Suggestion 6:

base: P2-Et
core: tert-butyl 4-(4-(5-(4-chlorophenyl)furan-2-carboxamido)phenyl)piperazine-1-carboxylate
electrophile: 2-(bromomethyl)-1-methylimidazole

Reasoning: Based on the previous data, it seems my hypothesis is correct in that the nitrogen groups are playing a significant role in the reaction yield. I would like to test another electrophile that contains two nitrogen groups and is similar to the 2-(bromomethyl)pyridine. This should also produce a high yield, keeping the core and base constant from the previous suggestion.

Yield: 100.0

---

Suggestion 7:

base: P2-Et
core: tert-butyl 4-(4-(5-(4-cyanophenyl)furan-2-carboxamido)phenyl)piperazine-1-carboxylate
electrophile: bromomethylbenzene

Reasoning: Based on previous data, it appears that switching the electrophile to include an additional nitrogen atom also produces the highest yield, as expected. To test my previous hypothesis about nitrogen playing an important role, I would like to switch the electrophile to bromomethylbenzene, which removes the nitrogen groups. I hypothesize that removing the nitrogen group from the electrophile will significantly decrease the yield, keeping the core and base constant from the previous suggestion.

Yield: 39.0

---

Suggestion 8:

base: LiHMDS
core: tert-butyl 4-(4-(5-(4-cyanophenyl)furan-2-carboxamido)phenyl)piperazine-1-carboxylate
electrophile: 2-(bromomethyl)-1-methylimidazole

Reasoning: The previous result confirms that the nitrogen groups on the electrophile play an important role in improving the reaction yield. Now, I will explore a different base. Since there are two different bases that result in 100% yield, I hypothesize that only the nitrogen groups on the core and electrophile play an important role, and that the base is less impactful to the reaction yield. To test this, I will switch the base to one of the remaining untested bases. I chose LiHMDS since it contains a Lithium atom, which I think will hinder the reaction.

Yield: 14.0

---

Suggestion 9:

base: BTTP
core: tert-butyl 4-(4-(5-(4-chlorophenyl)furan-2-carboxamido)phenyl)piperazine-1-carboxylate
electrophile: 2-(bromomethyl)-1-methylimidazole

Reasoning: The hypothesis that the Li atom will hinder the reaction was confirmed. This idea was based on the assumption that metal ions will sometimes act as a poison to reaction catalysts, but I was not entirely confident. However, it does seem that the Li significantly impacted the yield, which dropped to 14%. I will now test the hypothesis that the last remaining base will increase the yield from the previous suggestion due to no metal ions being present. The last base that is untested, BTTP, contains nitrogen groups that are not very accessible, which may lower yield a little bit. Additionally, there is a phosphorous atom, which may hinder the reaction. I think the reaction yield will still be high, but not 100% as with the other two bases.

Yield: 80.0

-------------------------------------------------------------------------------------------------

Suggestion 10:

base: P2-Et
core: tert-butyl 4-(4-(5-(4-chlorophenyl)furan-2-carboxamido)phenyl)piperazine-1-carboxylate
electrophile: 2-(bromomethyl)-6-methylpyridine

Reasoning: The previous hypothesis seems to be confirmed in that the BTTP base gave a high yield, but not the maximum. This is likely due to the phosphorous atom, which is likely hindering the reaction conversion. Based on previous results for the 100% yield, I will now determine whether adding a methyl group to the 2-(bromomethyl)pyridine next to the nitrogen site hinders the reaction yield at all. My hypothesis is that this electrophile, the 2-(bromomethyl)-6-methylpyridine will do similarly well to the 2-(bromomethyl)pyridine.

Yield: 83.0

-------------------------------------------------------------------------------------------------

Suggestion 11:

base: NaOtBu
core: tert-butyl 4-(4-(5-(4-chlorophenyl)furan-2-carboxamido)phenyl)piperazine-1-carboxylate
electrophile: 2-(bromomethyl)-6-methylpyridine

Reasoning: It appears that based on the previous suggestion, the side methyl group on the pyridine hinders the reaction, which indicates that the active sites are not as accessible in the reaction. I will now test the same electrophile and core, changing the base to NaOtBu which also produced 100% yield even though there are no nitrogen groups present. I am unsure why this base did so well, my guess is that the oxygen is playing an important role similar to the nitrogen groups on the core and electrophile. Therefore, this base should do similar to the last suggestion to give around an 80% yield.

Yield: 76.0

-------------------------------------------------------------------------------------------------

Suggestion 12:

base: P2-Et
core: tert-butyl 4-(4-(5-(4-chlorophenyl)furan-2-carboxamido)phenyl)piperazine-1-carboxylate

electrophile: 2-(bromomethyl)-1-methylimidazole

Reasoning: The previous suggestion gave a 76% yield, close to 80%, indicating that the two best bases are indeed performing similarly (P2-Et and NaOtBu). Now that we have observed the two bases that work well in this reaction, I would like to confirm that the core with the nitrogen will still perform better than the chlorine atom attached at the end of the last phenyl ring, using the best electrophile and base.

Yield: 100.0

-----------------------------------------------------------------------------------------------

Suggestion 13:

base: NaOtBu
core: tert-butyl 4-(4-(5-(4-cyanophenyl)furan-2-carboxamido)phenyl)piperazine-1-carboxylate
electrophile: 2-(bromomethyl)pyridine

Reasoning: It seems that my original hypothesis on the core chemistry was correct after all, based on the previous suggestion. Hence, the core only impacts the yield when not using the optimal electrophile and base, but otherwise is does not change yield. The combination of the electrophile and base seem to be the most important. We can test this by changing the electrophile and base to the other two optimal chemistries, 2-(bromomethyl)pyridine and NaOtBu. Since those two gave 100% yield, I expect that changing the core will have no impact.

Yield: 100.0

-----------------------------------------------------------------------------------------------

Suggestion 14:

base: NaOtBu
core: tert-butyl 4-(4-(5-(4-chlorophenyl)furan-2-carboxamido)phenyl)piperazine-1-carboxylate
electrophile: 2-(bromomethyl)-1-methylimidazole

Reasoning: The previous suggestion contained an error. The core should have been the chlorinated one, not the one with the nitrogen group. The previous suggestion was already given in a previous round, and therefore this suggestion will now test the other core.

Yield: 84.0

-----------------------------------------------------------------------------------------------

Suggestion 15:

base: P2-Et
core: tert-butyl 4-(4-(5-(4-cyanophenyl)furan-2-carboxamido)phenyl)piperazine-1-carboxylate
electrophile: 1-bromo-3-methylbut-2-ene

Reasoning: Based on the previous suggestion, the result shows that yield drops from 100% to 84% when changing the core. This is likely due to the chlorine interaction with the oxygen group on the base molecule, since using the other optimal base (P2-Et) resulted in 100% yield. That tells me that the chlorine and oxygen atom are interacting in some way, or hindering the reaction. For my last suggestion, I would like to test another electrophile, using the chlorinated core and P2-Et base. This is purely exploratory, and should not result in the optimal yield. In fact, it is expected to be a poor yield due to no nitrogen groups being present on the electrophile, similar to the bromomethylbenzene.

Yield: 45.0
-------------------------------------------------------------------------------------------

## M. Alternative Analysis for Chan-Lam Coupling Dataset

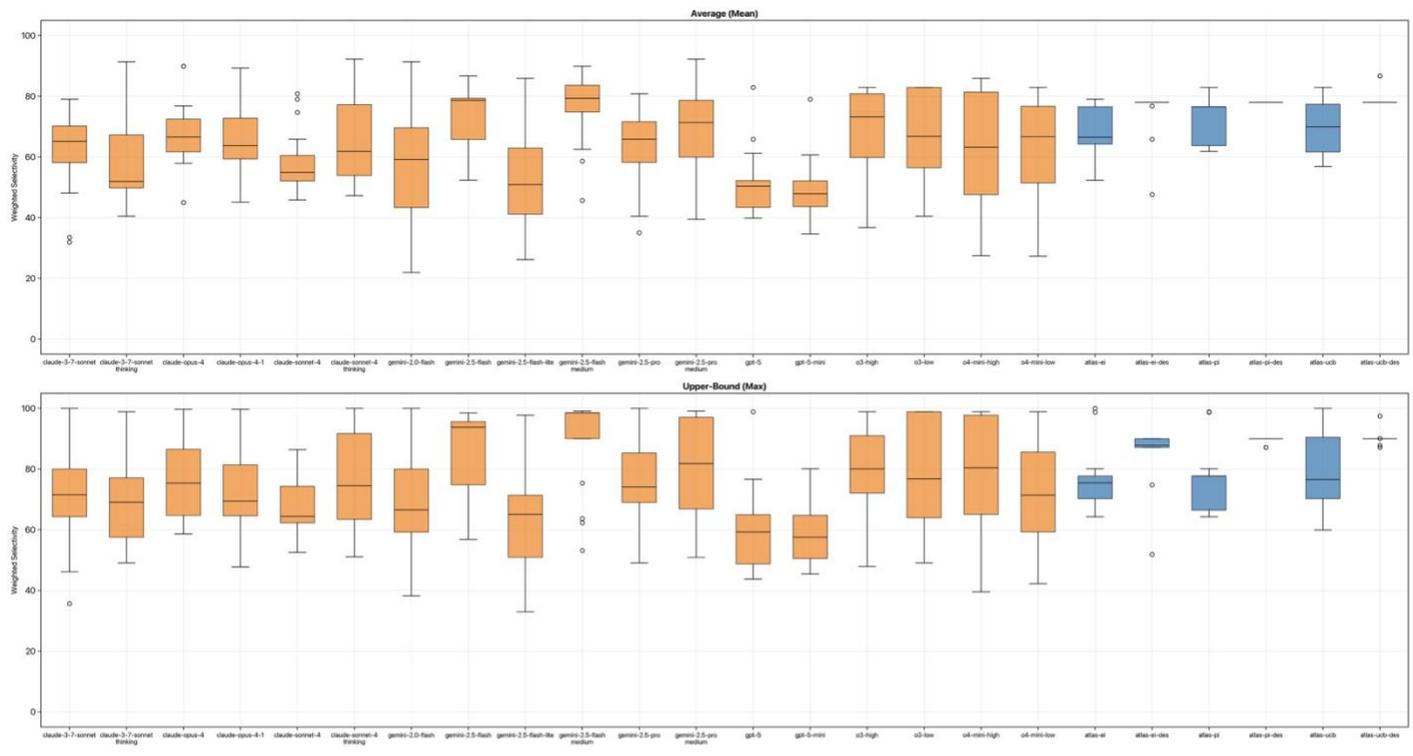

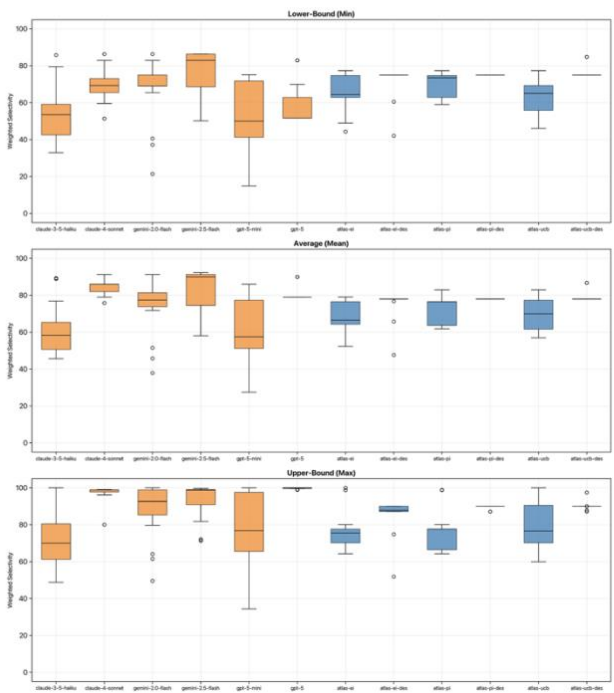

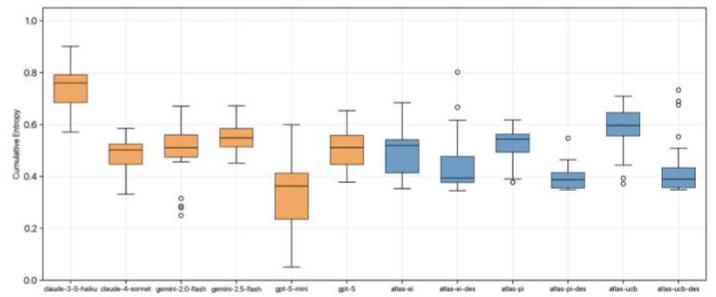

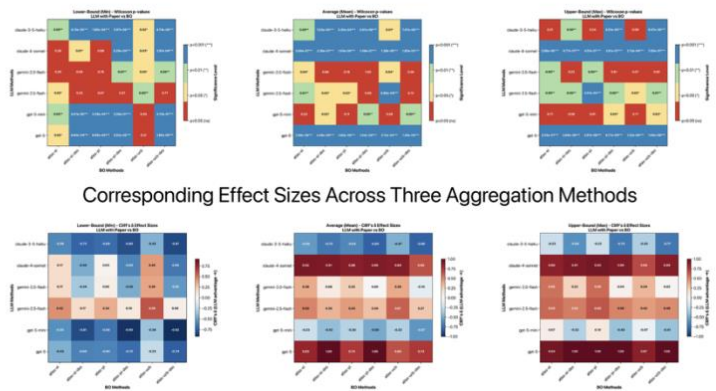

**Figure S14 — Extended Performance Analysis on Chan-Lam Dataset Using Alternative Aggregation Methods and Prompting Strategies.** A.) In the main text we present performance plots using lower-bound

aggregation within a group of measurements. Here we should comparisons using mean aggregation and upper-bound aggregation for all evaluated LLM (orange) and BO (blue) methods. B.) We investigate the impact of providing the full text of the paper presenting the Chan-Lam dataset as a prompt to the LLM. We present performance distributions across all three aggregation methods. C.) The corresponding selection diversity when providing the paper to the LLM-GO methods is shown. D.) Statistical analysis for each aggregation method is shown comparing LLM-GO with the paper to all BO methods.